\documentclass[lettersize,journal]{IEEEtran}
\usepackage{amsmath,amsfonts}
\usepackage{tabularx}
\usepackage{algorithmic}
\usepackage{algorithm}
\usepackage{array}
\newcommand{\cmark}{\ding{51}}%
\newcommand{\xmark}{\ding{55}}%
\usepackage[caption=false,font=normalsize,labelfont=sf,textfont=sf]{subfig}
\usepackage{textcomp}
\usepackage{stfloats}
\usepackage{verbatim}
\usepackage{graphicx}
\usepackage{cite}
\usepackage{multirow}
\usepackage{bbding}
\usepackage{pifont}
\usepackage{graphicx}
\usepackage{tikz}
\usepackage{multirow}
\usepackage{array}
\usepackage{caption}
\usepackage{colortbl}
\usepackage{makecell}
\usepackage{xcolor}
\usepackage{longtable}
\usepackage{ltxtable}
\usepackage{rotating}
\definecolor{datasetcolor}{RGB}{0,102,204}
\definecolor{networkcolor}{RGB}{255,140,0}
\definecolor{losscolor}{RGB}{0,128,0}
\usepackage{pgfplots}
\pgfplotsset{compat=newest}
\pgfplotsset{compat=1.16}
\usepackage{tikz}
\usepackage{pgfplots}
\pgfplotsset{compat=newest}
\usepackage{booktabs}   % nicer horizontal rules
\usepackage{pifont}     % check-/x-marks
\usepackage{tabularx}   % automatic column width
\usepackage{array} % for m{} column type
\usepackage{tabularx} % for table width management
\usepackage{ragged2e} % for justification in columns
\usepackage{booktabs} % for \toprule etc.
\usepackage{xcolor}   % for LightGray
\usepackage{caption}
\usepackage{array}
\usepackage{ragged2e}
\usepackage[table]{xcolor}
\usepackage{tabularx}
\usepackage{MnSymbol}
\usepackage{array}
\usepackage{placeins}

\usepackage[colorlinks=true, linkcolor=black, urlcolor=blue, citecolor=black]{hyperref}

\begin{document}

\title{Multi Camera Connected Vision System with Multi View Analytics: A Comprehensive Survey}

\author{Muhammad Munsif,~\IEEEmembership{Student Member,~IEEE}, Waqas Ahmad, Amjid Ali, Mohib Ullah, Adnan Hussain, Sung Wook Baik,~\IEEEmembership{Senior Member,~IEEE}
%\thanks{Manuscript received XXXX, XXXX; revised  XXXX, XXXX.}
\thanks{
This work was supported by the National Research Foundation of Korea (NRF) grant funded by the Korea Government, MSIT, Grant/Award Number: RS-2023-NR076686.
(Corresponding author: Sung Wook Baik)}
\thanks{ Muhammad Munsif, Waqas Ahmad, Amjid Ali, Mohib Ullah, Adnan Hussain, Sung
Wook Baik are with the Sejong University, Seoul 143-747, South Korea. (Email: munsif@ieee.org; waqasahmad@sju.ac.kr; amjidali3797@gmail.com; mohibullah@sju.ac.kr; Adnanhussain@sju.ac.kr; sbaik@sejong.ac.kr). }
}

% \markboth{Journal of \LaTeX\ Class Files,~Vol.~XXXX, No.~XXX, OCTOBER~2025}%


\maketitle

\begin{abstract}
Connected Vision Systems (CVS) are transforming a variety of applications, including autonomous vehicles, smart cities, surveillance, and human-robot interaction. These systems harness multi-view multi-camera (MVMC) data to provide enhanced situational awareness through the integration of MVMC tracking, re-identification (Re-ID), and action understanding (AU). However, deploying CVS in real-world, dynamic environments presents a number of challenges, particularly in addressing occlusions, diverse viewpoints, and environmental variability. Existing surveys have focused primarily on isolated tasks such as tracking, Re-ID, and AU, often neglecting their integration into a cohesive system. These reviews typically emphasize single-view setups, overlooking the complexities and opportunities provided by multi-camera collaboration and multi-view data analysis.
To the best of our knowledge, this survey is the first to offer a comprehensive and integrated review of MVMC that unifies MVMC tracking, Re-ID, and AU into a single framework. We propose a unique taxonomy to better understand the critical components of CVS, dividing it into four key parts: MVMC tracking, Re-ID, AU, and combined methods. We systematically arrange and summarize the state-of-the-art datasets, methodologies, results, and evaluation metrics, providing a structured view of the field's progression. Furthermore, we identify and discuss the open research questions and challenges, along with emerging technologies such as lifelong learning, privacy, and federated learning, that need to be addressed for future advancements. The paper concludes by outlining key research directions for enhancing the robustness, efficiency, and adaptability of CVS in complex, real-world applications. We hope this survey will inspire innovative solutions and guide future research toward the next generation of intelligent and adaptive CVS.

\end{abstract}


\section{Introduction}
\begin{figure}
    \centering
    \includegraphics[width=\linewidth, height=0.3\textheight]{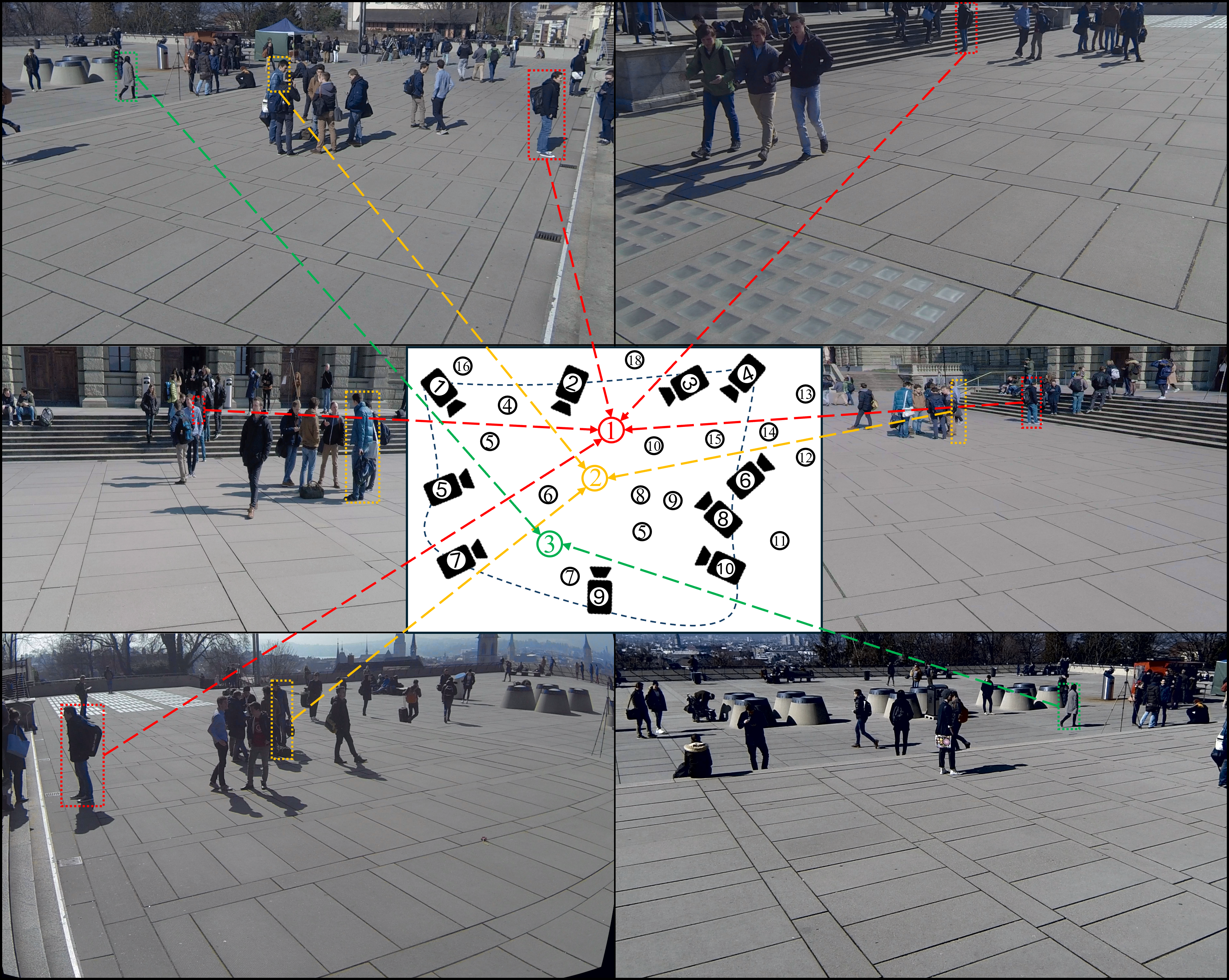}
    \caption{ Demonstration of a Multi-Camera Connected Vision System with synchronized views and a central diagram showing camera placement and overlapping views. Colored bounding boxes track the same individual across multiple angles, showcasing person tracking and multi-view analysis. The top-left corner image is captured with Camera 1, the top-right with Camera 4, the middle-left with Camera 8, the middle-right with Camera 10, the bottom-left with Camera 3, and the bottom-right with Camera 2.}
    \label{fig:Concept}
\end{figure}

\IEEEPARstart{C}{onnected}  vision system (CVS)  as illustrated in Figure \ref{fig:Concept}, refers to a computer vision (CV) assisted collaborative and distributed framework where multiple visual sensors operate in synchronization to capture, analyze, and share visual data, facilitating comprehensive understanding of dynamic scenes \cite{MobileYi}. The advancement in high-performance computing, sophisticated communication protocols, and scalable algorithmic solutions has notably accelerated the practicality of automated surveillance video analytics, enabling systems to efficiently extract meaningful information from continuous video streams. This information is then synthesized 
into interpretable formats, supporting real-time monitoring or post-event analysis and decision-making \cite{Jain2020Spatula}. Core CV tasks integrated in CVS frameworks, including multi-view multi-camera (MVMC) object tracking \cite{Shim_2025_CVPR}, re-identification (Re-ID) \cite{ran2025camera}, and MVMC action understanding (AU) \cite{nguyen2024multi} to form cohesive and powerful analytical pipelines \cite{MobileYi}. Unlike standalone vision modules, connected MVMC leverage synchronized multi-view inputs, effectively addressing complex challenges such as occlusion, maintaining object identities across diverse camera perspectives, and accurately interpreting behaviors \cite{tran2024efficient}. Such capabilities are particularly indispensable in contexts demanding extensive situational awareness, including smart surveillance systems\cite{yu2025citytrac}, autonomous vehicles \cite{sun2024multiple}, intelligent human-robot interactions \cite{nguyen2024multi}, and distributed scene understanding across multiple viewpoints \cite{peng2024prototype}.
By aggregating spatial-temporal information from an array of sensors, CVS significantly enhances the robustness and semantic richness of analytics, facilitating higher-level interpretations such as group behavior analysis and anomaly detection \cite{pereira2024video}. Consequently, CVS form a critical backbone for emerging applications in smart cities and collaborative robotics, where comprehensive real-time visual perception is essential for informed and intelligent decision-making.

\begin{figure*}
    \centering
    \includegraphics[width=1.0\linewidth, trim=16pt 18pt 18pt 12pt, clip]{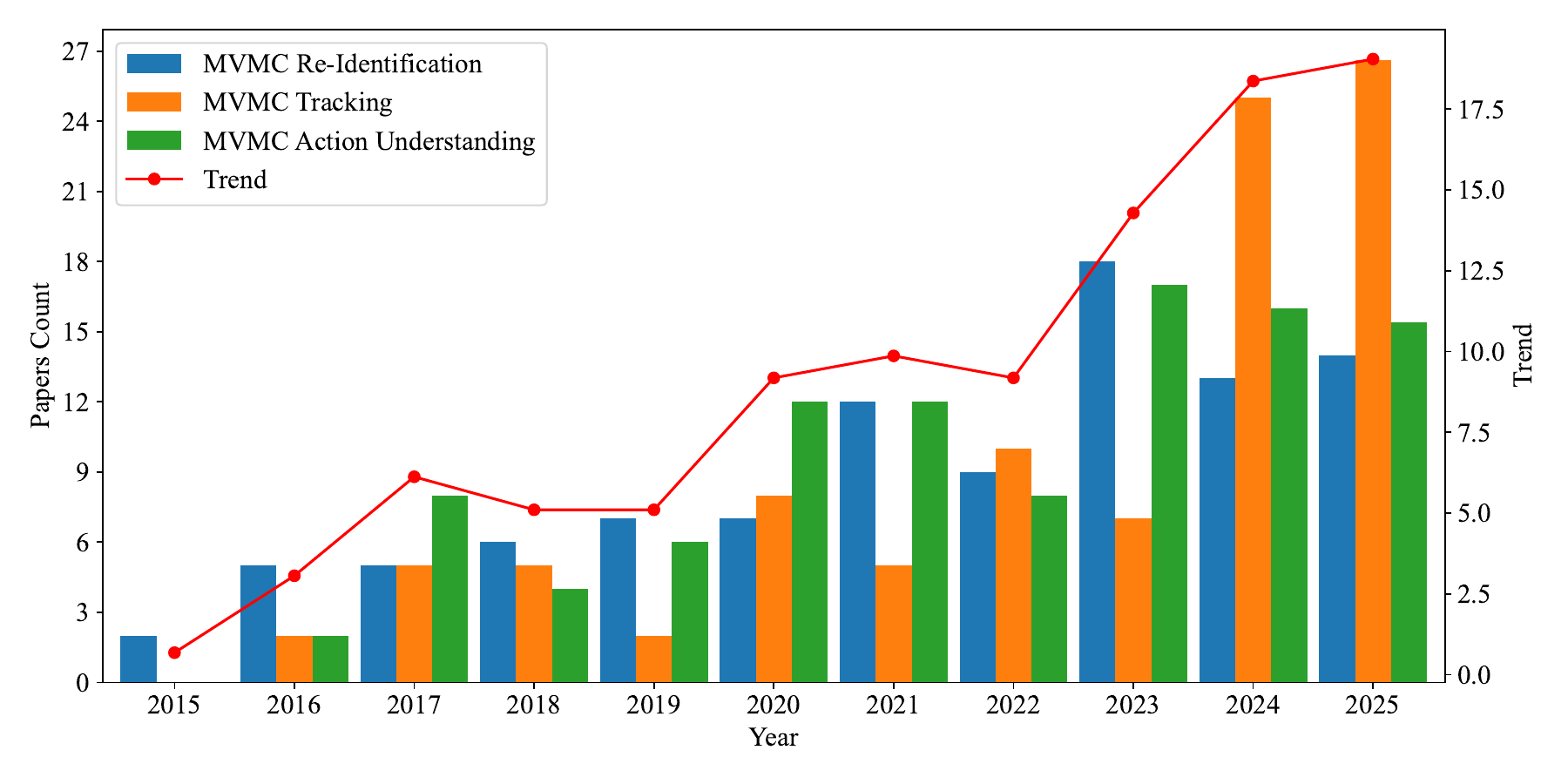}
    \caption{Annual publication trend (2015–August 2025) for multi-camera multi-view research, segmented into three core tasks: multi-view tracking, multi-camera re-identification, and multi-view action understanding. The 2025 data is extrapolated from data available for January–August to ensure comparability with prior years. }
    \label{fig:2}
\end{figure*}
\subsection{Unique Characteristics of Multi Camera Multi View CVS}
Connected MVMC vision systems are distinguished by their ability to significantly enhance scene understanding through the synchronized capture of visual data from multiple perspectives~\cite{MobileYi}. This integration offers clear advantages over single-camera systems, particularly in mitigating challenges such as occlusions, tracking ambiguities, and limited fields of view. At the core of intelligent CVS are key tasks including multi-object tracking\cite{nguyen2024multi}, Re-ID\cite{ye2021deep}, and behavior analysis\cite{shah2023multi}, which together enable comprehensive interpretation of complex and dynamic environments. One of the most critical capabilities of multi-camera configurations is MVMC tracking, as illustrated in Figure 1. Operating across large-scale camera networks from isolated spaces to extensive surveillance grids MVMC tracking improves robustness by associating targets across overlapping views and connected sensors. This ensures continuous and accurate tracking, even in highly dynamic and cluttered settings, making CVS indispensable in scenarios where long-term, identity-consistent monitoring from multiple viewpoints is essential for real-time interpretation and decision-making. Multi-camera Re-ID \cite{woo2024mtmmc} is another fundamental component, maintaining identity consistency across different camera views, including non-overlapping configurations. Re-ID algorithms determine whether a specific individual or object appears across multiple frames at different times or locations, thereby enhancing the reliability of large-scale surveillance systems. When tracking and Re-ID are successfully integrated, multi-camera systems excel at AU by aggregating and synchronizing data from multiple viewpoints~\cite{gao2024hypergraph}. This enables the system to infer not only where an object or individual is, but also what they are doing, even under challenging conditions such as occlusions, environmental noise, or drastic viewpoint changes. Such robust multi-perspective AU is vital for applications such as smart surveillance and other intelligence systems. Collectively, these capabilities position CVS as a powerful framework for achieving high-precision, real-time situational awareness and a rich semantic understanding of dynamic environments, making it an essential enabler for next-generation applications in diverse domains.
\subsection{Motivation}

\begin{figure}[!t]
    \centering
    \includegraphics[width=1.00\linewidth, trim=10pt 5pt 10pt 10pt, clip]{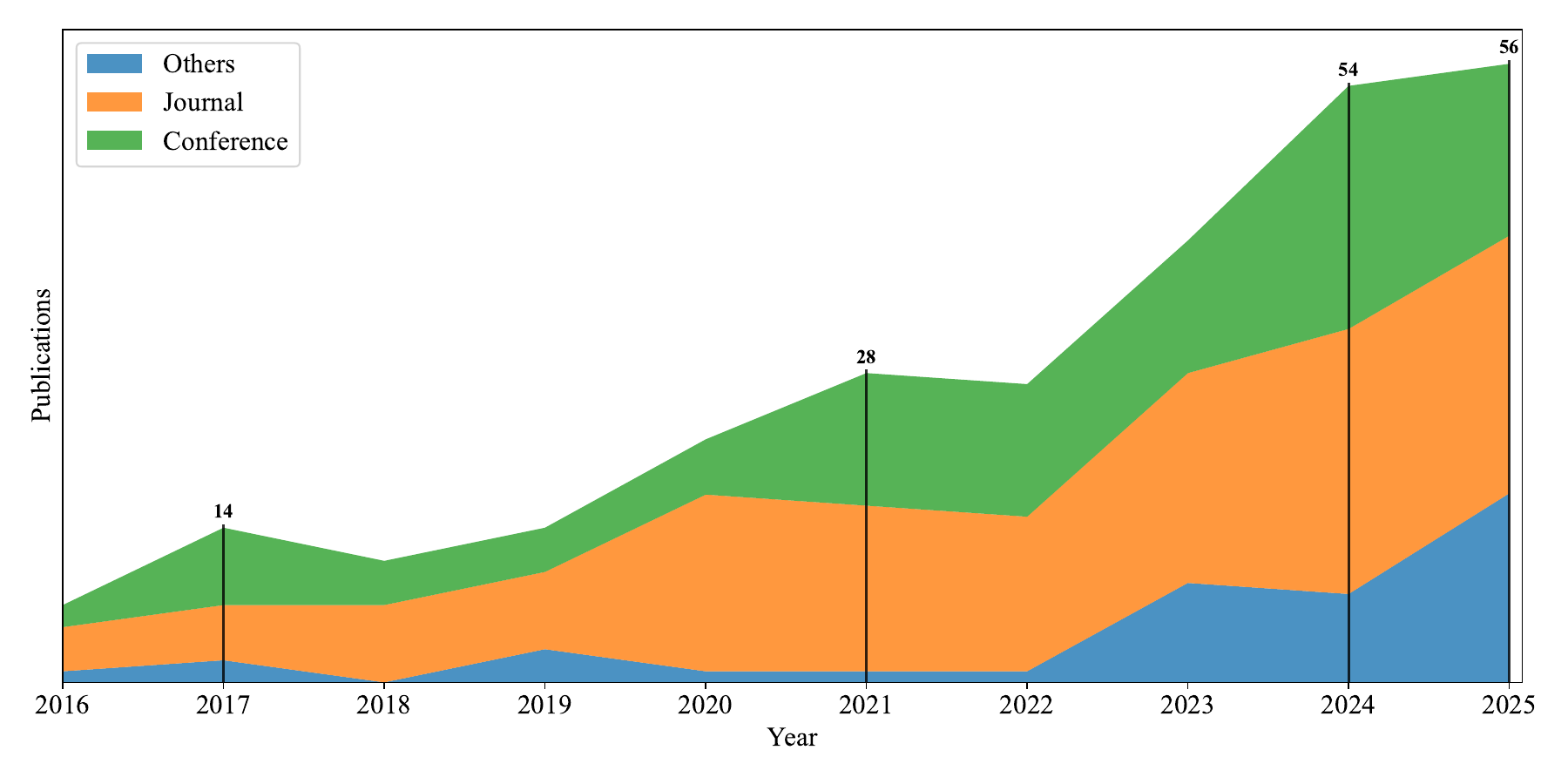}
    \caption{Distribution of MVMC-related publications across major academic venues. The 2025 data is extrapolated from data available for January–August to ensure comparability with prior years.}
    \label{fig:3}
\end{figure}

\begin{figure*}
    \centering
    \includegraphics[width=1\linewidth, trim=120pt 80pt 120pt 80pt, clip]{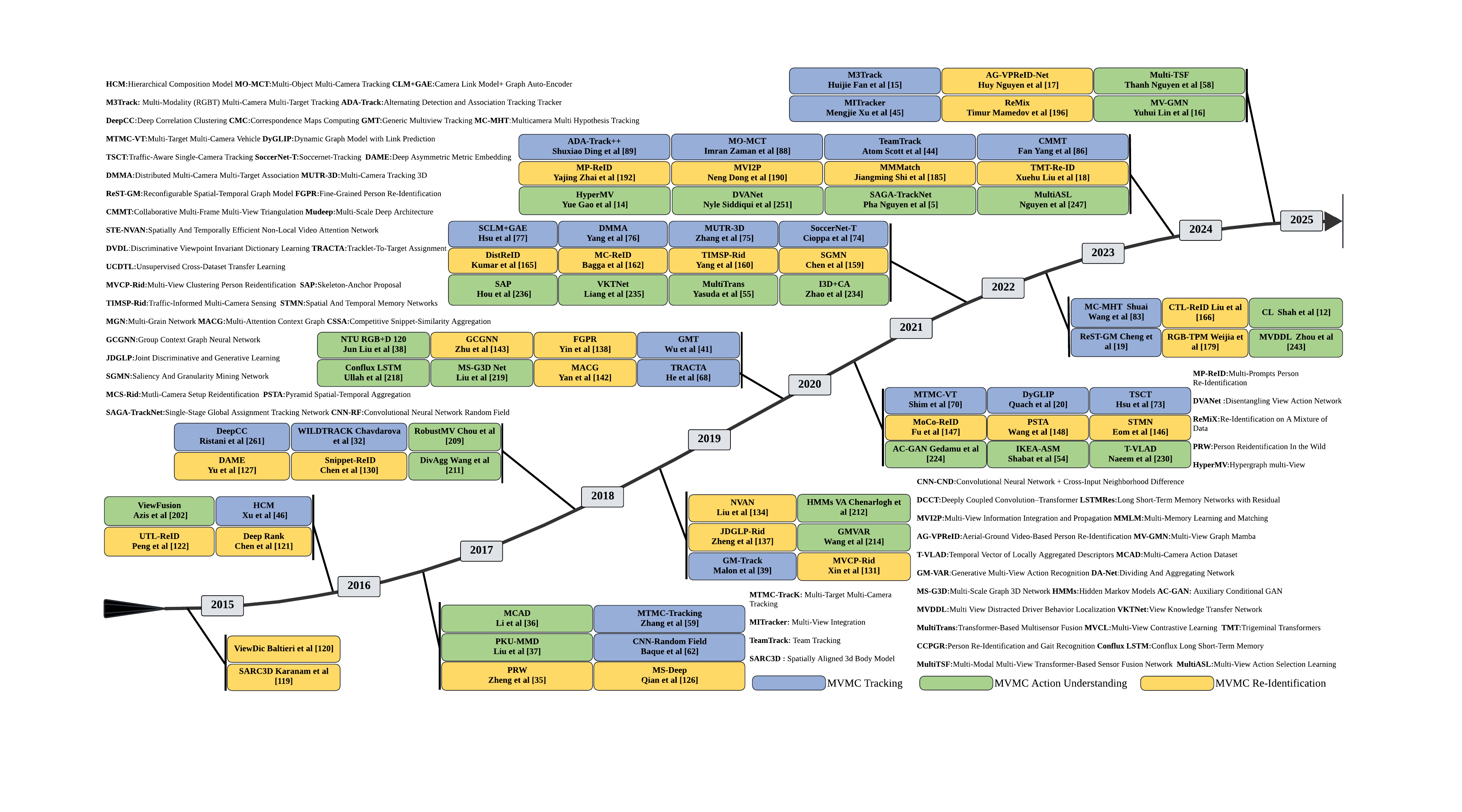}
    \caption{Advancements in multi-view, multi-camera tracking, Re-ID, and action understanding (2015 to 2025), highlighting key developments and innovations in each area over the past decade. }
    \label{Fish}
\end{figure*}

In recent years, research on connected MVMC  has accelerated rapidly, driven by advances in deep learning (DL) architectures, high-bandwidth communication protocols, and edge–cloud computational frameworks. As shown in Figure~\ref{fig:2}, the volume of scholarly contributions addressing MVMC-related tasks, including multi-view tracking (MV-T), Re-ID, and MVMC-AU has grown substantially since 2019, reaching its peak in 2025. The distribution of publication venues across journals (38\%), conferences (36\%), and other formats (26\%) (Figure~\ref{fig:3}) reflects the interdisciplinary nature of this research domain, where progress relies on both algorithmic innovation and practical system-level implementation.
The growing relevance of MVMC research is closely linked to the operational demands of large-scale surveillance, intelligent transportation, and public safety applications, where continuous and reliable monitoring across heterogeneous environments is critical. Single-camera systems often fail to sustain target identity through occlusions, viewpoint changes, and illumination variations, whereas synchronized multi-camera configurations effectively address these limitations through cross-view association. Recent advancements in MVMC representation learning, structurally shown in Figure \ref{Fish}, state space model (SMM) based learning  \cite{fan2025all,lin2025mv }, attention-based spatio-temporal modeling\cite{nguyen2025ag, nguyen2024multi,liu2024video}, and graph-based multi-camera association \cite{gao2024hypergraph,cheng2023rest, quach2021dyglip} have significantly improved tracking continuity and Re-ID accuracy in unconstrained environments. The availability of large-scale benchmark datasets with cross-view annotations has further catalyzed progress, enabling MVMC systems to evolve from low-level detection and tracking toward higher-level semantic understanding, including behavior analysis, anomaly detection, and predictive modeling.
Several surveys tabulated in Table \ref{tab:survey-comparison} have reviewed related CV tasks, yet they fall short of providing an integrated perspective across the full MVMC pipeline. For example, Pareek~\cite{pareek2021survey} and Kong \textit{et al.}~\cite{kong2022human} offer comprehensive reviews of human action recognition (HAR) techniques but focus primarily on single-view trimmed videos, lacking considerations for cross-camera identity consistency or spatial-temporal alignment. Marvasti \textit{et al.}~\cite{marvasti2021deep}, Sun \textit{et al.}~\cite{amosa2023multi}, and Zhang \textit{et al.}~\cite{zhang2024comprehensive} provide valuable insights into visual tracking, yet emphasize monocular or limited-view setups without addressing large-scale inter-camera coordination. Ye \textit{et al.}~\cite{ye2021deep} and Nayak \textit{et al.}~\cite{nayak2025comprehensive} focus on person Re-ID but largely prioritize image-based or non-overlapping camera configurations, overlooking integration with tracking and multi-view geometric reasoning. Surveys on multi-modal and ambient sensing, such as Arrotta \textit{et al.}~\cite{arrotta2025multi}, Stergiou \textit{et al.}~\cite{stergiou2025time}, and Sun \textit{et al.}~\cite{sun2022human}, address broader sensing modalities yet rarely consider multi-view video fusion or spatial reasoning across distributed camera networks. These limitations, as summarized in Table~\ref{tab:survey-comparison}, highlight the need for a unified and structured review that spans MVMC Tracking, MC-ReID, and MVMC-AU within a connected vision framework to consolidate state-of-the-art datasets, methodologies, results, evaluation metrics, and open challenges, providing both a comprehensive reference for researchers and practical guidance for deployment in real-world systems. 

\subsection{Contributions}
The limitations identified in existing survey papers underscore the need for a comprehensive and integrated review of CVS. To address this gap, to the best of our knowledge, for the first time, we present a structured and in-depth survey that consolidates current knowledge, methods, datasets, and evaluation metrics, identifies open challenges, and provides actionable insights for advancing the field. Given the growing importance of CVS in real-world applications, this work aims to serve as both a scholarly reference and a practical guide for researchers and practitioners. The main contributions of this survey are as follows:

\begin{enumerate}
   \item We conceptualize the CVS and its interdependent tasks, providing a unified framework that integrates MVMC tracking, Re-ID, and AU. Building on this foundation, we conduct a comprehensive survey of CVS research spanning the past decade (2015–2025), encompassing both peer-reviewed and non–peer-reviewed sources. The analysis includes a structured taxonomy of methods and a temporal breakdown of publications, offering valuable insights into the developmental trajectory of CVS and identifying emerging directions for future research.

    \item \textbf Unlike prior surveys that address tracking, Re-ID, and AU in isolation, we emphasize their interdependence within a connected MVMC system. We propose a unified taxonomy that captures how these core tasks operate synergistically in distributed camera networks, enabling robust, real-time scene understanding and comprehensive situational awareness.

    \item  Conducted a systematic review of state-of-the-art datasets and methods developed for CVS between 2015 and 2025, covering multi-view tracking, multi-camera Re-ID, and multi-view AU. Our analysis consolidates widely used datasets and evaluation metrics, and compares state-of-the-art performance across benchmark tasks, offering a clear perspective on the current capabilities and limitations of existing approaches.

    \item  We identify persistent challenges in deploying large-scale, real-world CVS, including scalability, cross-domain generalization, privacy preservation, and multi-modal integration. Building on these observations, we propose future research directions aimed at advancing the robustness, efficiency, and semantic richness of MVMC system, with an emphasis on bridging the gap between academic prototypes and operational deployments.
\end{enumerate}
\subsection{Organization}
The rest of this survey is organized as follows: In Section II, we outline the systematic method employed to curate relevant literature. Section III provides an overview of the most important connected vision MVMC datasets, accompanied by their respective test protocols. Our proposed taxonomy is presented in Section IV. In Section V, studied state-of-the-art techniques, shedding light on their evolutionary trends in recent years. Lastly, Section VI discusses the challenges of MVMC representation learning and outlines potential directions for future research, followed by Section VII, which concludes this survey paper.
\section{REVIEW METHODOLOGY}
A rigorous search protocol was executed to ensure the replicability of this survey and foster its utility for future researchers. Scholarly articles were initially identified through search engines such as Google Scholar and reputable digital libraries, including the ACM Digital Library, ScienceDirect, IEEE Xplore, and the CVF Open Access repository. The search strategy combined a wide range of carefully selected keywords relevant to the theme of multi-camera CVS and multi-view analytics, including terms such as “multi-view tracking,” “multi-camera AU,” “multi-view action recognition,” “person Re-ID,” “multi-camera systems,” “connected vision,” “multi-view analytics,” “multi-sensor fusion,” “camera networks,” “cross-view matching,” “multi-camera object detection,” “spatio-temporal reasoning,” and “distributed vision systems.” Following the initial retrieval of publications, the results were refined by excluding articles that either lacked direct relevance to CVS or failed to demonstrate sufficient technical rigor. Specifically, studies were excluded if they relied solely on non-vision sensors for scene or AU, did not introduce a novel methodology or framework, evaluated their findings only on proprietary or non-standard datasets, or neglected to benchmark results against contemporary state-of-the-art approaches. In cases where auxiliary modalities such as scalar or IoT data were combined with camera networks, the emphasis of this survey remained on the vision-centric components. To maintain consistency with the evolution of the field, a temporal constraint was imposed by considering only publications after 2015, coinciding with the emergence of scalable multi-camera tracking and recognition frameworks. To further enhance comprehensiveness, forward and backward citation tracking was applied, whereby both the works citing the initially retrieved articles and the references within these works were examined. This iterative process was continued until a robust collection of the most pertinent and technically rigorous publications was assembled, resulting in a definitive body of literature addressing techniques, methodologies, and frameworks for multi-camera, multi-view analytics within connected vision systems.
\definecolor{LightGray}{rgb}{0.9,0.9,0.9}
\newcolumntype{C}[1]{>{\centering\arraybackslash}m{#1}}
\newcolumntype{L}[1]{>{\raggedright\arraybackslash}m{#1}}
\newcolumntype{J}[1]{>{\justifying\noindent\arraybackslash}m{#1}}
\renewcommand{\arraystretch}{1.30}
% Make X-columns behave like m{<width>} so content is vertically centered
\renewcommand\tabularxcolumn[1]{m{#1}}
\sloppy
\begin{table*}[htbp]
\centering
\caption{Recent related survey papers on CVS‐focused tasks (tracking, re-identification, action understanding). Citation counts from Google Scholar, Aug.\ 2025.}
\renewcommand{\arraystretch}{1.15}
\begin{tabularx}{\textwidth}{@{}
        C{0.55cm}      % #
        L{2.7cm}       % Ref (now centered)
        C{1.35cm}      % Citation
        C{2cm}         % Venue
        *{3}{C{1.1cm}} % MV-T, MV-RID, MV-AU
       J{5.3cm} % Description (centered; X now vertically centered via \tabularxcolumn)
@{}}
\toprule
\multirow{2}{*}{\#} &
\multirow{2}{*}{\centering\arraybackslash Ref} &
\multirow{2}{*}{Citation} &
\multirow{2}{*}{Venue} &
\multicolumn{3}{c}{Focus} &
\multirow{2}{*}{\centering\arraybackslash Description} \\[-0.4em]
\cmidrule(lr){5-7}
& & & & MV-T & MC-RID & MV-AU & \\ \midrule
1  & P.~Pareek~\cite{pareek2021survey}                & 429  & AIR,21  & \xmark & \xmark & \cmark & Focuses on HAR; surveys ML/DL techniques, datasets, and challenges.\\
\hline
2  & M.~Ye \textit{et al.}~\cite{ye2021deep}         & 2329 & TPAMI,21& \xmark & \cmark & \xmark & Reviews person re-identification in closed- and open-world settings.\\
\hline
3  & S.~M.~\textit{et al.}~\cite{marvasti2021deep}   &  452 & TITS,21 & \cmark & \xmark & \xmark & Surveys deep-learning visual tracking; highlights challenges and directions.\\
\hline
4  & Y.~Kong \textit{et al.}~\cite{kong2022human}    &  953 & IJCV,22 & \xmark & \xmark & \cmark & Reviews DL-based HAR, datasets, evaluation metrics, and challenges.\\
\hline
5  & Z.~Sun \textit{et al.}~\cite{sun2022human}      &  817 & TPAMI,22& \xmark & \xmark & \cmark & Surveys single/multi-modal HAR, fusion strategies, and benchmarks.\\
\hline
6  & T.~I.~\textit{et al.}~\cite{amosa2023multi}     &  107 & NEUCOM,23& \cmark & \xmark & \xmark & Reviews multi-object tracking; outlines challenges and future work.\\
\hline
7  & Z.~Sun \textit{et al.}~\cite{sun2024multiple}   &    5 & TCSVT,24 & \cmark & \xmark & \xmark & Surveys multiple-pedestrian tracking, localisation, and MOT benchmarks.\\
\hline
8  & H.~Zhang \textit{et al.}~\cite{zhang2024comprehensive} &21 & TIM,24 & \cmark & \xmark & \xmark & Focuses on RGBT tracking; categorises evaluations and benchmarks.\\
\hline
9  & L.~Arrotta \textit{et al.}~\cite{arrotta2025multi}     &   0 & ESWA,25 & \xmark & \xmark & \cmark & Reviews multi-object HAR methods and datasets; highlights challenges.\\
\hline
10 & A.~Stergiou \textit{et al.}~\cite{stergiou2025time}   &   1 & IJCV,25 & \xmark & \xmark & \cmark & Explores action understanding with a focus on temporal modelling.\\
\hline
11 & K.~Nayak \textit{et al.}~\cite{nayak2025comprehensive} &   0 & ACM, 2025 & \xmark & \cmark & \xmark & Surveys group Re-ID across cameras; discusses layout variation, occlusion.\\
\hline
12 & \textbf{Ours}                                   &  --  & --        & \cmark & \cmark & \cmark & Review of MVMC tracking, Re-ID, AU methods, datasets, challenges, and future directions.\\
\bottomrule
\end{tabularx}
\label{tab:survey-comparison}
\end{table*}
\section{EVALUATION PROTOCOLS AND DATASET}
\subsection{Protocols}
Evaluation protocols for multi-camera CVS, particularly in tracking, Re-ID, and multi-view AU, can generally be classified into three categories: subject-dependent, subject-independent, and cross-view protocols. In the subject-dependent setting, both training and testing sets include data samples from all participants or identities, which helps assess performance under intra-subject variations but may limit generalization. In contrast, subject-independent protocols ensure that the test subjects or identities are entirely distinct from those in the training set, thereby evaluating the robustness of models against unseen actors or objects. Cross-view evaluation, which is particularly critical for multi-camera systems, involves separating training and testing data across distinct camera viewpoints to assess the generalization ability of models when exposed to unseen camera perspectives. Using these evaluation schemes, models are typically trained on one subset of the data and then tested by deriving features from the held-out samples, after which predictors or embedding-based matchers assign action categories, object identities, or trajectories. Beyond these standard setups, several nuanced considerations arise in multi-camera scenarios. It is essential to account for differences in camera viewpoints, calibration, synchronization, field of view overlap, and occlusion levels, all of which can heavily impact evaluation. Datasets such as DukeMTMC~\cite{ristani2016performance} and EPFL-RLC ~\cite{chavdarova2017deep} provide synchronized campus-level surveillance streams that support cross-subject and cross-view protocols for multi-target tracking, while WILDTRACK~\cite{chavdarovawildtrack} and MultiviewX focus explicitly on cross-view pedestrian detection and tracking with overlapping fields of view. In contrast, large-scale datasets like MMPTRACK~\cite{han2023mmptrack} and MTMMC~\cite{woo2024mtmmc} emphasize robustness in real-world conditions by incorporating multiple RGB and thermal cameras, enabling evaluation of multimodal fusion under subject-independent, cross-camera and view splits
. Similarly, for Re-ID tasks, datasets like DukeMTMC-ReID~\cite{zheng2017unlabeled} and PRW~\cite{zheng2017person} enforce cross-view protocols where identities must be matched across disjoint cameras. For AU, datasets such as MCAD~\cite{li2017multi}, PKU-MMD~\cite{liu2017pku}, and NTU RGB+D 120~\cite{liu2019ntu} explicitly define cross-view and cross-subject splits, ensuring that models are evaluated not only on actor variability but also on their ability to generalize across drastically different viewpoints. An exhaustive evaluation also requires the documentation of environmental conditions (indoor vs. outdoor, cluttered vs. controlled scenes), sensor configurations (RGB, depth, thermal, or multimodal), and challenges like occlusion, crowd density, and motion blur that are inherent in multi-camera systems. Data augmentation strategies, such as synthetic viewpoint generation or simulation of occlusions, have been increasingly employed to improve robustness. Performance is typically reported in terms of multi-object tracking metrics (MOTA, HOTA, IDF1), Re-ID metrics (mAP, rank-k accuracy), and action recognition (AR) metrics (Top-1/Top-5 accuracy, F1-score, average precision), depending on the task domain. Across the literature, these evaluation protocols subject-dependent, subject-independent, cross camera, and cross-view, have been widely adopted and adapted to match the unique challenges posed by connected multi-camera vision systems, and the datasets summarized in Table \ref{table:datasets_tracking} to \ref{table:datasets_action} serve as the backbone for benchmarking state-of-the-art solutions.
%\vspace{-0.577em}
\subsection{Datasets for MVMC Connected Vision Systems}
\begin{figure*}[!b]
    \centering
    \includegraphics[width=1.00\linewidth, trim=10pt 10pt 15pt 10pt, clip]{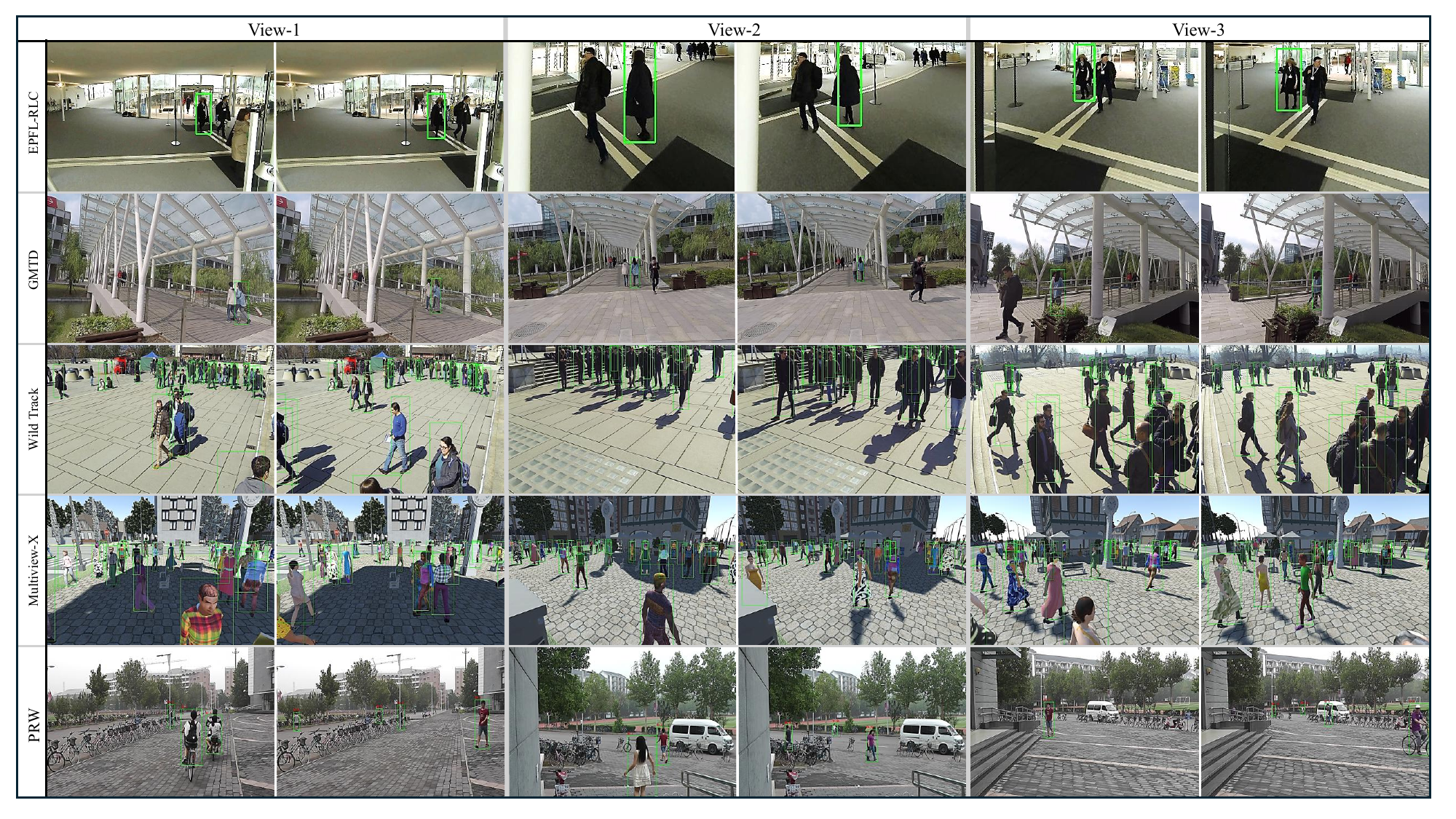}
    \caption{Samples from publicly available datasets of MVMC, multi-object tracking, and Re-ID in a CVS. It presents various tracking and Re-ID scenarios across multiple camera views. }
    \label{fig=Tracking_R_ID_Samples}
\end{figure*}

\begin{figure*}[!b]
    \centering
    \includegraphics[width=1.001\linewidth, trim=16pt 20pt 20pt 10pt, clip]{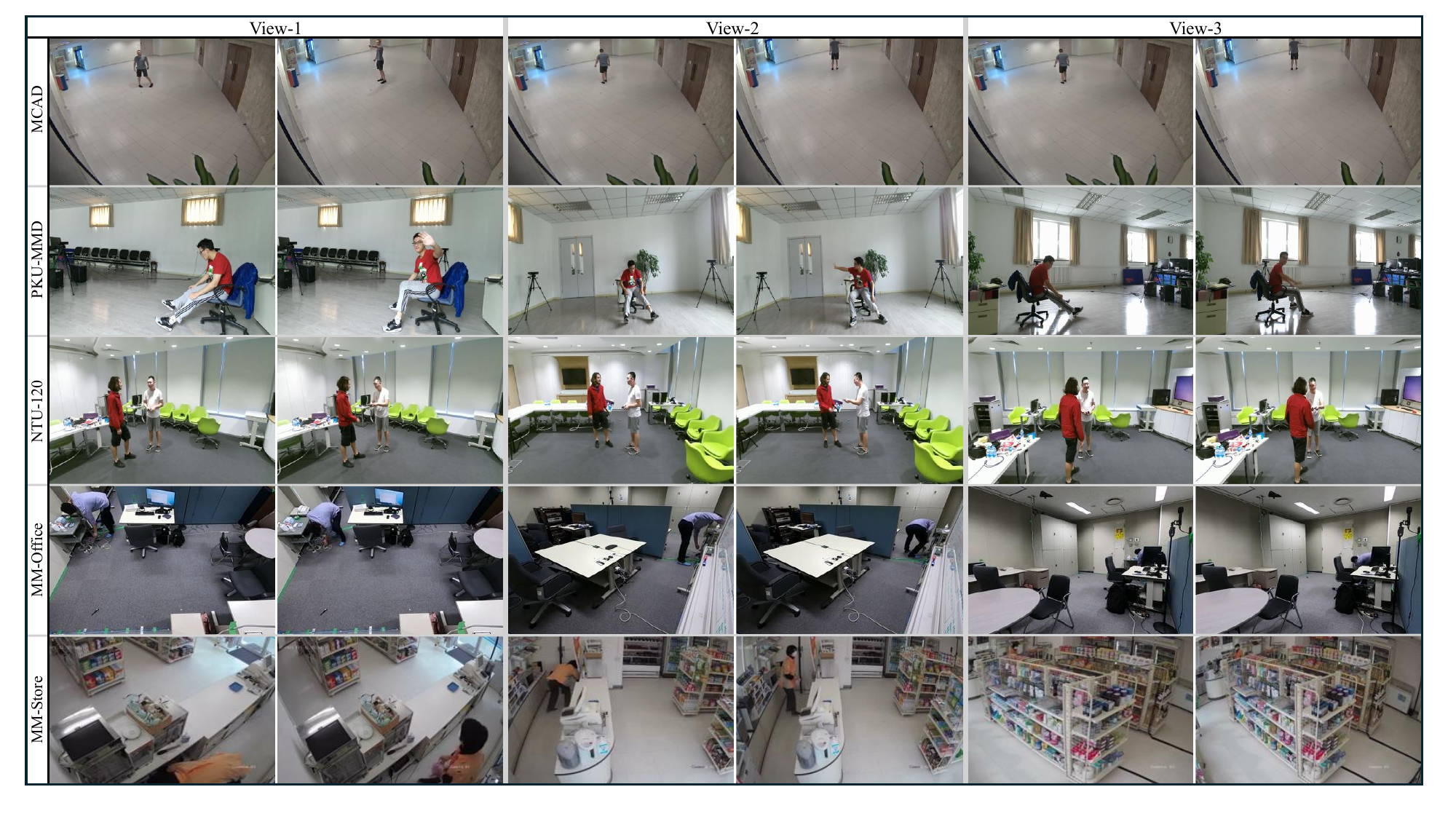}
    \caption{Samples from multi-camera, multi-view AU datasets. Presents video samples captured from different camera perspectives (View-1, View-2, View-3) across various datasets.}
    \label{fig=Action Recognition_Samples}
\end{figure*}

The effectiveness of multi-camera CVS heavily relies on the availability of robust datasets that facilitate the evaluation of algorithms in complex real-world settings. This section presents a comprehensive overview of key datasets employed for MVMC multi-object tracking (MOT), object Re-ID, and AU. Visual samples of some of the datasets are given in Figures \ref{fig=Tracking_R_ID_Samples} and \ref{fig=Action Recognition_Samples}. These datasets have been pivotal for benchmarking algorithms designed to address challenges such as occlusion handling, identity Re-ID across views, and HAR in multi-view systems. Moreover, they allow for the testing of algorithmic robustness in dynamic and cluttered environments, where variability in lighting, viewpoints, and motion patterns pose significant challenges. The continuous improvement of these datasets, including their annotation quality and diversity, directly contributes to the advancement of the field. As a result, they not only support the evaluation of existing methods but also inspire the development of new algorithms tailored to the evolving needs of multi-camera systems.

\renewcommand{\arraystretch}{1.33}
\sloppy
\begin{table*}[htbp]
\centering
\scriptsize
\caption{Summary of multi-camera multi-view multi-object tracking datasets.}
\label{table:datasets_tracking}
\begin{tabularx}{\textwidth}{@{}
    C{0.5cm}   % #
    L{2.5cm}   % Dataset
    C{1.1cm}   % Year
    C{1.1cm}   % Cameras
    C{1.69cm}   % Videos/Frames
    C{1.8cm}   % Environment
    J{6.9cm}   % Description (justified, no indent)
@{}}
\toprule
\textbf{\#} & \textbf{Dataset} & \textbf{Year} & \textbf{Cameras} & \textbf{Videos/Frames} & \textbf{Environment} & \textbf{Description} \\
\midrule
1  & \textbf{DukeMTMC}~\cite{ristani2016performance} & 2016 & 8   & 8     & Outdoor        & Contains over 2M annotated frames from eight synchronized campus cameras; widely used for multi-camera tracking.\\ \hline
2  & \textbf{EPFL-RLC}~\cite{chavdarova2017deep} & 2017 & 3   & 24K   & Indoor         & Pedestrian dataset recorded with three static overlapping cameras at 60 FPS; includes synchronized frames with occupancy annotations.\\ \hline
3  & \textbf{ToCaDa}~\cite{malon2018toulouse} & 2018 & 7   & 25    & Outdoor        & 25 synchronized cameras with annotated bounding boxes and synchronized audio tracks.\\ \hline
4  & \textbf{WILDTrack}~\cite{chavdarovawildtrack} & 2018 & 7   & 7     & Outdoor        & Multi-camera pedestrian tracking dataset with seven synchronized outdoor views; provides accurate 3D and 2D annotations.\\ \hline
5  & \textbf{MDOT}~\cite{zhu2020multi} & 2020& 5   & 155   & Outdoor        & Multi-drone single-object tracking dataset with 155 synchronized drone clips (113K frames with two drones, 145K with three).\\ \hline
6  & \textbf{GMTD}~\cite{wu2020visual} & 2020& 2/3 & 23    & Indoor/Outdoor & Proposed for evaluating tracker robustness with moving/uncalibrated cameras; frame-level annotation across diverse environments.\\ \hline
7  & \textbf{MultiviewX}~\cite{hou2020multiview} & 2020 & 6   & 6     & Outdoor        & Synthetic multiview detection dataset with six synchronized cameras, high-resolution views, and accurate BEV annotations.\\ \hline
8  & \textbf{MMPTrack}~\cite{han2023mmptrack} & 2021 & 4/6 & 2.98M & Indoor         & Large-scale MVMC dataset with 23 synchronized RGB+depth cameras across five scenes; auto-annotated and manually refined.\\ \hline
9  & \textbf{DIVOTrack}~\cite{hao2024divotrack} & 2024 & 3   & 75    & Outdoor        & Real-world cross-view dataset with ten scenarios; dense annotations across 953 cross-view tracks.\\ \hline
10 & \textbf{TeamTrack}~\cite{scott2024teamtrack} & 2024 & 2   & 6     & Indoor/Outdoor & Full-pitch tracking dataset in team sports using fisheye and drone top views.\\ \hline
11 & \textbf{MTMMC}~\cite{woo2024mtmmc} & 2024 & 16  & 16    & Indoor/Outdoor & Features 16 synchronized RGB-thermal cameras across campus and factory environments, supporting multimodal MTMC tracking.\\ \hline
12 & \textbf{MVTrack}~\cite{xu2025mitracker} & 2025 & 3/4 & 260   & Indoor/Outdoor & Designed for evaluating multi-view tracking; 260 synchronized RGB-D videos across 3–4 views with challenging scenarios.\\
\bottomrule
\end{tabularx}
\end{table*}
\subsection{MVMC Multi-Object Tracking Datasets}
\begin{figure}[!t]
    \centering
\includegraphics[width=1\linewidth, trim=11pt 11pt 10pt 11pt, clip]{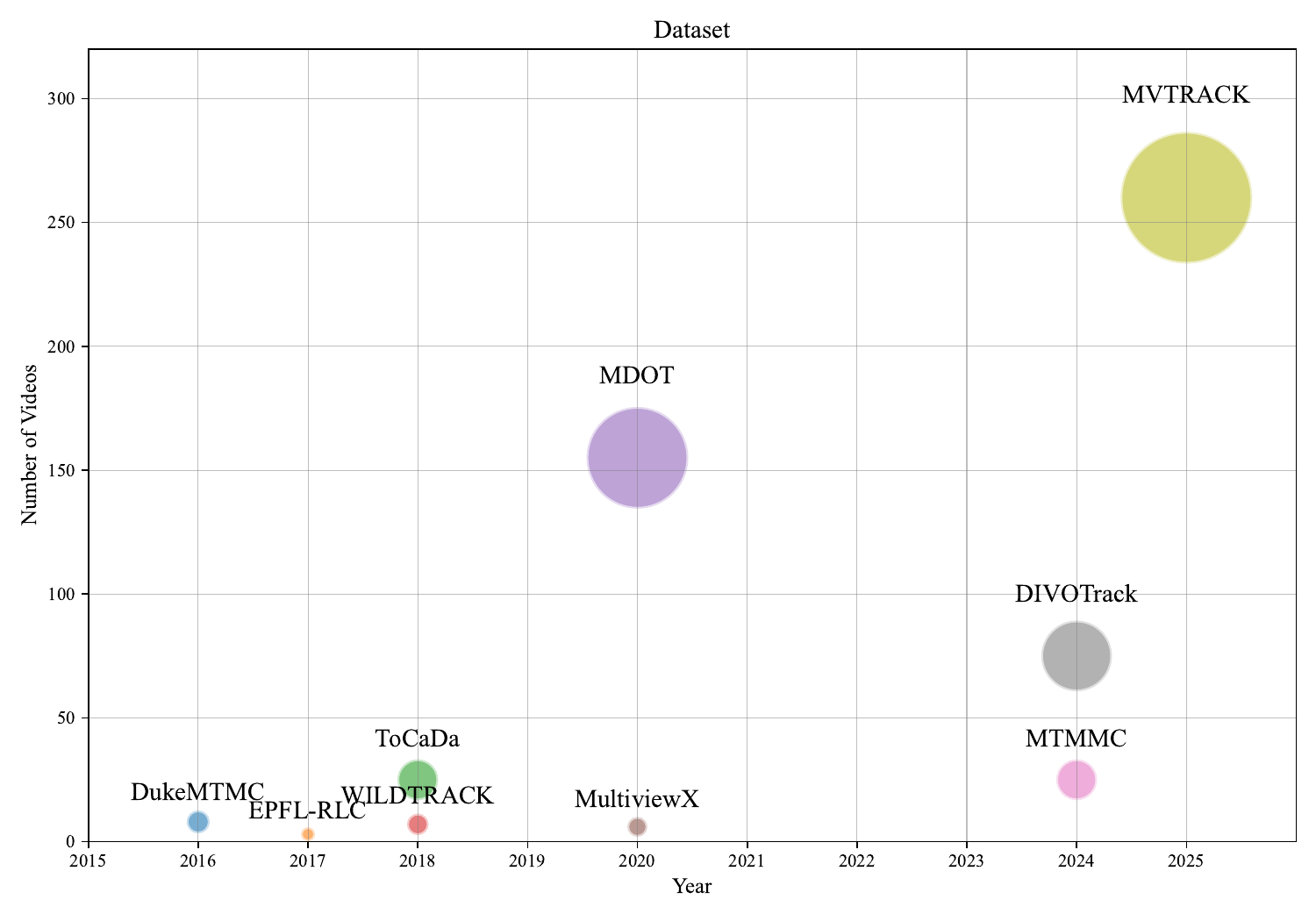}
    \caption{Number of videos across MVMC multi-object tracking datasets.}
    \label{Fig-Tracking_Datasets}
\end{figure}

MVMC-MOT is a critical component of CVS, where algorithms must address challenges such as occlusion, viewpoint variations, and the simultaneous tracking of multiple entities across multiple cameras. We have compiled and summarized publicly available datasets up to 2025, overview is presented in Table~\ref{table:datasets_tracking} and Figure \ref{Fig-Tracking_Datasets} illustrates the number of videos across various MVMC tracking datasets, providing insights into the scale of these benchmarks for evaluating tracking systems. Details of each dataset are as follows:
\textbf{DukeMTMC (2016):}  
The DukeMTMC\cite{ristani2016performance} dataset is a cornerstone for evaluating multi-camera pedestrian tracking and Re-ID algorithms. It consists of over 2 million annotated frames captured from eight synchronized HD cameras on the Duke University campus. The dataset tracks over 2,000 unique pedestrian identities, providing detailed annotations such as bounding boxes and identity labels. The inclusion of camera calibration data also aids in trajectory estimation and cross-camera association.
\textbf{EPFL-RLC Multi-Camera (2017):}  
The EPFL-RLC\cite{chavdarova2017deep} dataset is recorded at the EPFL Rolex Learning Center, consisting of three static cameras that capture overlapping views. It includes approximately 8,000 frames per sequence, with annotations for occupancy across a grid. This dataset is commonly used for indoor pedestrian detection, tracking, and multi-view camera calibration.
\textbf{ToCaDa (2019):}  
The ToCaDa\cite{malon2018toulouse} dataset captures synchronized video and audio sequences from 25 cameras, recorded across two distinct scenes in a campus setting. It includes annotations for pedestrian and vehicle movements, supporting research in cross-view tracking, object association, and event detection. The multimodal annotations (including audio) offer opportunities for multimodal fusion and cross-view event detection. A popular dataset,
\textbf{WILDTrack (2018):}  
WILDTRACK\cite{chavdarovawildtrack} provides synchronized outdoor pedestrian tracking data from seven static cameras positioned in a crowded public square. It offers over 40,000 bounding box annotations for more than 300 individual pedestrians. This dataset is crucial for developing and benchmarking algorithms that handle multi-view tracking in cluttered and dynamic environments.
\textbf{MDOT (2020):} The MDOT\cite{zhu2020multi} dataset is focused on multi-drone tracking for single-object scenarios.
\definecolor{LightGray}{rgb}{0.9,0.9,0.9}

\renewcommand{\arraystretch}{1.33}
\sloppy
\begin{table*}[!t]
\centering
\scriptsize
\caption{Summary of multi-camera object Re-identification datasets.}
\label{table:datasets_reid}
\begin{tabularx}{\textwidth}{@{}
    C{0.5cm}   % #
    L{2.5cm}   % Dataset
    C{0.7cm}   % Year
    C{0.6cm}   % Cameras
    C{1.9cm}   % Videos/Frames
    C{1.4cm}   % Environment
    J{8.0cm}   % Description
@{}}
\toprule
\textbf{\#} & \textbf{Dataset} & \textbf{Year} & \textbf{Cameras} & \textbf{Videos/Frames} & \textbf{Environment} & \textbf{Description} \\
\midrule
1  & \textbf{CAMPUS\cite{xu2016multi}} & 2016 & 4   & 83K   & Outdoor & A multi-view dataset with dense crowds and occlusions, captured by 3–4 synchronized HD cameras with fully annotated trajectories for MOT. \\ \hline
2  & \textbf{MARS\cite{zheng2016mars}} & 2016 & 6   & 20K+  & Outdoor & A video-based person Re-ID dataset, recorded across six synchronized cameras. It includes 3,248 distractor identities to simulate real-world conditions. \\ \hline
3  & \textbf{PRW\cite{zheng2017person}} & 2017 & 6   & 11K+  & Outdoor & A real-world dataset for pedestrian detection and re-identification in outdoor campus settings, with 34,304 bounding boxes across 932 individuals. \\ \hline
4  & \textbf{DukeMTMC-Reid\cite{zheng2017unlabeled}} & 2017 & 8   & 34K+  & Outdoor & An extension of the DukeMTMC dataset, specifically designed for Re-ID, featuring 1,812 pedestrian identities captured across eight synchronized cameras. \\ \hline
5  & \textbf{RPIfield\cite{zheng2018rpifield}} & 2018 & 12  & 12    & Outdoor & A large-scale outdoor Re-ID benchmark with approximately 4K distractor identities. It includes 30 hours of video captured from 12 synchronized cameras. \\ \hline
6  & \textbf{LPW\cite{song2018region}} & 2018 & 2/4 & 590K  & Outdoor & A real-world multi-camera Re-ID dataset with 7K+ pedestrian tracklets and 2K+ identities captured across three busy outdoor scenes. \\ \hline
7  & \textbf{CityFlow\cite{tang2019cityflow}} & 2019 & 40  & 40    & Outdoor & A city-scale benchmark for vehicle tracking and Re-ID, collected from 10 intersections. It contains 3.25 hours of HD video with 666 identities. \\ \hline
8  & \textbf{MEVID\cite{davila2023mevid}} & 2023 & 2/6 & 10.46M+ & Indoor/Outdoor & A large-scale video-based Re-ID dataset featuring 158 individuals, 8K tracklets, and 33 cameras across 17 locations. It supports cross-view identity matching. \\
\bottomrule
\end{tabularx}
\end{table*}
It includes 155 synchronized video sequences captured using two or three drones, totaling 259,793 frames. The dataset includes annotations for challenging tracking conditions, such as occlusion, camera motion, and illumination variation. It is used for research in robust multi-drone tracking, trajectory estimation, and multi-view fusion.
\textbf{GMTD (2020):}  
The GMTD\cite{wu2020visual} dataset consists of 23 synchronized sequences captured at 1080p resolution from two or three uncalibrated cameras. It covers diverse objects, including humans, animals, and inanimate objects, across both indoor and outdoor environments. The dataset’s detailed frame-level annotations, including object identities and occlusion handling, make it valuable for evaluating trajectory prediction and generic object tracking algorithms.  Similarly to the WILDTrack\cite{chavdarovawildtrack}, \textbf{MultiviewX (2020):}  
The MultiviewX\cite{hou2020multiview} dataset is a synthetic benchmark used for evaluating multi-view pedestrian detection algorithms. Captured using six static cameras in a 16x25m² area, the dataset includes annotated frames with high precision, including calibrated ground-plane projections. The dataset supports testing under severe occlusion and crowded environments.
\textbf{MMPTrack (2021):}  
The MMPTrack\cite{han2023mmptrack} dataset is one of the largest and most comprehensive multi-camera tracking benchmarks. Captured using 23 well-calibrated RGB cameras, it spans five different environments and includes approximately 9.6 hours of video. The dataset includes dense annotations for full-body tracking across various indoor scenarios and is widely used for developing and benchmarking multi-camera tracking algorithms.
\textbf{DIVOTrack (2024):}  
The DIVOTrack\cite{hao2024divotrack}dataset is collected for evaluating cross-view multi-object pedestrian tracking in open-world environments. It includes 953 cross-view tracks from 15 diverse scenarios captured with three moving cameras. The dataset provides dense annotations across 953 tracks and 81,000 frames, making it a valuable resource for evaluating algorithms for cross-view identity matching and robust tracking.
\textbf{TeamTrack (2024):}  
The TeamTrack\cite{scott2024teamtrack} dataset focuses on multi-sport tracking, recorded from dual viewpoints (drone top-view and fisheye side-view) across multiple sports including soccer, basketball, and handball. It includes around 155 minutes of synchronized footage with 4.37 million annotated frames. This dataset challenges algorithms in scenarios with frequent occlusions, similar visual appearances, and dynamic motion patterns.

\renewcommand{\arraystretch}{1.4}

\sloppy
\begin{table*}[!t]
\centering
\scriptsize
\caption{Summary of multi-camera multi-view action understanding datasets.}
\label{table:datasets_action}
\begin{tabularx}{\textwidth}{@{}
    C{0.5cm}   % #
    L{2.4cm}   % Dataset
    C{0.9cm}   % Year
    C{0.7cm}   % Cameras
    C{1.4cm}   % Videos/Frames
    C{1.7cm}   % Environment
    J{8cm}   % Description
@{}}
\toprule
\textbf{\#} & \textbf{Dataset} & \textbf{Year} & \textbf{Cameras} & \textbf{Videos/Frames} & \textbf{Environment} & \textbf{Description} \\
\midrule
1  & \textbf{MCAD\cite{li2017multi}} & 2017 & 5   & 14,298   & Indoor/Outdoor & A MVMC AR dataset, capturing 18 distinct action categories from 20 individuals using five synchronized CCTV cameras, ideal for cross-view AR. \\ \hline
2  & \textbf{PKU-MMD\cite{liu2017pku}} & 2017 & 3   & 21,545  & Indoor & A multi-modal dataset for human AR. It includes RGB, depth, infrared, and 3D skeleton data across 51 action categories. \\ \hline
3  & \textbf{MOD20\cite{perera2020multiviewpoint}} & 2020 & 3/4 & 2,324   & Outdoor & A multi-view outdoor AR dataset recorded from both drone and ground-level cameras, covering 23 AU categories such as running, cycling, and rock climbing. \\ \hline
4  & \textbf{NTU RGB+D 120\cite{liu2019ntu}} & 2019 & 3   & 114K+   & Indoor & One of the largest multi-modal (RGB, depth, infrared, and skeleton) datasets for 3D human AR, featuring 120 classes recorded using 155 camera viewpoints. \\ \hline
5  & \textbf{RoCoG\cite{de2020vision}} & 2020 & 3   & 1,574   & Outdoor & Provides real and synthetic multi-view videos for training and evaluating human-robot interaction recognition systems. \\ \hline
6  & \textbf{IKEA Assembly\cite{ben2021ikea}} & 2021 & 3   & 371   & Indoor & A dataset for fine-grained AR during furniture assembly, featuring 33 atomic actions annotated with human pose data and object segmentation. \\ \hline
7  & \textbf{MM-OFFICE\cite{yasuda2022multi}} & 2022& 4   & 880   & Indoor & A multi-modal dataset for AR in office scenes, recorded with four GoPro cameras and eight microphones. It includes weakly and strongly labeled annotations. \\ \hline
8  & \textbf{CHAD\cite{danesh2023chad}} & 2023 & 4   & 412   & Outdoor & A MVMC dataset for anomaly detection and AR, focused on scenarios in a commercial parking lot, with annotated normal and anomalous behaviors. \\ \hline
9  & \textbf{RoCoG\_V2\cite{reddy2023synthetic}} & 2023 & 2   & 482   & Outdoor & Designed for synthetic-to-real gesture recognition, containing real ground/air videos and large-scale synthetic counterparts for seven robot control gestures. \\ \hline
10 & \textbf{MM-STORE\cite{nguyen2025multisensor}} & 2025 & 6   & 2,970   & Indoor & A multi-modal AR dataset recorded in a convenience store, with RGB and audio data, annotated for 18 event categories, designed for weakly supervised learning. \\
\bottomrule
\end{tabularx}
\end{table*}

\textbf{MTMMC (2024):}  
The MTMMC\cite{woo2024mtmmc} dataset includes 16 synchronized RGB and thermal connected cameras deployed in diverse indoor and outdoor environments. It comprises more than 3 million frames and 3,669 unique pedestrian identities. This dataset is useful for research on multi-modal fusion strategies, cross-camera identity association, and tracking performance under complex real-world conditions.
\textbf{MVTRACK (2025):}  
The MVTRACK\cite{xu2025mitracker} dataset is collected for evaluating multi-view object tracking. It includes 234,000 frames across 260 video sequences captured by 3–4 synchronized static cameras. It provides detailed annotations on objects under various conditions, such as background clutter, occlusion, and low resolution, making it suitable for benchmarking tracking algorithms. Additionally, the dataset’s diverse scenarios make it a valuable resource for testing the robustness and adaptability of tracking models in real-world applications.
\subsection{Multi Camera Re-Identification Datasets}
Object Re-ID plays a crucial role in CVS, surveillance, and monitoring applications, as it enables the consistent matching of individuals across non-overlapping camera views within multi-camera CVS. This task is particularly challenging due to significant variations in illumination, pose, viewpoint, occlusion, and background clutter, all of which can drastically impact appearance consistency across cameras. Over the years, a variety of benchmark datasets have been introduced to facilitate the development and evaluation of Re-ID algorithms under these challenging real-world conditions. We compile and summarize the publicly available datasets up to 2025, which provide valuable benchmarks for assessing algorithmic performance, as presented in Table~\ref{table:datasets_reid}. Details of each dataset are as follows:

\textbf{Campus (2016):}  
The Campus\cite{xu2016multi} dataset is a multi-view benchmark with four sequences (Garden 1, Garden 2, Auditorium, Parking Lot) recorded by 3–4 synchronized HD cameras. It features around 1,525 objects with dense crowds, frequent occlusions, and unique ID annotations across views for evaluating multi-camera multi-object Re-ID.
\textbf{MARS (2016):}  
The MARS\cite{zheng2016mars} dataset is a video-based Re-ID benchmark consisting of 1,261 unique pedestrians, recorded across six synchronized cameras. It is used to evaluate video-based matching algorithms that exploit temporal and spatial cues for identity matching.
\textbf{PRW (2017):}  
The PRW\cite{zheng2017person} dataset contains 34,304 bounding boxes for 932 pedestrians, recorded using six synchronized cameras. It is collected to evaluate both pedestrian detection and Re-ID algorithms in real-world outdoor settings, where varying lighting, pose, and occlusion present significant challenges.
\begin{figure*}[!b]
    \centering
    \includegraphics[width=1\linewidth, trim=20pt 100pt 20pt 140pt, clip]{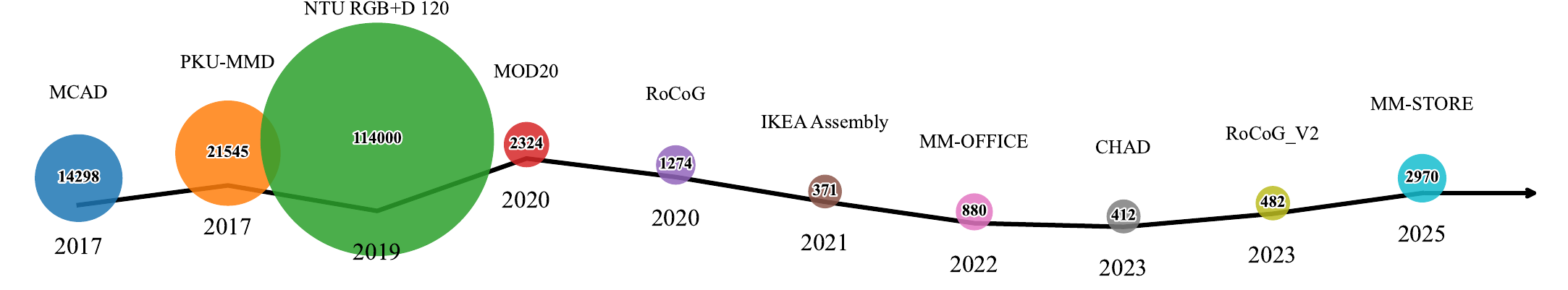}
    \caption{Videos across MVMC action understanding datasets. }
    \label{fig=AU-Datasets}
\end{figure*}
\textbf{DukeMTMC-Reid (2017):}  
An extension of DukeMTMC\cite{ristani2016performance}, the DukeMTMC-Reid\cite{zheng2017unlabeled} dataset provides 36,441 manually annotated cropped pedestrian images across eight synchronized HD cameras. It contains 1,812 unique pedestrian identities and is widely used to benchmark algorithms that aim to match identities across different views under challenging conditions.
\textbf{RPIfield (2018):}  
The RPIfield\cite{zheng2017unlabeled} dataset includes 30 hours of multi-view video captured by 12 synchronized outdoor cameras, tracking 112 known actors and approximately 4,000 distractor identities. It is used to study identity matching and temporal persistence in multi-camera surveillance environments.
\textbf{LPW (2018):}  
The LPW\cite{song2018region} dataset features 7,694 tracklets for 2,731 unique identities, recorded across three busy scenes covered by 2–4 synchronized cameras. It includes pedestrians performing various actions like walking, running, and cycling, and is used to benchmark algorithms for pedestrian tracking and Re-ID under realistic conditions.  \textbf{CityFlow (2019):}  
Other than person Re-ID datasets, the CityFlow\cite{tang2019cityflow} dataset is a city-scale benchmark for multi-target multi-camera vehicle tracking and Re-ID. It includes 3.25 hours of synchronized HD video from 40 cameras across 10 intersections, with 666 vehicle identities and 229K annotated bounding boxes. Camera geometry and calibration are provided to support spatio-temporal reasoning for MTMC tracking and vehicle Re-ID.
\textbf{MEVID (2023):}  
MEVID\cite{davila2023mevid} is a large-scale dataset containing 8,092 tracklets of 158 individuals across 33 cameras. It includes both indoor and outdoor scenes and supports research on domain adaptation, multi-view identity matching, and robust Re-ID under varying environmental conditions.
\subsection{MVMC Action Understanding Datasets}
AU across multiple cameras and views is a fundamental task for understanding human activities in diverse applications of CVS after tracking and Re-ID, including surveillance and behavior analysis. 
Unlike single-view AR, the MVMC setting introduces additional challenges such as viewpoint changes, occlusion, varying illumination, and synchronization across cameras, all of which significantly affect recognition performance. To support the development and evaluation of robust algorithms, several benchmark datasets have been proposed that capture human activities across multiple viewpoints under realistic conditions. We compile and summarize the publicly available MVMC AU datasets up to 2025, which serve as key benchmarks in this field, as presented in Table~\ref{table:datasets_action}. Furthermore, the distribution of videos across these datasets is illustrated in Figure~\ref{fig=AU-Datasets}.
\textbf{MCAD (2017):}  
The MCAD\cite{li2017multi} dataset consists of 14,298 action samples spanning 18 action categories, recorded across five synchronized CCTV cameras. It is widely used for evaluating MVAR algorithms, especially in surveillance environments.
\textbf{PKU-MMD (2017):}  
The PKU-MMD\cite{liu2017pku} dataset includes over 5.4 million frames across 51 action categories, recorded using three synchronized Kinect cameras. It supports research in multi-modal human AR, including skeleton-based analysis and continuous action detection.
\textbf{MOD20 (2020):}  
MOD20\cite{liu2017pku} consists of 2,324 action clips recorded from both drone and ground-level cameras. It includes 20 action categories, making it suitable for benchmarking multi-view action recognition (MVAR) in outdoor environments.
\textbf{NTU RGB+D 120 (2019):}  
One of the largest and most comprehensive AR datasets, NTU RGB+D 120\cite{liu2019ntu} contains over 114,000 frames across 155 viewpoints, capturing 120 action classes. It supports both skeleton-based and multimodal AR under diverse conditions.
\textbf{IKEA Assembly (2021):}  
The IKEA Assembly\cite{ben2021ikea} dataset is designed for detailed AR during furniture assembly. It includes 371 video samples across three camera views, with annotations for 33 atomic actions, human pose estimation, and object tracking.
\textbf{MM-OFFICE (2022):}  
MM-OFFICE\cite{yasuda2022multi} includes 880 video clips recorded in office environments using four RGB cameras and eight microphones. It is used for weakly supervised AR and multi-modal sensor fusion.
\textbf{CHAD (2023):}  
CHAD\cite{danesh2023chad} focuses on anomaly detection in a commercial parking lot, providing 1.15 million frames from four synchronized cameras. It includes both normal and anomalous frame annotations.
\textbf{RoCoG (2023):}  
\textbf RoCoG\cite{reddy2023synthetic, de2020vision} have two versions that are collected for robot gesture recognition, containing real and synthetic clips captured from 360° camera orientations. It supports research in human-robot interaction and gesture recognition.
\textbf{MM-STORE (2025):}  
\textbf{MM-STORE} contains 2,970 video clips recorded across six RGB cameras and omnidirectional microphones in a commercial store. It supports research in multi-modal AR and event tagging.

\renewcommand{\arraystretch}{0.83}
\sloppy
\begin{table*}[!t]
\centering
\tiny
\caption{Summary of MVMC Tracking Methods.}
\label{table:mvmc_tracking}

\resizebox{\textwidth}{!}{%
\begin{tabularx}{1.07\textwidth}{@{}
 C{0.3cm}   % #
    L{2.0cm}   % Method
    L{2cm}   % Venue/Year (merged)
    L{2.5cm}   % Feature Extractor
    L{2.4cm}   % M-Tracker
    L{3.0cm}     % Dataset
    L{2.9cm}   % Evaluation Metric
    C{1.5cm}   % Code
@{}}
\toprule
\textbf{\#} & \textbf{Method} & \textbf{Venue,Year} & \textbf{Feature Extractor} & \textbf{M-Tracker} & \textbf{Dataset} & \textbf{Evaluation Metric} & \textbf{ Source Code} \\
\midrule
1  & MVP-Track \cite{xu2016multi} & CVPR,16& DCNN & Hierarchical Trajectory & EPFL, PETS 2009, CAMPUS & MODA, MOTA, MT, ML, IDSW & -- \\ \midrule
2  & MTMC-Track \cite{zhang1712multi} & arXiv,17& AlignedReID & Hungarian Matching & DukeMTMC & IDF1, IDP, IDR & -- \\ \midrule
3  & Cross-View Track \cite{xu2017cross} & AAAI,17& DCNN & ST-APG & CAMPUS, PPL-DA & TA, TP, MT, ML, IDSW & -- \\ \midrule
4  & MCMT-Track \cite{wen2017multi} & IJCV,17& -- & STV hypergraph & PETS 2009 & MOTA, MOTP, MT, ML, IDSW & -- \\ \midrule
5  & DOR-MCMT Det \cite{baque2017deep} & ICCV,17& VGG-16 & Deep Occlusion & ETHZ, EPFL, PETS & MODA, MODP & \href{https://github.com/pierrebaque/DeepOcclusion?tab=readme-ov-file}{Link} \\ \midrule
6  & MCP-Det \cite{martin2018enhancing} & Sensors,18& DPM, YOLO9000 & Threshold Based & PETS S3MF1, EPFL-RLC & Precision, F-score & -- \\ \midrule
7  & WILDTRACK-PD \cite{chavdarova2018wildtrack} & CVPR,18& GoogleNet, ResNet-18 & non-Markovian, KSP, ptrack & WILDTRACK & Acc, AUC, MOTA, MOTP & -- \\ \midrule
8  & MV-Multi-Target \cite{he2019efficient} & IEEE Sensors,19& HOG, YOLOv3 & DeepSORT, MTIC, ICF & EPFL & IDF1, MOTA, MOTP, MODA & -- \\ \midrule
9  & GM-Track \cite{wu2019generic} & arXiv,19& ResNet-50 & GMT & GMTD & Average and Weighted Acc & -- \\ \midrule
10 & MV-3D Track \cite{ong2020bayesian} & TPAMI,20& YOLOv3, Faster R-CNN & Multi-View 3D Tracking & WildTracks, Curtin Multi-Camera & ML, IDSW, MOTA, MOTP, OSPA & -- \\ \midrule
11 & TPN-S \cite{wu2020visual} & TIP,20& ResNet-50 & GMT & GMTD, CAMPUS & Acc & -- \\ \midrule
12 & Tracklet-to-Target \cite{he2020multi} & TIP,20& ResNet-18, HOG & DHash trac, TRACTA & CAMPUS, EPFL, MCT & MOTA, MOTP, MODA, MODP & \href{https://github.com/GehenHe/TRACTA}{Link} \\ \midrule
13 & CoMOTE \cite{sun2020collaborative} & IEEE HPCC,20& -- & Collaborative Multi-Object & WildTrack & CLMOT, MOTA, MOTP, IDs, Acc & -- \\ \midrule
14 & Vehicle Track \cite{shim2021multi} & CVPR,21& ResNeXt-50, DLA-34 & Hungarian algorithm & CityFlow & IDF1, IDP, IDR & -- \\ \midrule
15 & CIR Track \cite{li2021multi} & CVPR,21& ResNet-101 with IBN-Net-a & TrackletNet & AI City Challenge 21& IDF1, IDTP, IDFP, IDFN, mAP & -- \\ \midrule
16 & UOT+pos \cite{le2021unbalanced} & ICPR,21& ResNet-50, OpenPose, R-CNN & Unbalanced Optimal & PETS2009, EPFL & IDF1, IDP, IDs, MOTA, MOTP & -- \\ \midrule
17 & DyGLIP \cite{quach2021dyglip} & CVPR,21& Omni-Scale Network & DyGLIP & CityFlow, CAMPUS, MCT, EPFL & MT, ML, MCTA, MOTA, MOTP, IDSW & \href{https://github.com/uark-cviu/DyGLIP}{Link} \\ \midrule
18 & TSCT+TA+META \cite{hsu2021multi} & TIP,21& ResNet-50 & TSCT, TCLM & CityFlow & MOTA, MOTP, IDF1, Recall, MT & -- \\ \midrule
19 & SN-Tracking \cite{cioppa2022soccernet} & CVPR,22& DLA-34, PCB Pyramid & DeepSORT, FairMOT & SoccerNet Tracking & HOTA, MOTA & \href{https://github.com/SoccerNet/sn-tracking}{Link} \\ \midrule
20 & MUTR3D \cite{zhang2022mutr3d} & CVPR,22& ResNet-101 & MUTR3D & nuScenes & MOTA, IDS & -- \\ \midrule
21 & DMMA \cite{yang2022distributed} & Scientific Reports,22& CN+MobileNet & DMMA & MOT16 & MOTA & -- \\ \midrule
22 & SCLM+GAE \cite{hsu2022multi} & WACV,22& ResNet-50 & SCLM+GAE & CityFlow-19, CityFlow-20 & IDF1, IDP, IDR & -- \\ \midrule
23 & MTrack \cite{yu2022towards} & CVPR,22& DLA-34 & FairMOT, MTrack & MOT15, MOT16, MOT17, MOT20 & IDF1, MOTA, IDSW & -- \\ \midrule
24 & MCT \cite{guo2022multi} & Neurocom,22& CNN & DeepSORT, TraDes & EPFL & HOTA, DeTA, LocA, IDF1 & -- \\ \midrule
25 & Indoor MOT \cite{jang2022lightweight} & Sensors,22& KLT & DTW, FastMOT & -- & F1 Score & -- \\ \midrule
26 & BG-Reranking \cite{yang2022box} & CVPR,22& HRNet, ResNeXt101 & ICA, SCMT, DeepSORT & CityFlowV2 & IDF1, IDP, IDR, mAP & -- \\ \midrule
27 & LMGP \cite{nguyen2022lmgp} & CVPR,22& DG-Net & LMGP & WILDTRACK, CAMPUS, PETS-09 & IDF1, MOTA, MOTP, MT, ML, IDSW & -- \\ \midrule
28 & ReST \cite{cheng2023rest} & ICCV,23& OSNet & ReST & WILDTRACK, CAMPUS, PETS-09 & IDF1, MOTA, MOTP, MT, ML & \href{https://github.com/chengche6230/ReST?tab=readme-ov-file}{Link} \\ \midrule
29 & MC-MHT \cite{wang2023blockchain} & TII,23& CNN, YOLO & MC-MHT, MCTChain & CAMPUS & MOTA, MOTP, MT, ML & -- \\ \midrule
30 & MH-Track \cite{yousefi2023tracking} & ASP,23& CNN, Kalman Filter, GA & MSHM & PETS 2009 & MOTA, MOTP, MT, ML, FM, IDSW & -- \\ \midrule
31 & DIVOTrack \cite{hao2024divotrack} & IJCV,23& DLA-34 & CrossMOT & DIVOTrack & CVIDF1, CVMA, MOTA & \href{https://github.com/shengyuhao/DIVOTrack}{Link} \\ \midrule
32 & MVFlow+MUSSP \cite{engilberge2023multi} & WACV,23& ResNet-34, KSP, MuSSP & DeepSort & PETS2009, WILDTRACK & MODA, MOTA, MOTP, IDF1 & \href{https://github.com/cvlab-epfl/MVFlow}{Link} \\ \midrule
33 & CMMT \cite{yang2024unified} & CVM,24& YOLOX, StrongReID & SORT+PDNC+CMMT & WILDTRACK, MMPTRACK & IDF1, MOTA, MT, ML, PCP & -- \\ \midrule
34 & AMOT \cite{fang2024coordinate} & Multimedia-S,24& YOLOv4 & MARL & 3D Soccer Court Environment & Coverage rate & -- \\ \midrule
35 & TeamTrack \cite{scott2024teamtrack} & CVPR,24& YOLOv8 & ByteTrack, BoT-SORT & TeamTracK & HOTA, MOTA, IDF1 &\href{https://github.com/AtomScott/TeamTrack}{Link} \\ \midrule
36 & MO-MCT \cite{zaman2024robust} & MTA,24& ResNet-152 & DeepSORT, Kalman filter & AI City & IDF1, Precision, Recall & -- \\ \midrule
37 & ADA-Track++ \cite{ding2024ada} & CVPR,24& VoVNet-99, ViT-Large & ADA-Track++ & NUScenes & AMOTA, AMOTP, IDSW & \href{https://github.com/dsx0511/ADA-Track?utm_source}{Link} \\ \midrule
38 & CSR+EUP \cite{kim2024cluster} & CVPR,24& ResNet50-IBN & BoT-Sort & Challenge Track 1, CrowdPose & HOTA, DetA, AssA, LocA & -- \\ \midrule
39 & GMT \cite{fan2024gmt} & arXiv,24& DLA-34, CenterNet & GMT & VisionTrack & MOTA, IDF1, HOTA, IDSW & -- \\ \midrule
40 & TrackTacular \cite{teepe2024lifting} & CVPR,24& BEVFormer & One-shot, CenterNet & MultiviewX, WILDTRACK & IDF1, MOTA, MOTP, MT, ML & \href{https://github.com/tteepe/TrackTacular}{Link} \\ \midrule
41 & OCMCTrack \cite{specker2024ocmctrack} & CVPR,24& SOLIDER & ConfTrack, OCMCTrack & AICity'24 & mAP, HOTA, DetA, AssA, LocA & -- \\ \midrule
42 & MTMC-CIS \cite{cherdchusakulchai2024online} & CVPR,24& OSNet & ByteTrack & AICity'24 & HOTA, DetA, AssA, LocA & -- \\ \midrule
43 & RIIPS \cite{yoshida2024overlap} & CVPR,24& OSNet-AIN & Hierarchical clustering-based & AICity'24 & HOTA, DetA, AssA, LocA & -- \\ \midrule
44 & RockTrack \cite{li2024rocktrack} & arXiv,24& --- & RockTrack & nuScenes & AMOTA, AMOTP, IDS & -- \\ \midrule
45 & STMC \cite{herzog2024spatial} & arXiv,24& LightMBN, ResNet-101 & Graph-based & CityFlow, Synthehicle & IDF1, MOTA & \href{https://github.com/fubel/stmc}{Link} \\ \midrule
46 & ASTM-Net \cite{zhang2024video} & TCSVT,24& ResNeXt-101, SE-ResNet-101 & ASTM-Net & HST, CityFlow, UA-DETRAC & IDF1, IDP, IDR, MOTA, IDSW & -- \\ \midrule
47 & CLM-Track \cite{lin2024city} & PAMI,24& ResNeXt101-IBN-a & DeepSORT, Greedy Matching & CityFlow V2 & IDF1, IDP, IDR, Precision, Recall & -- \\ \midrule
48 & TrakAthlete4D \cite{agarwal2024trakathlete4d} & BMVC,24& YOLOX-L Pretrained & TrakAthlete4D & GMVD & MODA, MODP, Precision, Recall & -- \\ \midrule
49 & FastTrackTr \cite{liao2024fasttracktr} & arXiv,24& RT-DETR, ResNet-50 & FastTrackTr & DanceTrack, MOT17, VisDrone2019 & HOTA, DetA, IDF1, MOTA & -- \\ \midrule
50 & MVTr \cite{yang2024end} & CVIU,24& ResNet-18, Transformer & MVTr, Hungarian & WildTrack, Multi-ViewX & IDF1, MOTA, MOTP, MT, ML, IDSW & -- \\ \midrule
51 & EarlyBird \cite{teepe2024earlybird} & WACV,24& ResNet-18, ResNet-50, Swin-T & EarlyBird & WildTrack, MultiViewX & IDF1, MOTA, MOTP, MODA, MODP & \href{https://github.com/tteepe/EarlyBird}{Link} \\ \midrule
52 & Argus \cite{yi2024argus} & IEEE TMC,24& ResNet-101, YOLO-v5n & Conv-Track, Spatula-Track & CityFlowV2, CAMPUS, MMPTRACK & MOTP, MOTA & -- \\ \midrule
53 & Multi-Cuts \cite{herzog2024spatial} & arXiv,24& ResNet-101, YOLOvX & Multi-Cuts & CityFlow, Synthehicle & IDF1, MOTA & \href{https://github.com/fubel/stmc}{Link} \\ \midrule
54 & MITracker \cite{xu2025mitracker} & CVPR,25& ViT & MITracker & MVTrack, GMTD & AUC, PNorm, Precision & \href{https://github.com/XuM007/MITracker}{Link} \\ \midrule
55 & All Day-MCMT \cite{fan2025all} & CVPR,25& DLA-34 & GMT & M3Track & MOTA, HOTA, MOTP, IDF1 & \href{https://github.com/QTRACKY/ADMCMT}{Link} \\ \midrule
56 & MCTR \cite{niculescu2025mctr} & WACV,25& DETR, YOLOX, ReST & MCTR & MMPTrak, AI City Challenge & HOTA, IDF1, MOTA & -- \\ \midrule
57 & TrackTrack \cite{shim2025focusing} & CVPR,25& YOLOvX-x, SBS50 & TrackTrack & MOT17-20, DanceTrack & CLEAR, IDF1, HOTA, AssA & -- \\ \midrule
58 & McByte \cite{stanczyk2025no} & CVPR,25& YOLOvX, SAM & McByte & SportsMoT, SoccerNet-Tracking & HOTA, IDF1, MOTA & -- \\ \midrule
59 & Fish-Track \cite{chai2025framework} & arXiv,25& YOLOv8 & ByteTrack & MOT17, Fish Data & HOTA, MOTA, AssA, DetA, IDF1 & -- \\ \midrule
60 & OMO-Track \cite{luo2025omnidirectional} & CVPR,25& ResNet-50, YOLOv11v & FlexiTrack & JRDB, QuadTrack & HOTA, MOTA, AssA, DetA, IDF1 & -- \\ \midrule
61 & CityTrac \cite{yu2025citytrac} & IoTJ,25& -- & CityTrac & CityFlow V2, Geolife & Track-R, F-Rate & -- \\ \midrule
62 & FusionTrack \cite{li2025fusiontrack} & arXiv,25& ResNet-50 & FusionTrack & MDMOT & MOTA, MOTP, IDF1, HOTA, IDS & -- \\ \midrule
63 & AAM-View Track \cite{alturki2025attention} & arXiv,25& Custom ResNet & Kalman Filter & WildTrack, MultiViewX & MOTA, MOTP, IDF1, MT, ML & -- \\ \midrule
64 & MCBLT \cite{wang2024mcblt} & arXiv,25& Transformer, GNN & Multi-Camera Multi-Object 3D & AICity'24, WildTrack & Rank-1, 5, Acc, mAP & -- \\ \midrule
65 & IV-Direction \cite{ali2025integrating} & ISPRS,25& ResNet-18, Custom CNN & Kalman Filter, DeepSORT & WildTrack & MODA, MODP, MOTA & -- \\ \midrule
66 & GCF-MTMCT \cite{hu2025enhanced} & RobCE,25& ResNet-18, ResNet-18 Decoder & Global Context Fusion Network & WildTrack & MODA, MODP, F1-score & -- \\ \midrule
67 & MVPOV \cite{alturki2025enhanced} & CVPR,25& ResNet, CenterNet & Multi-View Pedestrian Detection & WildTrack, MultiViewX & MODA, MODP & -- \\ \midrule
68 & UMPN \cite{engilberge2025unified} & arXiv,25& GNN, Greedy Algorithm & Unified Message Passing & WildTrack, MOT20, SCOUT & MOTA, MOTP, IDF1, HOTA & -- \\ \midrule
69 & MVL-Net \cite{zhang2025mvl} & IEEE TMM,25& ResNet-50 & MVL-Net & MV3DHumans & Precision, Recall & -- \\ \midrule
70 & MVTL-UAV \cite{wu2025multi} & Drones,25& DLA-34 & MVTL-UAV & DIVOTrack & CVMA, CVIDF1 & -- \\
\bottomrule
\end{tabularx}
}
\end{table*}

The datasets reviewed in this section are pivotal for advancing the field of MVMC-connected vision systems. Each dataset contributes to different aspects of multi-view analytics, from multi-object tracking to object Re-ID and AU. The MOT datasets, such as DukeMTMC and ToCaDa, are essential for evaluating algorithms that handle occlusion, tracking in dynamic environments, and multi-target tracking across different camera views. The Re-ID datasets, including DukeMTMC-Reid and MARS, enable the evaluation of identity matching algorithms that need to handle real-world challenges such as varying poses, lighting conditions, and occlusions. Finally, the AU datasets like NTU RGB+D 120 and PKU-MMD provide rich data for benchmarking AU algorithms across multiple viewpoints and modalities. Figures \ref{Fig-Tracking_Datasets} and \ref{fig=AU-Datasets} illustrate the comparative sizes and complexities of datasets across the multi-camera AU domain, helping to contextualize the scale of these benchmarks for future research. Furthermore, these datasets collectively propel research in multi-camera vision systems, fostering the development of more accurate, robust, and generalizable algorithms for MVMC multi-object tracking, Re-ID, and MVAU. As a result, they play a critical role in bridging the gap between theoretical advancements and practical applications in real-world environments.

\begin{figure*}[!t]
    \centering
    \includegraphics[width=1\linewidth]{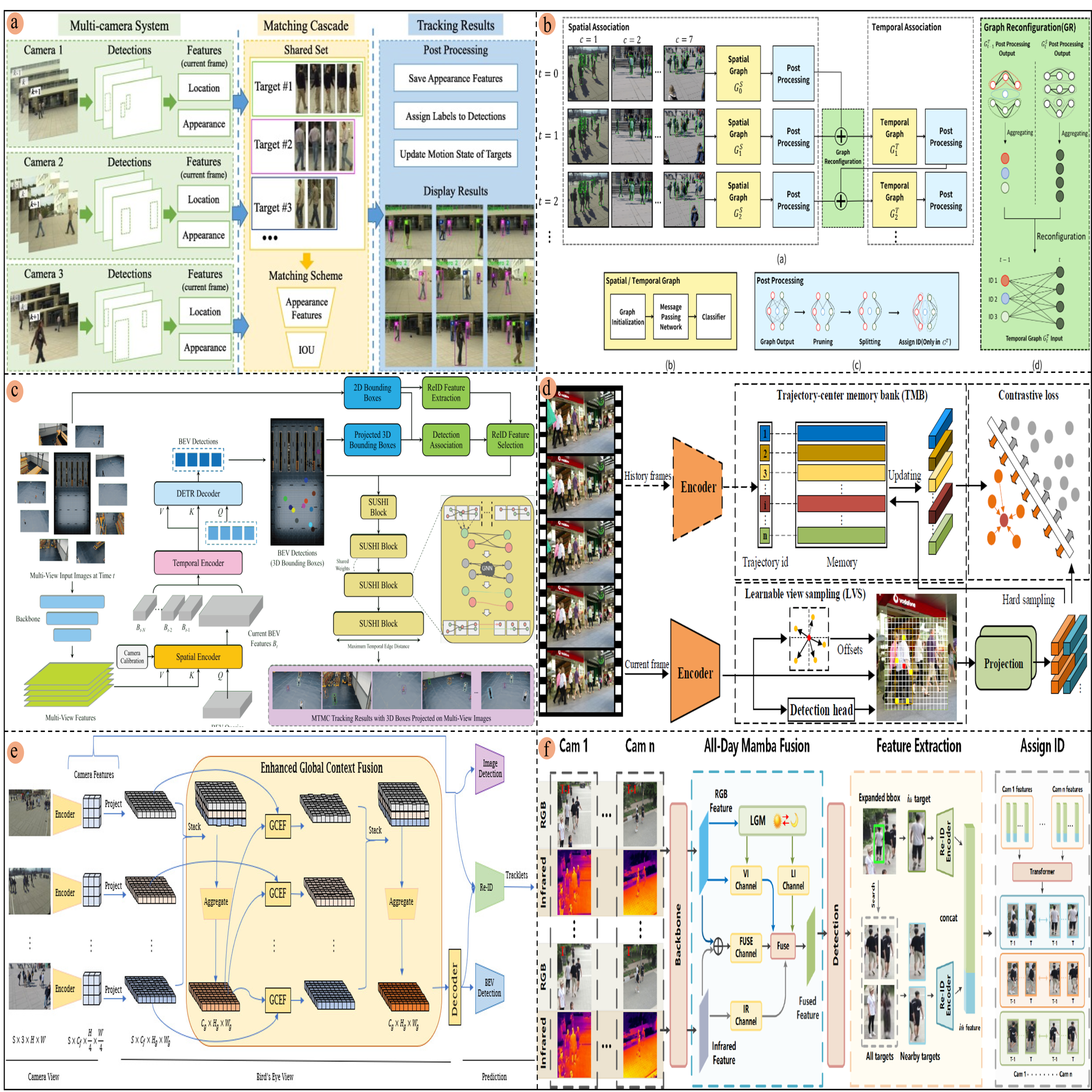}
    \caption{Visual representation of advanced Multi-View and Multi-Camera Tracking frameworks, showcasing different techniques for tracking multiple targets across multiple cameras, including methods for a) motion modeling \cite{guo2022multi}, b) spatial-temporal associations \cite{cheng2023rest}, c) 3D tracking \cite{wang2024mcblt}, d) discriminative learning \cite{yu2022towards}, e) global context fusion \cite{hu2025enhanced}, f) all-day tracking \cite{fan2025all}. Each framework addresses unique challenges in MVMC, multi-target tracking scenarios.}
    \label{fig=Tracking-Frameworks}
\end{figure*}
\section{PROPOSED TAXONOMY}
 To provide a clear and structured understanding of the diverse approaches in multi-camera CVS with multi-view analytics, we introduce a novel taxonomy that systematically organizes the existing methods. While several studies have categorized techniques in related areas, none have specifically addressed the full spectrum of approaches under the MVMC context. For instance, in \cite{zhang2024comprehensive,marvasti2021deep} focused on tracking methods across single-camera systems, and in \cite{sun2022human, stergiou2025time} examined AR. However, these efforts do not comprehensively cover the intricacies and diverse approaches of MVMC environments, where methods must account for synchronization, occlusion handling, and heterogeneous viewpoints. In this work, we aim to fill this gap by proposing a comprehensive taxonomy for MVMC systems. Our taxonomy is divided into four key dimensions: MVMC Tracking, MVMC Re-ID, MVMC AU, and MVMC integrated  approaches. Each of these categories addresses fundamental aspects of MVMC systems, offering insights into MVMC tracking and Re-ID challenges, AR techniques, and integrated methods that combine multiple facets of vision systems. Additionally, we examine how these dimensions interconnect, highlighting the synergies and challenges inherent to combining tracking, Re-ID, and AU across multiple cameras.
To ensure a thorough categorization of methods, we further analyze the unique characteristics of each approach, focusing on key aspects such as existing data fusion techniques and cross-view learning. By providing this systematic framework, we offer a comprehensive perspective on the state-of-the-art MVMC methods, with clear sub-categories and considerations for each. All methods are methodically presented in subsequent sections.

\renewcommand{\arraystretch}{0.8}
\sloppy

\begin{table*}[!t]
\centering
\tiny
\caption{Summary of MVMC Re-ID Approaches.}
\label{table:mvmc_reid}

\resizebox{\textwidth}{!}{%
\begin{tabularx}{1.3\textwidth}{@{}
    C{0.4cm}   % #
    L{2cm}   % Method
    L{2.0cm}   % Venue/Year
    L{4cm}   % Feature Extractor
    L{3.9cm}   % M-ReID Approach
    L{3.9cm}   % Dataset
    L{3.1cm}   % Evaluation Metric
    C{1.5cm}   % Code
@{}}
\toprule
\textbf{\#} & \textbf{Method} & \textbf{Venue,Year} & \textbf{Feature Extractor} & \textbf{M-ReID Approach} & \textbf{Dataset} & \textbf{Evaluation Metric} & \textbf{Source Code} \\
\midrule
1 & ViewDic \cite{karanam2015person} & ICCV,15& Color, LFDA, LBP, HoGHoF & Viewpoint-Invariant Dictionary & iLIDS-VID, CAVIAR4REID & CMC & -- \\ \midrule
2 & SARC3D \cite{baltieri2015mapping} & IJCV,15& Histogram, HOG & SARC3D & 3DPeS & CMC, AUC & -- \\ \midrule
3 & DeepRank \cite{chen2016deep} & TIP,16& Custom CNN & Learning-to-Rank & VIPeR, CUHK01, CAVIAR4REID & CMC, Rank, TTR & -- \\ \midrule
4 & UTL-ReID \cite{peng2016unsupervised} & CVPR,16& HOG, LBP, Histogram & Unsupervised Transfer Learning & VIPeR, PRID, CUHK01 & CMC, Rank-1 & -- \\ \midrule
5 & Handoff-ReID \cite{shah2016multi} & Neurocom,16& HSIFT, CHF, SVM & Multi-Camera Handoff & iLIDS, ViPer, Custom & CMC, Rank & -- \\ \midrule
6 & ODBoA \cite{ma2016orientation} & arXiv,16& HOG, LAB, HSV & O-Driven Bag of Appearances & CUHK01, PKU-ReID, Market-1203 & CMC, Rank-1 & -- \\ \midrule
7 & WHF-ReID \cite{mumtaz2017weighted} & IPTA,17& LOMO, GOG, WHOS, PCA & Weighted Hybrid Features & VIPeR, ETHZ-1, CUHK02 & CMC, Rank-1,10 & -- \\ \midrule
8 & PRW \cite{zheng2017person} & CVPR,17& R-CNN, ResNet & FCN & PRW, Market-1501, Inria & mAP, Rank-1,10, IoU & \href{https://github.com/liangzheng06/PRW-baseline?utm_source}{Link} \\ \midrule
9 & MS-Deep \cite{qian2017multi} & ICCV,17& CNN & Multi-Scale FCN & CUHK03, CUHK01, PRID2011 & Rank-1,5,10 & -- \\ \midrule
10 & DAME \cite{yu2018unsupervised} & TPAMI,18& ResNet-56 & Deep Asymmetric Metric Embedding & CUHK01, CUHK03, VIPeR, PRID & Rank-1 & \href{https://github.com/liangzheng06/PRW-baseline?utm_source}{Link} \\ \midrule
11 & JSTL-DGD \cite{feng2018learning} & TIP,18& AlexNet, JSTL\_DGD & View-Specific Networks & VIPeR, CUHK03, SYSU, Market-1501 & CMC & -- \\ \midrule
12 & CSP \cite{dai2018cross} & PR,18& LOMO, GOG & Cross-View Semantic Projection & VIPeR, GRID, PRID450s, CUHK01/03 & CMC & -- \\ \midrule
13 & Snippet-ReID \cite{chen2018video} & CVPR,18& ResNet-50 & Snippet Aggregation & PRID2011, iLIDS-VID, MARS & Rank-1,5,10, mAP & -- \\ \midrule
14 & MV-Clust \cite{xin2019semi} & PR,19& CaffeNet, VGG16, Res50 & K-means / NK-means & Market-1501, DukeMCMT-reID & CMC, mAP, Rank-1-5-10 & -- \\ \midrule
15 & SliceReID \cite{zhang2019improving} & IET-CV,19& ResNet-50 & Multi-Transfer Learning & DukeMTMC, Market-1501, CUHK03 & Rank-1,5,10,20, mAP & -- \\ \midrule
16 & MV-Sim \cite{ainam2019view} & Access,19& ResNet-50 & Multi-View Similarity (n-pair) & Market-1501, CUHK03, DukeMCMT & Rank-1,5,10, mAP & \href{https://github.com/jpainam/multi-view}{Link} \\ \midrule
17 & NVAN \cite{liu2019spatially} & ICIT,19& NVAN & Non-Local Attention & MARS, DukeMTMC-VID & Rank-1, mAP & \href{https://github.com/jackie840129/STE-NVAN}{Link} \\ \midrule
18 & NLATN \cite{rao2019non} & AVSS,19& CNN & Siamese Temporal Network & PRID2011, iLIDS-VID, SDU-VID & Rank-1,5,10,20, CMD & \href{https://github.com/rshivansh/Non-Local-Attention-Person-Re-ID?tab=readme-ov-file}{Link} \\ \midrule
19 & MC-VehReID \cite{tan2019multi} & CVPR,19& R-CNN, SENet, FPN & DeepSORT+Spatio-Temporal & CityFlow-ReID, VehicleID, CityFlow & Rank-1,5,10 & -- \\ \midrule
20 & JDG-ReID \cite{zheng2019joint} & CVPR,19& ResNet-50, PatchGAN & Joint Disc.+Generative & Market-1501, DukeMCMT-reID, MSMT17 & CMC, mAP, FID & -- \\ \midrule
21 & FG-ReID \cite{yin2020fine} & IJCV,20& ResNet-50, RNN-mask, LSTM & Fine-Grained Features & FGPR, MARS, iLIDS & Rank-1,5, mAP & -- \\ \midrule
22 & TempFusion \cite{jiang2020rethinking} & AAAI,20& ResNet-50 & Temporal Fusion & CUHK03, DukeMTMC, Market-1501 & Rank-1,5, mAP & -- \\ \midrule
23 & GraphFusion \cite{zhou2019person} & TNNLS,20& RGB, HSV, Lab, HOG, LBP, LOMO & Adaptive Graph Learning & VIPeR, PKU-Reid, CUHK01/03 & Rank-1 & -- \\ \midrule
24 & RMatch \cite{lin2020recurrent} & MTAP,20& ResNet, RNN & Spatial Alignment Matching & Market-1501, DukeMCMT, CUHK03 & CMC, Rank, mAP & -- \\ \midrule
25 & MACG \cite{yan2020learning} & TPAMI,20& ResNet-50 & Multi-Attention Context Graph & CUHK-SYSU , MCTS, DukeMCMT & mAP, Rank & \href{https://github.com/daodaofr/group_reid?utm_source}{Link} \\ \midrule
26 & GCGNN \cite{zhu2020group} & TMM,20& ResNet-34 & Group Context GNN & iLIDS Group, Road Group, Duke Group & CMC, Rank & -- \\ \midrule
27 & TSSL-ReID \cite{chang2020transductive} & PR,20& ResNet-50 & Transductive Semi-Supervised & Market-1501, DukeMCMT, CUHK03 & CMC, Rank-1,5 & -- \\ \midrule
28 & UPA \cite{ding2021beyond} & TIFS,21& DenseNet-121, ShuffleNet & Universal Adv. Perturbation & Market-1501, DukeMCMT, MARS & mAP, mDR, Rank & -- \\ \midrule
29 & STMN \cite{eom2021video} & ICCV,21& ResNet-34, LSTM & Spatio-Temporal Memory & MARS, Duke-VID, LS-VID & mAP, Rank-1 & \href{https://github.com/cvlab-yonsei/STMN}{Link} \\ \midrule
30 & MoCo-ReID \cite{fu2021unsupervised} & CVPR,21& ResNet-50 & MoCo v2 Pretraining & Lu Person, DukeMTMC, Market-1501 & mAP, Rank-1 & \href{https://github.com/DengpanFu/LUPerson?tab=readme-ov-file}{Link} \\ \midrule
31 & PSTA \cite{wang2021pyramid} & ICCV,21& ResNet-50 & Pyramid ST Aggregation & MARS, Duke-VID, iLIDS-VID & mAP, CMC, Rank-1,5,10 & \href{https://github.com/WangYQ9/VideoReID_PSTA?tab=readme-ov-file}{Link} \\ \midrule
32 & City-VehReID \cite{wu2021multi} & CVPR,21& EfficientDet, ResNet & City-Scale Vehicle ReID & VeRi-776 & mAP, Rank-1, IDF1 & -- \\ \midrule
33 & MVFF \cite{xu2021multi} & KBS,21& ResNet-50, GNN & Multi-View Feature Fusion & Market-1501, Duke, CUHK03 & CMC, mAP & \href{https://github.com/Yinsongxu/MVMP_MFFN}{Link} \\ \midrule
34 & OA-MTMC \cite{specker2021occlusion} & CVPR,21& ResNet-101 & Occlusion-Aware MTMC & VehicleX, CityFlowV2 & mAP, Rank-1 & -- \\ \midrule
35 & SSAN \cite{ding2021semantically} & arXiv,21& ResNet-50, Bi-LSTM & Self-Aligned Network & CUHK-PEDES, ICFG-PEDES & Rank-1,5,10 & \href{https://github.com/zifyloo/SSAN}{Link} \\ \midrule
36 & VTM \cite{zheng2021viewpoint} & Neurocom,21& ResNet-50 & Viewpoint Transform Matching & Market-1501, DukeMCMT, CUHK03 & mAP, CMC, Rank & -- \\ \midrule
37 & GCN-ReID \cite{wang2020cross} & JIFS,21& HOG, SILTP & Graph Conv Network & SYSU-sReID, CUHK03, Market-1501 & Rank-1,5,10,20 & -- \\ \midrule
38 & MCAM \cite{song2021mask} & Neurocom,21& CaffeNet, ResNet-50, MSCAN & Mask-Guided Contrastive Attention & MARS, Market-1501, CUHK03, Duke & CMC, Rank-1, mAP & -- \\ \midrule
39 & Crossroad \cite{liu2021city} & CVPR,21& ResNet50-IBN, ResNet101-IBN-a & YOLOv5x, JDETracker & MTMCT, CityFlow & IDF1, Prec, Rec & \href{https://github.com/LCFractal/AIC21-MTMC?utm_source}{Link} \\ \midrule
40 & SportsReID \cite{comandur2022sports} & arXiv,22& ResNet-50, OSNet, DeiT-Tiny & Sports Player ReID & SoccerNet & mAP, Rank-1 & \href{https://github.com/shallowlearn/sportsreid}{Link} \\ \midrule
41 & MVDN \cite{yang2022cross} & MTAP,22& CapsuleNet & Multi-View Decomposition & CUHK03, Market-1501, VIPeR & Rank-1, mR1 & -- \\ \midrule
42 & SGMN \cite{chen2022saliency} & TCSVT,22& ResNet-50 & Saliency + Granularity & iLIDS-VID, MARS, Duke-VID, LS-VID & Rank-1,5,20, mAP & -- \\ \midrule
43 & TIMS \cite{yang2022traffic} & TITS,22& ResNet-50, YOLOv3 & TrackletNet & LHTV, CityFlow, VeRi-776 & IDF1, IDP, IDR, Acc & -- \\ \midrule
44 & MVA-ReID \cite{liu2022person} & PRML,22& ResNet-50 & Multi-View Attention & Market-1501, DukeMCMT, CUHK03-L & Rank-1, mAP & -- \\ \midrule
45 & MC-ReID \cite{bagga2022person} & ICIT,22& CNN & Multi-Camera System & CUHK03 & Rank-1, mAP & -- \\ \midrule
46 & ODA-ReID \cite{rami2022online} & CVPR,22& ResNet-50, DBSCAN, MMT & Online Domain Adaptation & Market-1501, DukeMCMT, MSMT17 & mAP, Rank-1 & -- \\ \midrule
47 & ColorAttack \cite{gong2022person} & CVPR,22& DenseNet, IBN-ResNet-101 & Color Attack + Defense & Market-1501, DukeMCMT & mAP, Rank & -- \\ \midrule
48 & DistReID \cite{kumar2022person} & ICIT,22& YOLOv4, DeepSORT, Cosine & Distributed Tracking+ReID & EPFL, MARS & MOTA, Acc & -- \\ \midrule
49 & CTL-ReID \cite{liu2023deeply} & TNNLS,23& ResNet-50, ViT & Conv-Transformer Learning & MARS, iLIDS-VID, PRID2011 & mAP, Rank-1,5 & \href{https://github.com/flysnowtiger/DCCT}{Link} \\ \midrule
50 & DeepChange \cite{xu2023deepchange} & ICCV,23& MobileNetV3, ViT-B/16 & Clothes-Change Benchmark & DeepChange & Rank-1,5,10,20, mAP & \href{https://github.com/PengBoXiangShang/deepchange?utm_source}{Link} \\ \midrule
51 & PartFormer \cite{ni2023part} & ICCV,23& ViT-Base & Part-Aware Transformer & Market-1501, MSMT17, CUHK03 & Rank-1, mAP & \href{https://github.com/liyuke65535/Part-Aware-Transformer}{Link} \\ \midrule
52 & UniReID \cite{che2023efficient} & ICACA,23& ResNeXt & Unified Multi-Class ReID & Market-1501, VeRi-776 & CMC, Rank-1,5,10, mAP & -- \\ \midrule
53 & KRec \cite{chen2019person} & JVCIR,23& ResNet-50, KNN & Re-Ranking (k-reciprocal) & Market-1501, CUHK03, DukeMCMT & Rank-1,5,10, mAP & -- \\ \midrule
54 & JDL \cite{peng2023joint} & SPIC,23& ResNet-50, Attention & Joint Learning (Diverse Knowledge) & Market-1501, CUHK03, VeRi-776 & CMC, Rank-1,5, mAP & -- \\ \midrule
55 & CSA-Net \cite{wang2023context} & arXiv,23& ResNet-50 & Context Sensing Attention & MARS, LS-VID, Duke-VID & Rank-1, mAP & -- \\ \midrule
56 & MC-CPD \cite{peng2023revisiting} & TPAMI,23& ABD-Net, TransReID, RDQN & Multi-Camera Continuous ReID & VIPeR, CUHK03, Market-1501, Duke & Rank-1,5,10 & -- \\ \midrule
57 & RC-ReID \cite{liu2023reliable} & TCSVT,23& ResNet-50, DBSCAN & Random-Camera Guided & Market-1501, DukeMCMT, MSMT17 & CMC, mAP, Rank & -- \\ \midrule
58 & ProtoReID \cite{wang2023rethinking} & arXiv,23& ResNet-50, ProNet++/ProNet & Projection-on-Prototypes & Market-1501, CUHK03, MSMT17 & CMC, Rank-1, mAP & -- \\ \midrule
59 & SRF-Net \cite{tian2023self} & IEEE TIM,23& ResNet-50, SRFNet & Self-Regulation Feature Network & Market-1501, CUHK03, DukeMCMT & mAP, Rank-1 & -- \\ \midrule
60 & MSPA \cite{khan2023visual} & IEEE JSTSP,23& ResNet-50, ConvLSTM, Pyramids & Multi-Scale Pyramid Attention & Market-1501, DukeMCMT-reID & CMC, Rank, mAP & -- \\ \midrule
61 & ColorPrompt \cite{gu2023color} & arXiv,23& ResNet-50, ShuffleNet-V2 & Data-Free Continual UDA & Market-1501, CUHK-SYSU03 & Rank-1, mAP & -- \\ \midrule
62 & CCPG \cite{li2023depth} & CVPR,23& AP3D, PSTA, BiCnet-TKS, PiT & Cloth-Changing Benchmark & CCPG & Accuracy, mAP & -- \\ \midrule
63 & MDMT \cite{liu2023robust} & TMM,23& ResNet-50, SIFT, KNN, RANSAC & Multi-Drone Occlusion ReID & MDMT & MOTA & -- \\ \midrule
64 & RGB-TPM \cite{wei2023video} & Vis. Comput.,23& ResNet-50, VGG & RGB Triple Pyramid Model & MARS, iLIDS-VID, PRID2011 & Rank, mAP & -- \\ \midrule
65 & MVIA \cite{xiahou2023identity} & TCSVT,23& ResNet-50, IC-GAN & Multi-Viewpoint Aggregation & Market-1501, DukeMCMT, CUHK03 & CMC, Rank-1, mAP & -- \\ \midrule
66 & DRE \cite{liu2024diverse} & arXiv,24& ViT-B/16 & Diverse Representation Embedding & Market-1501, CUHK-SYSU, Duke & Rank-1, mAP & -- \\ \midrule
67 & Hi-AFA \cite{dong2024hierarchical} & Access,24& OSNet & Hierarchical Attentive Aggregation & Market-1501, Duke, MSMT17, CUHK03 & mAP, CMC, Rank & -- \\ \midrule
68 & TMT-Re-ID \cite{liu2024video} & IEEE T-ITS,24& ST/CT, LSTM, 3D CNN & Trigeminal Transformer & iLIDS-VID, MARS, Duke-VID, LS-VID & Rank-1,5, mAP & -- \\ \midrule
69 & MMMatch-VI \cite{shi2024multi} & ECCV,24& ResNet-50 & Multi-Memory Matching & SYSU-MM01, RegDB & CMC, Rank, mAP & -- \\ \midrule
70 & UFFM+AMC \cite{che2025enhancing} & KBS,24& BoT, CLIP-ReID, ResNet-50, KNN & Uncertainty Fusion + Auto-Weighted & Market-1501, Duke, MSMT17 & CMC, Rank-1, mAP & -- \\ \midrule
71 & SCFP \cite{zhao2024end} & AI Review,24& ViT & Corrupted Feature Prediction & P-DukeMTMC, Market-1501, CUHK03 & Rank-1,5, mAP & -- \\ \midrule
72 & DJAA \cite{chen2024anti} & TPAMI,24& ResNet-50 & Anti-Forgetting Adaptation & Market, CUHK-SYSU, CUHK03 & Rank-1, mAP & -- \\ \midrule
73 & ACLS \cite{khan2024intelligent} & EAAI,24& ResNet-50, Transformer & Intelligent Correlation Learning & Market-1501, DukeMCMT, CUHK03 & CMC, mAP, Rank & -- \\ \midrule
74 & MVI2P \cite{dong2024multi} & Inf. Fusion,24& ResNet-50 & Multi-View Integration & Occluded-Duke, DukeMCMT, Market-1501 & CMC, mAP & -- \\ \midrule
75 & ACPL \cite{chen2024unsupervised} & AAAI,24& ResNet-50 & Adaptive Clustering (Group) & ROADGroup, CSG, SYSU & CMC, Rank, mAP & -- \\ \midrule
76 & MP-ReID \cite{zhai2024multi} & AAAI,24& CLIP, ChatGPT, ViT-B/16 & Multi-Prompts Cross-Modal Align. & Market-1501, DukeMCMT-reID & mAP, Rank & -- \\ \midrule
77 & TAF+TAA+B \cite{ma2025recursively} & EAAI,25& HRNet, ResNet-50 & Recursive Fine-Grained ST & MARS, Duke-VID, PRID2011 & CMC, mAP & -- \\ \midrule
78 & RFG-CC \cite{yin2025robust} & Mathematics,25& ResNet-50, SCHPNet & Robust Fine-Grained (Clothes) & PRCC, LTCC, DeepChange, VC-Clothes & Rank-1, mAP & -- \\ \midrule
79 & DetReIDX \cite{hambarde2025detreidx} & arXiv,25& YOLOv8, CLIP-ReID & UAV ReID Dataset & DetReIDX & Rank-1,5,10, mAP & \href{https://github.com/kailashhambarde/DetReIDX/tree/main}{Link} \\ \midrule
80 & ReMix \cite{mamedov2025remix} & WACV,25& ResNet-50-IBN, DBSCAN & Mixture Training (Encoders) & Market-1501, CUHK03-NP, DukeMCMT & CMC, Rank-1, mAP & -- \\ \midrule
81 & 3D-DG-ReID \cite{jiao2025generalizable} & IEEE TIFS,25& ResNet-50, SMPL, R-Texformer & Domain-Generalizable (3D) & Market-1501, Duke, CUHK-SYSU & CMC, Rank, mAP & -- \\ \midrule
82 & CA-GMDAL \cite{ran2025camera} & PR,25& ResNet-50, DBSCAN, GCN & Camera-Aware Multi-Domain Adapt. & Market-1501, MSMT17, PersonX & CMC, mAP, Rank & -- \\ \midrule
83 & MOR-RNN \cite{mk2025enhanced} & Neurocom,25& CNN, LSTM & Mixed-Order Relation-Aware RNN & Market-1501, DukeMCMT-reID & CMC, Rank, ROC & -- \\ \midrule
84 & FedMutual \cite{liu2025personalized} & ACM TOMM,25& ResNet-50, DBSCAN & Federated Mutual Learning & DukeMCMT, Market-1501, CUHK01 & CMC, mAP & -- \\ \midrule
85 & AG-VPReID \cite{nguyen2025ag} & CVPR,25& Texformer, PhysPT, YOLOv8x & Aerial–Ground Video ReID & MARS, LS-VID, G2A VReID & CMC, Rank-1, mAP & \href{https://github.com/agvpreid25/AG-VPReID-Net}{Link} \\ \midrule
86 & HDA-Net \cite{zhang2025hda} & ICETCI,25& ViT-Base/12, TFSF, HIA & Hierarchical Dynamic-Aware Attention & Market-1501, DukeMCMT-ReID & Rank-1,5,10, mAP & -- \\
\bottomrule
\end{tabularx}
}
\end{table*}

\subsection{MVMC Tracking}
MVMC tracking is the cornerstone of CVS, where the objective is to maintain identity-preserving trajectories of multiple targets across spatially distributed viewpoints. Unlike single-camera MOT, MVMC tracking must cope with challenges such as occlusions, camera hand-offs, and large viewpoint variations. Its importance spans a wide range of applications, from intelligent transportation and crowd surveillance to collaborative robotics. The literature as tabulated in \ref{table:mvmc_tracking}, can be broadly categorized into traditional and early convolutional neural networks (CNN) based methods, graph-based association approaches, spatio-temporal, contrastive learning methods, end-to-end transformer and BEV-based architectures, and application-specific frameworks. Popular frameworks of MVMC tracking are presented in Figure \ref{fig=Tracking-Frameworks}.
\subsubsection{Traditional and Early CNN-Based Methods}
The first generation of MVMC tracking systems relied heavily on handcrafted appearance descriptors and simple motion models. Classical pipelines extracted HOG, LBP, or color histograms from detected bounding boxes and used bipartite graph matching or Hungarian assignment to link observations across frames and views \cite{wen2017multi, xu2016multi, chavdarova2018wildtrack}. Kalman filters or constant-velocity motion predictors were employed to estimate trajectories, while geometry priors such as homography mappings provided additional cues for cross-camera handover. With the rise of CNNs, appearance modeling shifted from handcrafted descriptors to learned embeddings. Early CNN-based methods combined feature extractors such as AlexNet or VGG with conventional data association mechanisms \cite{baque2017deep, zhang1712multi}, improving robustness against illumination and pose changes. However, these methods still struggled in crowded scenarios and lacked scalability to large networks of cameras.
\subsubsection{Graph-Based Association Approaches}
To address the shortcomings of handcrafted matching, a major research thrust reformulated MVMC tracking as a graph partitioning or matching problem. In these methods, detections or short tracklets are represented as nodes in a graph, while spatial, temporal, and appearance affinities form the edges. Optimization techniques such as lifted multicut partitioning have been employed to group nodes belonging to the same identity, even under long-term occlusions. For example, LMGP combines lifted multicut with 3D geometry projections to associate pedestrians across overlapping views \cite{nguyen2022lmgp}. Similarly, the ReST framework constructs spatial-temporal graphs and reconfigures them over time through message passing, achieving globally consistent identities \cite{cheng2023rest}. These graph-based models demonstrate high resilience against occlusions and fragmented detections, and they remain competitive baselines on standard datasets.
\subsubsection{Spatio-Temporal and Contrastive Learning}
Another line of research leverages discriminative learning of spatio-temporal embeddings. Rather than explicitly solving graph optimization problems, these approaches learn feature spaces where embeddings of the same trajectory remain close across views. Contrastive trajectory learning frameworks, for instance, introduce memory banks to store historical trajectories and apply contrastive loss to distinguish between different identities\cite{yu2022towards}. Models such as GMT incorporate sliding time windows and global association transformers to link trajectories consistently across long temporal spans \cite{fan2024gmt}. The DIVOTrack framework extends this by jointly learning single-view and cross-view embeddings, ensuring both intra-view continuity and inter-view identity matching \cite{hao2024divotrack}. Such approaches highlight the shift toward representation learning as the driving force behind modern MVMC tracking.
\subsubsection{End-to-End Transformer and BEV-Based Frameworks}
Recent advances in deep architectures have led to end-to-end frameworks that unify detection, association, and cross-view reasoning within a single pipeline. Transformer-based models such as ADA-Track++ employ alternating detection and association queries, where self-attention mechanisms model intra- and inter-camera dependencies \cite{ding2024ada}. MCBLT applies DETR-inspired decoders and graph-based temporal association to handle long sequences \cite{wang2024mcblt}, while FusionTrack integrates transformer-based Re-ID modules with tracklet memory for robust association in unconstrained multi-view environments \cite{li2025fusiontrack}. Another promising direction is lifting multi-view features into a shared bird’s-eye-view (BEV) space, where geometric alignment simplifies both detection and tracking. Systems such as MITracker \cite{xu2025mitracker} and MUTR3D \cite{zhang2022mutr3d} exemplify this BEV-driven trend, enabling accurate 3D tracking across multiple synchronized cameras.
\begin{figure*}[!t]
    \centering
    \includegraphics[width=1\linewidth]{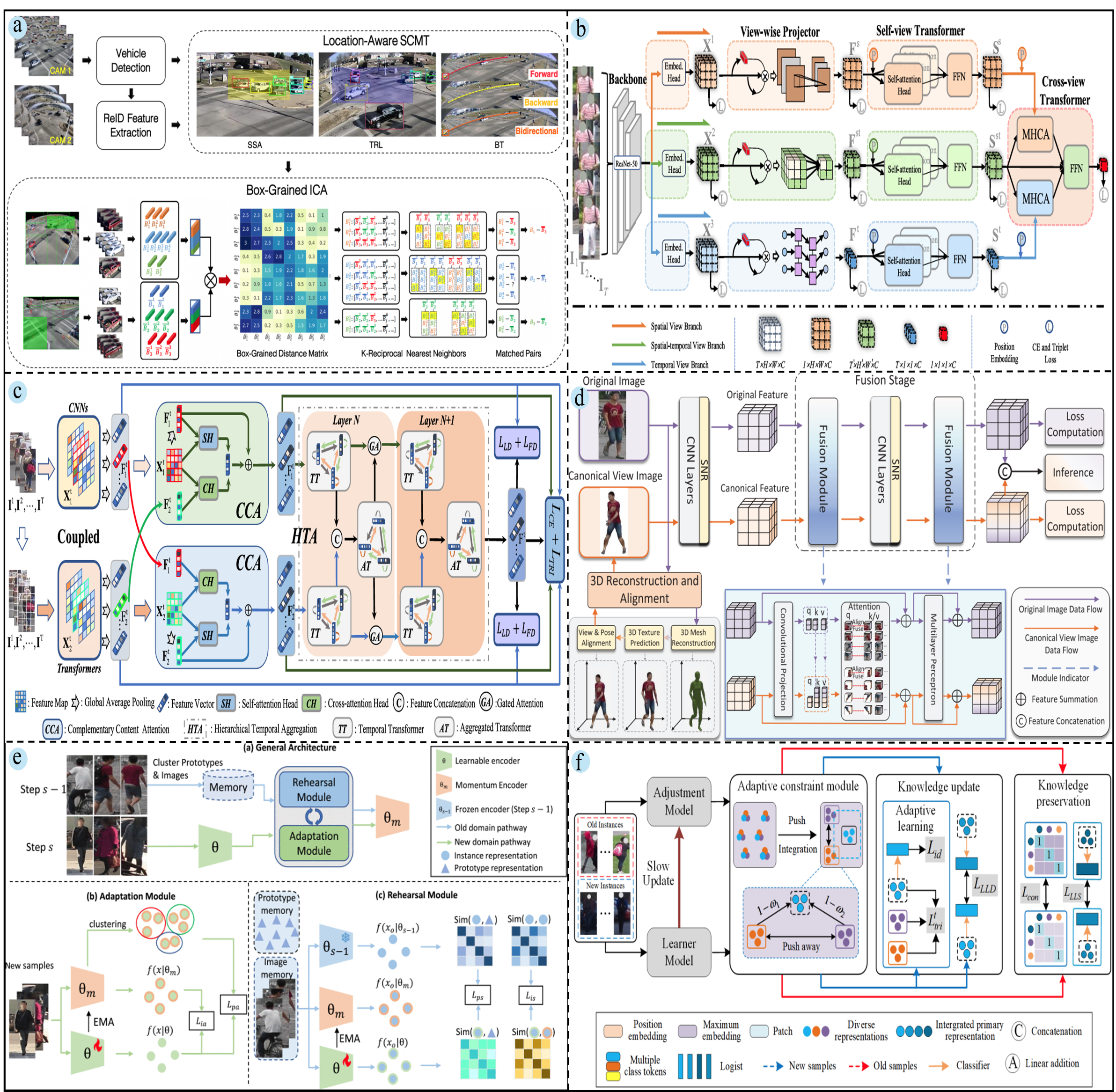}
    \caption{Overview of SOTA Multi-Camera Re-ID frameworks, showcasing various methods for object Re-ID across multiple cameras, including techniques for a) city-scale multi-camera vehicles Re-ID \cite{lu2023citytrack}, b) trigeminal transformers for person Re-ID \cite{liu2024video},  c) anti-forgetting adaptation \cite{chen2024anti}, d) convolution-transformer learning \cite{liu2023deeply}, e) 3D perspective re-identification \cite{jiao2025generalizable}, and f) lifelong person Re-ID \cite{liu2024diverse}. Each method addresses unique challenges in the multi-camera domain.
 }
    \label{fig=Re-ID_Frameworks}
\end{figure*}
\renewcommand{\arraystretch}{0.9}
\begin{table*}[!t]
\centering
\caption{Summary of MVMC Action Understanding Techniques.}
\label{table:mvmc_action}

\resizebox{\textwidth}{!}{%
\begin{tabularx}{1.5\textwidth}{@{}
    C{0.5cm}   % #
    L{3.0cm}   % Method
    L{2.2cm}   % Venue+Year
    L{5.0cm}   % Feature Extractor
    L{4.0cm}   % M-Action Approach
    L{4.6cm}   % Dataset
    L{3cm}   % Evaluation Metric
    C{2cm}   % Code
@{}}
\toprule
\textbf{\#} & \textbf{Method} & \textbf{Venu/Year} & \textbf{Feature Extractor} & \textbf{M-Action Approach} & \textbf{Dataset} & \textbf{Evaluation Metric} & \textbf{Source Code} \\
\midrule
1  & ViewFusion \cite{azis2016weighted}   & IET-CV,16   & 3D Joints, Pairwise, HoCubes & DTW + KNN & Custom & Accuracy, Conf. Matrix & -- \\ \midrule
2  & MHI-HOG \cite{murtaza2016multi}     & IET-CV,16   & HOG on MHIs & Nearest Neighbour & MuHAVi-uncut & Accuracy, ROC & -- \\ \midrule
3  & RobustFeat \cite{chou2017automatic} & NIPS,17     & Box, Cloud Features & NMC & KTH, Weizmann & Accuracy, LOOCV & -- \\ \midrule
4  & MotionFreq \cite{kase2017multi}     & ICIP,17     & Motion Frequency & SVM & IXMAS, i3DPost & Accuracy, LOOCV & -- \\ \midrule
5  & MCAD \cite{li2017multi}             & WACV,17     & STIP, Cuboid, IDT & SVM & MCAD & Accuracy, Mean Acc & -- \\ \midrule
6  & PKU-MMD \cite{liu2017pku}           & arXiv,17    & CNN, TSN, BN-Inception & STA-LSTM, JCRRNN & PKU-MMD & C-V Acc., mAPa, mAPv & -- \\ \midrule
7  & MTL-Act \cite{gao2017evaluation}    & MTAP,17     & HOG, HOF, MBH & Multi-Task Learning & IXMAS, NUCLA & Accuracy, LOSO & -- \\ \midrule
8  & CMTL-Repr \cite{hao2017multi}       & JVCIR,17    & iDT (HOG, HOF, MBHx, MBHy) & Sparse+Clustered MTL & M2I, IXMAS & Top-1 Acc & -- \\ \midrule
9  & MMA \cite{gao2018mma}               & MTAP,18     & BoW, IDT, C3D & NMC, NNC, GMMC & MMA & Acc & -- \\ \midrule
10 & RobustMV-AR \cite{chou2018robust}   & Access,18   & GMM, Box, Cloud Features & Handcrafted & MuHAVi, KTH, Weizmann & Accuracy, LOOCV & -- \\ \midrule
11 & HVI-Act \cite{liu2018hierarchically}& TCSVT,18    & HOG, HOF, MBHx & KNN + HVI & IXMAS, NUMA, MuHAVi & Accuracy, Cross-view & -- \\ \midrule
12 & DivAgg \cite{wang2018dividing}      & ECCV,18     & WDMM, HOG, LBP & SVM & NTU RGB+D 120, NUCLA & Cross-view/subject Acc. & \href{https://github.com/HakuSean/DA-Net?tab=readme-ov-file}{Link} \\ \midrule
13 & MS-CNN \cite{chenarlogh2019multi}   & ICSPIS,19   & 3D-CNN & FCN & IXMAS, New IXMAS & Acc & -- \\ \midrule
14 & DL+KSVD \cite{wang2019multi}        & J. Phys.,19 & AlexNet & K-SVD & IXMAS, WVU & Acc & -- \\ \midrule
15 & VCDN \cite{wang2019generative}      & ICCV,19     & ResNet-101 TSN & VCDN & UWA3D, MHAD & Classification Acc. & \href{https://github.com/wanglichenxj/Generative-Multi-View-Human-Action-Recognition}{Link} \\ \midrule
16 & VV-RGBD \cite{ji2019large}          & arXiv,19    & Joints, Temporal Dyn. & FCN+Softmax & Varying-view RGB-D & CS-CV Acc & -- \\ \midrule
17 & CA-LSTM \cite{bai2020collaborative} & arXiv,20    & LSTM Features & LSTM + Attention & NTU120, Ev-Action, UWA3D & Top-1 Acc & -- \\ \midrule
18 & CL-Net \cite{vyas2020multi}         & ECCV,20     & 3D CNN & CL-Net & NTU60, NUCLA & Classification Acc. & \href{https://github.com/svyas23/cross-view-action}{Link} \\ \midrule
19 & ConfluxLSTM \cite{ullah2021conflux} & Neurocom,20 & VGG19 & Conflux LSTM & MCAD, NUCLA & Acc & -- \\ \midrule
20 & MOD20 \cite{perera2020multiviewpoint}& THMS,20    & VGG-f, Motion CNN & KRP-FS, BKRP & MOD20 & Classification Acc. & \href{https://github.com/asankagp/mod20}{Link} \\ \midrule
21 & MS-G3D \cite{liu2020disentangling}  & CVPR,20     & MS-G3D & ST-GCN + GAP + Softmax & NTU120, NTU60, Kinetics-400 & Classification Acc. & \href{https://github.com/kenziyuliu/ms-g3d}{Link} \\ \midrule
22 & NTU120 \cite{liu2019ntu}            & TPAMI,20    & VGG, MobileNet & ST-LSTM, GCA-LSTM & NTU RGB+D 120 & Acc & -- \\ \midrule
23 & BoSW \cite{hu2020joint}             & Info. Sci.,20 & BoSW & Joint Model & IXMAS, WVU, M2I & Accuracy, NMI & -- \\ \midrule
24 & TSN+GRU \cite{bui2020multi}         & Procedia CS,20 & Inception-BN & GRU+TSN & NUCLA, MicaHandGesture & Acc., Cross-view & -- \\ \midrule
25 & SSS-Repr \cite{shao2020learning}    & TCSVT,20   & 3D-CNN, LSTM, MSNN & Self-Similarity & NTU60, UESTC, NUCLA & Acc., Cross-view & -- \\ \midrule
26 & MB-MHI \cite{naeem2020multiple}     & AJSE,20    & ResNet-101, HOG & SVM & MuHAVi-MAS8, MAS14 & Acc, LOCO, LOAO & -- \\ \midrule
27 & CGAN-AR \cite{gedamu2021arbitrary}  & PR,21      & ResNet, CGAN, WGAN-GP & FCN & NTU60, NTU120, UESTC & Accuracy, Cross-view & \href{https://github.com/GedamuA/TB-GAN}{Link} \\ \midrule
28 & AttnNet \cite{nguyen2021attention}  & Procedia CS,21 & ResNet-18/34, R(2+1)D & FCN & NTU120, NUCLA, MicaHand & C-V/S Acc & -- \\ \midrule
29 & LMSL \cite{wang2021continuous}      & TCSVT,21   & Handcrafted Features & LMSL & MHAD, UWA3D, DHA, MuHAVi & Acc., AUC, Rec, F1 & -- \\ \midrule
30 & X-ODistill \cite{xu2021cross}       & Neurocom,21 & I3D & FC + Softmax & IXMAS, NUCLA & Acc & -- \\ \midrule
31 & MLRMV \cite{liu2021mlrmv}           & IVC,21     & HOG, HOF, MBH, DT & SVM & NTU120, WVU, NUCLA & Accuracy, FPS & -- \\ \midrule
32 & IKEA-ASM \cite{ben2021ikea}         & WACV,21    & ResNet-50 & I3D + Softmax & IKEA ASM, Kinetics, NTU120 & Top-1/5 Acc., mAP & \href{https://github.com/IkeaASM/ikea_asm_dataset}{Link} \\ \midrule
33 & CKD \cite{kumar2021collaborative}   & IVC,21     & I3D, LSTM, GAT & FCN & NTU120, PKU-MMD & Acc & -- \\ \midrule
34 & T-VLAD \cite{naeem2021t}            & PRL,21     & C3D & SVM & IXMAS, MuHAVi, UCF50 & Acc & -- \\ \midrule
35 & UP-LBP \cite{nigam2021multiple}     & ICIP,21    & LBP & SVM & CASIA, IXMAS & Conf. Matrix & -- \\ \midrule
36 & PC-MDA \cite{tran2021pairwise}      & Access,21  & ResNet-50, C3D, LSTM & KNN & IXMAS, MuHAVi, NTU60, MICA & Acc & -- \\ \midrule
37 & JDLA \cite{liu2022task}             & DSP,22     & SSM, HOG, HOF & Linear Classifier & IXMAS, NUCLA, WVU & Acc & -- \\ \midrule
38 & I3D+CA \cite{zhao2022compositional} & PLOS One,22& I3D, Cross Attention & FCN & LEMMA, IKEA ASM & Prec, Rec, F1, Acc. & -- \\ \midrule
39 & MultiTrans \cite{yasuda2022multi}   & ICASSP,22  & ResNet-34, VGGish & Transformer Fusion & MM-Office & mAP & -- \\ \midrule
40 & VKTeN \cite{liang2022view}          & IVC,22     & ResNet-101 & CGANs, SSN & MHAD, UWA3D & Top-1 Acc & -- \\ \midrule
41 & SAP \cite{hou2022shifting}          & Pro-MM,22  & MS-G3D, ResNet-50, Attention & SAP & NTU60, NTU120, Kinetics-Skeleton & Top-1/5, C-V Acc & \href{https://github.com/ideal-idea/SAP}{Link} \\ \midrule
42 & VCD \cite{zhong2022vcd}             & ICASSP,22  & ResNet3D & Disentanglement & NUCLA, NTU60 & Acc., Cross-view & -- \\ \midrule
43 & VSP-F \cite{zhang2023modeling}      & CVPR,23    & ResNet-50 & PCL+SCL & IKEA ASM & Phase Acc. & \href{https://github.com/hengRUC/VSP?utm_source}{Link} \\ \midrule
44 & UVS \cite{liu2023unsupervised}      & IVC,23     & iDT+Fisher Vectors & Unsupervised Segmentation & WVU, NTU120, NUCLA & Acc & -- \\ \midrule
45 & MVSA-Net \cite{asali2023mvsa}       & arXiv,23   & TD-Convs, Conv-GRU, YOLOv5 & FCN & Onion Sorting & Accuracy, 5-fold CV & -- \\ \midrule
46 & 3MDAD \cite{lin2023multi}           & arXiv,23   & Swin Transformer & KL-Distillation & IXMAS, NUCLA, 3MDAD & Accuracy, Cross-view & -- \\ \midrule
47 & MAWKDN \cite{quan2023mawkdn}        & TCSVT,23   & BN-Inception, Wavelets & Cross-view Attention KD & MMAct, UTD, Berkely-MHAD & C-S/A Acc & -- \\ \midrule
48 & ViewCon \cite{shah2023multi}        & WACV,23    & S3D & Contrastive Loss & NTU120, NTU60, NUCLA & Acc. (ViewCon) & \href{https://github.com/kshah33/ViewCon}{Link} \\ \midrule
49 & Driver-MVAR \cite{zhou2023multi}    & CVPR,23    & ViT-L/16 & ViT Classifier & SynDD2 & Clip-level Acc., mOS & -- \\ \midrule
50 & Syn2Real \cite{reddy2023synthetic}  & ICRA,23    & ResNet50, X3D & DANN, CO2A & RoCoG & Top-1 Acc. & \href{https://github.com/reddyav1/RoCoG-v2?utm_source}{Link} \\ \midrule
51 & MKDT \cite{lin2023multi}            & arXiv,23   & Swin Transformer & MKDT & IXMAS, NUCLA, 3MDAD & Acc & -- \\ \midrule
52 & DRDN \cite{liu2023dual}             & TIP,23     & ResNet-3D & Disentanglement & NUCLA, NTU60, IXMAS & Acc., Cross-view & -- \\ \midrule
53 & MVFouls \cite{held2023vars}         & CVPR,23    & ResNet & Video Assistant Referee & SoccerNet, NTU120, NUCLA & Acc & -- \\ \midrule
54 & JRSMs \cite{kumar2023view}          & RMKMATE,23 & JRSM, CNN & Surface Map ConvNet & KLU3DAction, NTU60, PKU-MMD & Acc., Prec., Recall & -- \\ \midrule
55 & ASL-Act \cite{nguyen2024action}     & ACMMM Asia,24 & ResNet-18/34, ViT & Frame+Classifiers & MM-Office & mAP-C, mAP-S & \href{https://github.com/thanhhff/MultiASL?utm_source}{Link} \\ \midrule
56 & Guided-MELD \cite{yasuda2024guided} & arXiv,24   & ResNet-34, VGGish & Self-Distillation & MM-Office, MM-Store & mAP, ROAUC & -- \\ \midrule
57 & Real \cite{panev2024exploring}      & WACV,24    & SlowFast, X3D, MViT & Multi-view Synthetic Fusion & REMAG & Top-1 Acc. & -- \\ \midrule
58 & CVFN \cite{yang2024cross}           & ICVISP,24  & NTU60 & ResNet3D & NTU120, NTU60 & Top-1 Acc. & -- \\ \midrule
59 & DVANet \cite{siddiqui2024dvanet}    & AAAI,24    & R3D & Gated Adaptive Fusion & NUCLA, NTU120, PKU-MMD & Top-1 Acc. & \href{https://github.com/NyleSiddiqui/MultiView_Actions}{Link} \\ \midrule
60 & HyperMV \cite{gao2024hypergraph}    & TPAMI,24   & ResNet-18 & Hypergraph NN & THUMV-EACT-50, DHP19 & Top-1/3/5 Acc. & \href{https://github.com/lujiaxuan0520/THU-MV-EACT-50?utm_source}{Link} \\ \midrule
61 & MVAFormer \cite{yamane2024mvaformer}& ICIP,24   & ViT-Base & Transformer & MMAct & Acc., Prec., Rec, F1 & -- \\ \midrule
62 & DrivingBehav \cite{nguyen2024multi} & CVPR,24    & VideoMAE, UniformerV2, X3D & NMS + FCN & AI City Challenge & Acc., Prec., Rec, F1 & \href{https://github.com/SKKUAutoLab/aicity_2024_driving_action}{Link} \\ \midrule
63 & UVS-H-BLS \cite{liu2024multi}       & IVC,24     & IDT, Atom, FV & Broad Learning System & NTU120, NUCLA, WVU & Acc & -- \\ \midrule
64 & Mv2MAE \cite{shah2024mv2mae}        & arXiv,24   & ViT-Small & Masked Autoencoder & NTU60, NTU120, PKU-MMD, RoCoG & Cross-view Acc. & -- \\ \midrule
65 & ConvSTLSTM \cite{kurchaniya2024framework}& ICCICN,24 & VGG16, LSTM, URI-LBP & ConvST-LSTM & CASIA, IXMAS & Accuracy, Loss & -- \\ \midrule
66 & MV-TSHGNN \cite{ma2024multi}        & TIP,24     & SHGNN, THGNN, STHGNN & Time-Series HGNN & NTU60, NTU120 & Acc, Prec., F1 & -- \\ \midrule
67 & WS-VAD \cite{pereira2024video}      & IJCB,24    & I3D, YOLO, VGG16, ResNet, DenseNet & Multi-Camera Overlap VAD & HQFS, Up-Fall, PETS2009 & Precision, F1 & -- \\ \midrule
68 & SAGA-Track \cite{nguyen2024multi}   & PR,24      & ResNet-50, Transformer, Kalman & Global Association & nuScenes & mAP, mATE, mASE & -- \\ \midrule
69 & MV-GMN \cite{lin2025mv}             & arXiv,25   & DeiT, Skeleton, MSSNet & State Space Model & NTU120, NTU60, PKU-MMD & Acc & -- \\ \midrule
70 & MMASL \cite{nguyen2025action}       & ACM TOMM,25 & ViT, AST & MIL & MM-Office & mAP-C, mAP-S & \href{https://github.com/thanhhff/MMASL/}{Link} \\ \midrule
71 & MultiSensor-Home \cite{nguyen2025multisensor}& arXiv,25 & ViT & Transformer Fusion & MM-Office & mAP-C, mAP-S & \href{https://github.com/thanhhff/MultiTSF}{Link} \\ \midrule
72 & UPOG-MVNet \cite{liu2025beyond}     & arXiv,25   & X3D (3D CNN) & FC+Softmax & UAV Human, MOD20, DroneAction & Top-1 Acc. & -- \\ \midrule
73 & Multi-VSL \cite{dinh2025sign}       & WACV,25    & I3D, Swin, MViT, VTNPF, YOLOv9 & SLR & Multi-VSL (200–1000) & Top-1/5 Acc. & \href{https://github.com/Etdihatthoc/Multi-VSL_WACV_2025?utm_source}{Link} \\ \midrule
74 & MAF-Net \cite{xie2025maf}           & PLOS One,25 & VGG19, Self-Attn, Bi-LSTM, ST-GCN & MAF-Net & NTU60, UTD, SYSU, MMAct & Acc, F1 & -- \\ 
\bottomrule
\end{tabularx}
}
\end{table*}

\subsubsection{Application-Oriented MVMC Tracking Frameworks}
In parallel to benchmark-driven development, numerous application-specific MVMC trackers have been proposed for domains such as urban mobility, sports analytics, and panoramic surveillance. Argus, for instance, enables distributed analytics across smart camera networks by coordinating workload distribution and identity assignment \cite{yi2024argus}. OmniTrack combines end-to-end association with hybrid tracking-by-detection to maintain consistent tracking in omnidirectional video streams. The All-Day MCMT framework fuses RGB and infrared modalities, providing illumination-invariant tracking across daytime and nighttime conditions \cite{fan2025all}. City-scale trackers such as CityTrack incorporate topology-aware modules and traffic semantics to improve vehicle tracking across urban environments \cite{yu2025citytrac}. Collectively, these frameworks demonstrate the adaptability of MVMC tracking to real-world deployment, bridging the gap between controlled research benchmarks and operational surveillance networks.

\begin{figure*}[!t]
    \centering
    \includegraphics[width=1\linewidth]{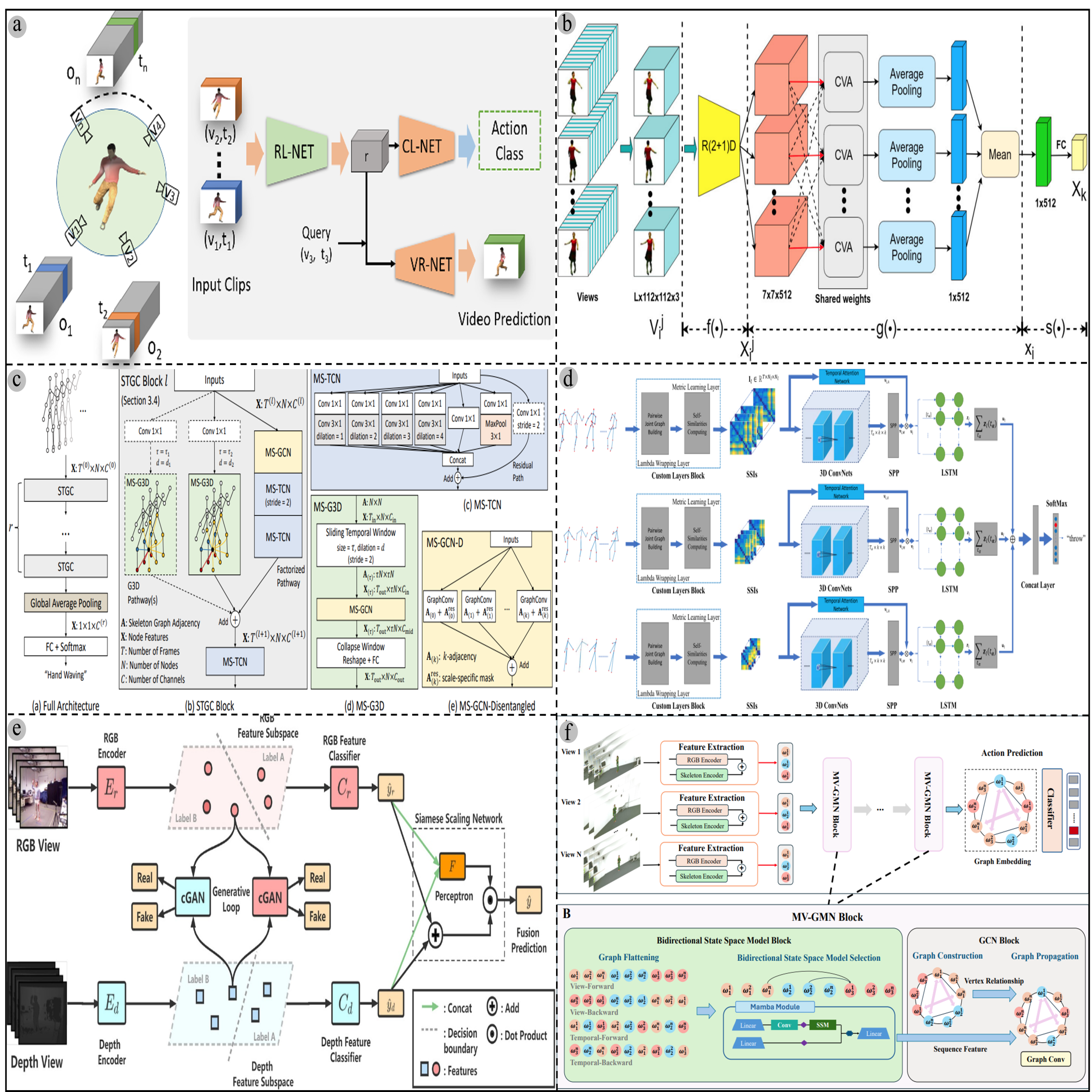}
    \caption{Overview of state-of-the-art Multi-View Multi-Camera AU frameworks, highlighting methods for AR across multiple views and cameras, including approaches for a) cross-view video prediction for action recognition \cite{vyas2020multi}, b) attention-based networks for multi-view action recognition \cite{nguyen2021attention}, c) graph convolutions for skeleton-based action recognition \cite{liu2020disentangling}, d) self-similarity learning for cross-view action recognition \cite{shao2020learning},  e) view knowledge transfer for action recognition \cite{liang2022view}, and f) state space models for multi-view action recognition \cite{lin2023multi}. Each method addresses unique challenges in multi-view, multi-camera Re-ID for action recognition.}
    \label{fig=AU-Frameworks}
\end{figure*}
\subsection{MVMC Re-ID}
MC-ReID is a critical task in CVS, ensuring consistent identity matching of individuals or vehicles across disjoint camera views. Unlike single-camera Re-ID, which deals primarily with intra-camera appearance variation, MC-ReID must contend with drastic viewpoint changes, heterogeneous illumination, occlusion, and limited temporal overlap. The literature can broadly be categorized into traditional feature-based methods, deep embedding and attention-based methods, temporal and video-based models, robust occlusion and viewpoint-adaptive approaches, and domain adaptation and lifelong learning frameworks. The summarized literature is given in Table \ref{table:mvmc_reid} and well-known frameworks are shown in Figure \ref{fig=Re-ID_Frameworks}.

\subsubsection{Traditional Feature-Based Methods}
Early MC-ReID approaches relied on handcrafted descriptors, such as color histograms, SIFT, or HOG, combined with distance metrics or dictionary learning for feature matching \cite{azis2016weighted, murtaza2016multi, shah2016multi, ma2016orientation}. Metric learning techniques such as LFDA and Mahalanobis distance were also adopted to reduce intra-class variance while maximizing inter-class separability. While these methods achieved moderate performance on small-scale benchmarks like iLIDS and PRID, they suffered in unconstrained settings where pose, illumination, and viewpoint variability were significant. Nonetheless, they laid the groundwork for subsequent learning-based approaches by emphasizing the importance of discriminative feature representations.

\subsubsection{Deep Embedding and Attention-Based Methods}
With the advent of DL, CNN-based embedding models quickly surpassed handcrafted features. Architectures such as ResNet and Inception became the backbone for extracting discriminative representations, with metric learning losses (triplet, contrastive) driving performance on large-scale benchmarks like Market-1501 and DukeMTMC-ReID \cite{chen2016deep, qian2017multi, liu2019spatially}. Transformer-based approaches have further advanced this trend. For instance, the Part-Aware Transformer leverages part-level attention to disentangle local and global body cues \cite{ni2023part}, while context-sensing networks refine embeddings through spatial and channel attention \cite{wang2023context}. Attribute-aware dual-branch models extend this by fusing soft biometrics (e.g., gender, clothing attributes) with global appearance cues, enhancing robustness in crowded environments.
\begin{figure}[!t]
    \centering
    \includegraphics[width=1\linewidth, height=0.322\textheight, trim=450pt 85pt 450pt 55pt, clip]{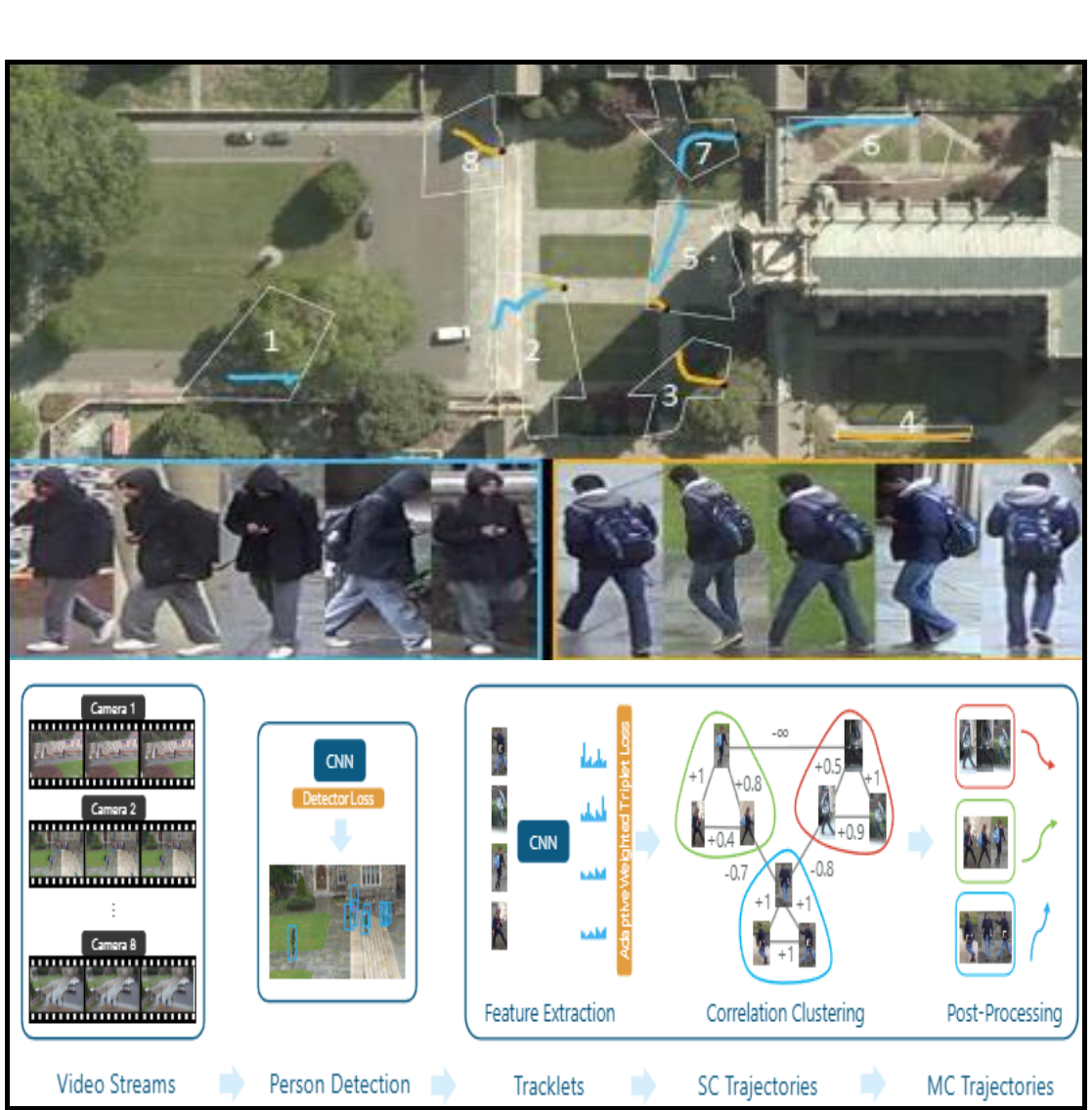}
    \caption{MVMC tracking and Re-Identification as an example of an integrated system. The top sub figure is a representations of a university campus with seven interconnected cameras, both overlapping and non-overlapping. The second sub figure illustrates the tracking pipeline, including person detection, feature extraction, correlation-based trajectory inference, and multi-stage reasoning for trajectory refinement across single and multi-camera views \cite{ristani2016performance}.
}
    \label{fig=Integrated}
\end{figure}
\subsubsection{Temporal and Video-Based Models}
A growing line of research leverages temporal information from video tracklets. Unlike image-based methods, which treat frames independently, temporal Re-ID incorporates motion continuity and cross-frame context. The Trigeminal Transformer exemplifies this direction by integrating spatial, temporal, and spatio-temporal streams with cross-view attention \cite{liu2024video}. Similarly, recursive frameworks such as TAF+TAA+B refine intra-frame pose-aware features and aggregate inter-frame dependencies \cite{ma2025recursively}, while hierarchical models like PSTA employ pyramid-level supervision to progressively refine spatio-temporal embeddings \cite{wang2021pyramid}. These methods consistently outperform static models on video datasets such as MARS and DukeMTMC-VID, demonstrating the utility of temporal cues in Re-ID.

\subsubsection{Occlusion- and Viewpoint-Robust Approaches}
Occlusion and drastic viewpoint changes remain major challenges in MC-ReID. Solutions often involve explicit occlusion modeling or data augmentation. Multi-view integration frameworks propagate features across views \cite{xiahou2023identity, dong2024multi}, while corrupted feature prediction models learn to smooth corrupted embeddings \cite{zhao2024end}, mitigating the impact of missing parts. Pose-guided generative approaches synthesize alternative viewpoints of a given individual, effectively expanding the appearance space \cite{jiao2025generalizable}. These advancements make MC-ReID more viable in real-world surveillance, where crowded environments are the norm.

\subsubsection{Domain Adaptation and Lifelong Learning}
Real-world deployments require Re-ID models that generalize across datasets and domains without extensive re-training. Camera-aware graph-based adaptation methods construct inter-camera graphs and use GCNs to propagate knowledge across domains \cite{jiao2025generalizable}. Anti-forgetting adaptation approaches stabilize unsupervised training with rehearsal modules that preserve past knowledge, mitigating catastrophic forgetting \cite{chen2024anti}. Diverse representation embedding methods balance rapid adaptation with long-term retention \cite{liu2024diverse}. More recently, federated mutual learning enables decentralized Re-ID training across multiple sites without centralizing sensitive data \cite{liu2025personalized}. Collectively, these methods reflect a shift toward scalable, generalizable, and lifelong Re-ID systems suitable for operational CVS pipelines.
\subsection{MVMC Action Understanding}
Multi-view AU (MV-AU) extends beyond detection, tracking, and identity preservation to the higher-level interpretation of human activities across distributed camera networks. Unlike single-view AR, MV-AU must resolve severe viewpoint changes, synchronize distributed inputs, and infer complex group behaviors. The literature can be broadly divided into classical handcrafted approaches, deep CNN and recurrent architectures, skeleton- and graph-based recognition, and transformer and hypergraph-driven models. The collected summarized literature is given in \ref{table:mvmc_action} and popular frameworks of MV AU are presented in Figure \ref{fig=AU-Frameworks}.
\subsubsection{Classical Handcrafted Approaches}
Early MV-AU systems extracted handcrafted descriptors such as optical flow, motion trajectories, and spatio-temporal interest points. Methods like motion history images (MHI) combined with HOG descriptors \cite{murtaza2016multi} or temporal VLAD \cite{naeem2021t} encoded motion cues across views, enabling recognition of simple activities such as walking or boxing. Multi-task learning pipelines such as DivAgg provided cross-view generalization on datasets like IXMAS and NUCLA \cite{wang2018dividing}. These handcrafted approaches provided early baselines for datasets like IXMAS and MuHAVi, but they were inherently limited by sensitivity to noise and lack of scalability to real-world complexity.
\subsubsection{Deep CNN and Recurrent Architectures}
The deep learning era introduced CNN and LSTM-based models for MV-AU. Multi-stream CNNs processed different modalities (RGB, optical flow, depth), while recurrent networks captured temporal dependencies. Conflux LSTM architectures, for instance, integrated per-view CNN feature maps through parallel LSTM streams, followed by inter-view fusion for classification \cite{ullah2021conflux}. Generative frameworks such as VCDN used cross-view discovery modules to align RGB and depth streams \cite{wang2019generative}, while cross-view video prediction models such as CL-Net combined representation learning with video forecasting \cite{vyas2020multi}. These pipelines achieved strong performance on large-scale benchmarks such as NTU RGB+D and PKU-MMD \cite{liu2017pku}, establishing deep CNN–RNN hybrids as dominant architectures for MV-AU.
\subsubsection{Skeleton- and Graph-Based Recognition}
Skeleton-based recognition provides compact, appearance-invariant representations by modeling joints as graph nodes. Graph convolutional networks (GCNs) and temporal convolutional networks (TCNs) have been widely adopted to capture spatio-temporal dependencies in skeleton sequences. MS-G3D employs multi-scale graph temporal convolutions \cite{liu2020disentangling}, while disentangled GCNs improve representation by separating adjacency scales. MV-GMN further integrates state-space modeling with GCNs, capturing bidirectional temporal dependencies across multiple views \cite{lin2025mv}. These methods excel in multimodal benchmarks like NTU120 and PKU-MMD, where skeleton annotations are available, and have become essential components in robust MV-AU pipelines.

\subsubsection{Transformer- and Hypergraph-Based Frameworks}
Recent methods embrace self-attention and hypergraph representations to capture long-range dependencies across views. Hypergraph-based MV-AU constructs vertex-attention hypergraphs to integrate multi-view features \cite{gao2024hypergraph}, outperforming pairwise fusion strategies. Self-supervised frameworks such as MV2MAE apply masked autoencoder pretraining to learn cross-view consistency \cite{shah2024mv2mae}, while multimodal transformer architectures fuse audio and video streams for comprehensive event detection in distributed environments \cite{yamane2024mvaformer, yasuda2022multi}. These advancements reflect the growing convergence between MV-AU and general-purpose multimodal foundation models, marking a clear trend toward pretraining, self-supervision, and large-scale cross-modal learning.
\subsection{MVMC Integrated Approaches}
While MVMC tracking, MC-ReID, and MV-AU have traditionally been studied in isolation, integrated approaches seek to unify them within end-to-end pipelines that more closely resemble real-world CVS as given in Figure \ref{fig=Integrated}. Such frameworks are motivated by the need for holistic scene understanding, where tracking, identity preservation, and behavior analysis are interdependent. The literature can be divided into early hybrids, end-to-end multi-task frameworks, and full connected vision pipelines.

\subsubsection{Early Hybrid Frameworks}
Initial attempts at integration focused on coupling tracking with Re-ID. Methods such as AWTL combined appearance-based Re-ID features with temporal clustering to improve multi-camera trajectory consistency. Other frameworks embedded Re-ID modules within vehicle tracking pipelines, allowing identity information to reinforce association decisions \cite{zheng2017person, tan2019multi}. These early hybrids demonstrated the value of joint optimization, but they were limited by modular design and lacked scalability.

\subsubsection{End-to-End Multi-Task Frameworks}
The next stage of development introduced end-to-end multi-task pipelines explicitly designed to integrate detection, tracking, and Re-ID. The MTMMC baseline exemplifies this approach, combining RGB-thermal detection with ByteTrack and Re-ID modules to achieve robust performance under diverse conditions. GeoReID extended the paradigm by incorporating geometric consistency into cross-camera matching, while CMLM-Veh employed multi-level matching mechanisms for simultaneous MOT and Re-ID \cite{li2025fusiontrack}. These frameworks represent a growing consensus that joint optimization improves robustness and generalization.

\subsubsection{Holistic Connected Vision Pipelines}
The most recent advances move beyond detection and identity preservation to incorporate AU, aiming for complete connected vision pipelines. Argus, for example, enables distributed video analytics across smart camera networks by coordinating tracking, Re-ID, and scene-level analytics in a collaborative framework \cite{yi2024argus}. Similarly, MV-BMOT merges Bayesian filtering with multi-view graph-based tracking for city-scale deployments, achieving scalable real-time performance \cite{yu2025citytrac}. These emerging pipelines underscore the shift toward holistic CVS systems, where low-level perception and high-level reasoning are integrated seamlessly to support intelligent decision-making in smart cities, transportation, and robotics.
\section{EVALUATION METRICS AND STATE-OF-THE-ART}
\subsection{Evaluation Metrics}
To ensure fair comparison across different MVMC connected vision tasks, several evaluation metrics are widely used. For MOT, the most common is the multiple object tracking accuracy (MOTA), which penalizes missed detections, false positives(FP), and identity switches(IDSW), as given in Equation~(1) \cite{ristani2016performance, xu2016multi, fang2024coordinate}:

\begin{equation}
\mathrm{MOTA} = 1 - \frac{\sum_{t}(\mathrm{FN}_t + \mathrm{FP}_t + \mathrm{IDSW}_t)}{\sum_{t}\mathrm{GT}_t}
\end{equation}

Localization precision is calculated by the multiple object tracking precision (MOTP), which measures the average distance (or IoU) between predicted and ground-truth objects, as shown in Equation~(2) \cite{wen2017multi, ong2020bayesian}:

\begin{equation}
\mathrm{MOTP} = \frac{\sum_{t}\sum_{i} d_{t,i}}{\sum_{t} c_t}
\end{equation}

For more robust benchmarking across recall thresholds, the averaged variants AMOTA and AMOTP are often reported as given in Equations~(3) and (4), respectively \cite{ding2024ada, li2024rocktrack}:

\begin{equation}
\mathrm{AMOTA} = \frac{1}{|\mathcal{R}|}\sum_{r \in \mathcal{R}}\mathrm{MOTA}(r)
\end{equation}

\begin{equation}
\mathrm{AMOTP} = \frac{1}{|\mathcal{R}|}\sum_{r \in \mathcal{R}}\mathrm{MOTP}(r)
\end{equation}

A more recent measure, higher order tracking accuracy (HOTA), jointly considers detection (DetA) and association (AssA) quality, as defined in Equation~(5) \cite{cioppa2022soccernet, scott2024teamtrack, yoshida2024overlap}:

\begin{equation}
\mathrm{HOTA} = \sqrt{\mathrm{DetA} \cdot \mathrm{AssA}}
\end{equation}

Identity-based metrics are also critical. The identity F1 score (IDF1) combines identity precision (IDP) and identity recall (IDR), shown in Equation~(6) \cite{zhang1712multi, li2021multi, shim2021multi}:

\begin{equation}
IDF1 = \frac{2 \ast IDTP}{2 \ast IDTP + IDFP + IDFN}
\end{equation}

\renewcommand{\arraystretch}{1.1}
\begin{table}[t]
\centering
\caption{Performance comparison of different state-of-the-art multi-view multi-camera tracking models on the Campus \cite{xu2016multi} dataset across multiple sequences (Garden1, Garden2, Auditorium, and Parking lot). The results are reported in terms of Multiple Object Tracking Accuracy (MOTA$\uparrow$), Multiple Object Tracking Precision (MOTP$\uparrow$), Mostly Tracked targets (MT$\uparrow$), and Mostly Lost targets (ML$\downarrow$).}
\label{tab:campus_sota}

\resizebox{0.49\textwidth}{!}{%
\begin{tabular}{c|l|c|cccc}
\toprule
\textbf{Sequence} & \textbf{Method} & \textbf{Venue,Year} & \textbf{MOTA$\uparrow$} & \textbf{MOTP$\uparrow$} & \textbf{MT$\uparrow$} & \textbf{ML$\downarrow$} \\
\midrule

\multirow{7}{*}{\textbf{Garden1}} 
& POM \cite{ristani2018features}   & CVPR,18  & 22.4 & 64.2 & --   & 43.7 \\
& HCT \cite{zhang1712multi}        & arXiv,17 & 49.0 & 71.9 & 31.3 & 6.3  \\
& TRAC \cite{he2020multi}          & TIP,20   & 58.5 & 74.3 & 30.6 & 1.6  \\
& DyGLIP \cite{quach2021dyglip}    & CVPR,21  & 71.2 & 91.6 & 31.3 & 0.0  \\
& LMGP \cite{nguyen2022lmgp}       & CVPR,22  & 76.9 & 95.9 & 62.9 & 1.6  \\
& ReST \cite{cheng2023rest}        & ICCV,23  & 77.6 & 99.1 & 100  & 0.0  \\
& MHT \cite{wang2023blockchain}    & TII,23   & 82.4 & 78.2 & 100  & 0.0  \\

\midrule
\multirow{6}{*}{\textbf{Garden2}} 
& POM \cite{ristani2018features}   & CVPR,18  & 14.0 & 63.8 & 14.3 & 7.1  \\
& HCT \cite{zhang1712multi}        & arXiv,17 & 25.8 & 71.6 & 33.3 & 11.1 \\
& TRAC \cite{he2020multi}          & TIP,20   & 35.5 & 75.3 & 16.9 & 11.3 \\
& DyGLIP \cite{quach2021dyglip}    & CVPR,21  & 87.0 & 98.4 & 66.7 & 0.0  \\
& ReST \cite{cheng2023rest}        & ICCV,23  & 86.0 & 99.9 & 100  & 0.0  \\
& MHT \cite{wang2023blockchain}    & TII,23   & 79.0 & 76.9 & 91.5 & 8.4  \\

\midrule
\multirow{5}{*}{\textbf{Auditorium}} 
& POM \cite{ristani2018features}   & CVPR,18  & 16.2 & 61.0 & 16.7 & 16.6 \\
& HCT \cite{zhang1712multi}        & arXiv,17 & 20.6 & 69.2 & 33.3 & 11.1 \\
& TRAC \cite{he2020multi}          & TIP,20   & 33.7 & 73.1 & 37.3 & 20.9 \\
& DyGLIP \cite{quach2021dyglip}    & CVPR,21  & 96.7 & 99.5 & 95.2 & 0.0  \\
& ReST \cite{cheng2023rest}        & ICCV,23  & 81.2 & 98.8 & 92.1 & 0.0  \\

\midrule
\multirow{6}{*}{\textbf{Parking Lot}} 
& POM \cite{ristani2018features}   & CVPR,18  & 11.0 & 60.0 & --   & 53.3 \\
& HCT \cite{zhang1712multi}        & arXiv,17 & 24.1 & 66.2 & 40.5 & 26.6 \\
& TRAC \cite{he2020multi}          & TIP,20   & 39.4 & 74.9 & 15.5 & 10.3 \\
& DyGLIP \cite{quach2021dyglip}    & CVPR,21  & 72.8 & 98.6 & 26.6 & 0.0  \\
& LMGP \cite{nguyen2022lmgp}       & CVPR,22  & 78.1 & 97.3 & 62.1 & 0.0  \\
& ReST \cite{cheng2023rest}        & ICCV,23  & 77.7 & 99.8 & 100  & 0.0  \\

\bottomrule
\end{tabular}
}
\end{table}
where identity precision (IDP) and identity recall (IDR) are defined in Equations~(7) and (8), respectively:

\begin{equation}
\mathrm{IDP} = \frac{\mathrm{IDTP}}{\mathrm{IDTP} + \mathrm{IDFP}}
\end{equation}

\begin{equation}
\mathrm{IDR} = \frac{\mathrm{IDTP}}{\mathrm{IDTP} + \mathrm{IDFN}}
\end{equation}

Two additional indicators evaluate track coverage over an object’s lifespan. The mostly tracked (MT) score measures the percentage of ground truth objects that are correctly tracked for most of their duration, as defined in Equation~(9) \cite{nguyen2022lmgp, cheng2023rest}:

\begin{equation}
\mathrm{MT} = \frac{\mathrm{Number \; of \; Mostly \; Tracked \; Objects}}{\mathrm{Total \; Number \; of \; Ground \; Truth \; Objects}} \times 100
\end{equation}

Conversely, the mostly lost (ML) score measures the percentage of objects that were not tracked for most of their duration, as shown in Equation~(10) \cite{yousefi2023tracking, herzog2024spatial}:

\begin{equation}
\mathrm{ML} = \frac{\mathrm{Number \; of \; Mostly \; Lost \; Objects}}{\mathrm{Total \; Number \; of \; Ground \; Truth \; Objects}} \times 100
\end{equation}

In addition, the identity discrepancy score (IDS) quantifies mismatches between predicted and ground-truth identities. It is defined as the proportion of discrepant matches over the total number of matches, given in Equation~(11) \cite{xu2017cross, hao2024divotrack}:

\begin{equation}
\mathrm{IDS} = \frac{\mathrm{Discrepant \; Matches}}{\mathrm{Total \; Matches}} \times 100
\end{equation}

For Re-ID tasks, performance is typically reported using the mean average precision (mAP) and the cumulative matching characteristic (CMC). The mAP averages the per-class precision–recall areas, as shown in Equation~(12) \cite{zheng2016mars}:

\begin{equation}
mAP = \frac{1}{Q} \sum_{q=1}^{Q} AP(q),
\end{equation}
The CMC curve measures the probability that the correct identity appears in the top-$k$ retrieved results. It is defined as in Equation~(13):
\begin{equation}
CMC(k) = \sum_{r=1}^{k} P(r)
\end{equation}

For AU, standard classification metrics are adopted. Precision and recall are given by Equations~(14) and (15), respectively \cite{wang2018dividing, vyas2020multi, gao2024hypergraph}.

\begin{equation}
\mathrm{Precision} = \frac{\mathrm{TP}}{\mathrm{TP} + \mathrm{FP}}
\end{equation}

\begin{equation}
\mathrm{Recall} = \frac{\mathrm{TP}}{\mathrm{TP} + \mathrm{FN}}
\end{equation}

Overall accuracy is defined as in Equation~(16) \cite{murtaza2016multi, wang2019generative, nguyen2021attention}:

\begin{equation}
\mathrm{Accuracy} = \frac{\mathrm{TP} + \mathrm{TN}}{\mathrm{TP} + \mathrm{FP} + \mathrm{FN} + \mathrm{TN}}
\end{equation}

and the harmonic mean of precision and recall is the F1-score, as shown in Equation~(17) \cite{shao2020learning, yamane2024mvaformer, nguyen2024multi}:

\begin{equation}
\mathrm{F1} = \frac{2 \cdot \mathrm{Precision} \cdot \mathrm{Recall}}{\mathrm{Precision} + \mathrm{Recall}}
\end{equation}

Together, these metrics provide a comprehensive view of detection quality, identity consistency, localization precision, and classification reliability in multi-camera connected vision systems.

\renewcommand{\arraystretch}{1.10}
\begin{table}[t]
\centering
\caption{Performance comparison of different state-of-the-art multi-view multi-camera tracking models on the Wildtrack \cite{chavdarovawildtrack} dataset. The results are reported in terms of Multiple Object Tracking Accuracy (MOTA$\uparrow$), Multiple Object Tracking Precision (MOTP$\uparrow$), Mostly Tracked targets (MT$\uparrow$), and Mostly Lost targets (ML$\downarrow$).}
\label{tab:wildtrack_sota}

\resizebox{0.48\textwidth}{!}{%
\begin{tabular}{l|c|cccc}
\toprule
\textbf{Method} & \textbf{Venue,Year} & \textbf{MOTA$\uparrow$} & \textbf{MOTP$\uparrow$} & \textbf{MT$\uparrow$} & \textbf{ML$\downarrow$} \\
\midrule
Wildtrack \cite{chavdarova2018wildtrack} & CVPR,18   & 72.2 & 60.3 & 72.0 & 25.0 \\
GLMB \cite{ong2020bayesian}              & TPAMI,20  & 69.7 & 73.2 & 136  & 37.0 \\
ReST \cite{cheng2023rest}                & ICCV,23   & 81.6 & 81.8 & 79.4 & 4.7  \\
MVFlow \cite{engilberge2023multi}        & WACV,23   & 91.3 & 57.0 & 38.0 & 2.0  \\
EarlyBird \cite{teepe2024earlybird}      & WACV,24   & 89.5 & 86.6 & 78.0 & 4.9  \\
TrackTacu \cite{teepe2024lifting}        & CVPR,24   & 91.8 & 85.4 & 87.8 & 4.9  \\
BEV \cite{alturki2025attention}          & arXiv,25  & 92.7 & 88.8 & 87.8 & 4.9  \\
MCBLT \cite{wang2024mcblt}               & arXiv,25  & 95.6 & 92.6 & 80.5 & 7.3  \\
GCEF \cite{hu2025enhanced}               & ICRCE,25  & 91.0 & 89.0 & 85.4 & 4.9  \\
UMPN \cite{engilberge2025unified}        & arXiv,25  & 93.9 & 86.9 & --   & --   \\
\bottomrule
\end{tabular}
}
\end{table}
\renewcommand{\arraystretch}{1.12}
\begin{table}[b]
\centering
\caption[EPFL dataset performance comparison]{Performance comparison of different state-of-the-art MVMC tracking models on the EPFL \cite{chavdarova2017deep} dataset across multiple sequences (Terrace, Passageway, and Basketball). The results are reported in terms of Multiple Object Tracking Accuracy (MOTA$\uparrow$) and Multiple Object Tracking Precision (MOTP$\uparrow$).}
\label{tab:epfl_sota}

\resizebox{0.48\textwidth}{!}{%
\begin{tabular}{c|l|c|c|c}
\toprule
\textbf{Sequence} & \textbf{Method} & \textbf{Venue,Year} & \textbf{MOTA$\uparrow$} & \textbf{MOTP$\uparrow$} \\
\midrule

\multirow{3}{*}{\textbf{Terrace}} 
& HTC \cite{xu2016multi}          & CVPR,16   & 72.3 & 71.6 \\
& POM \cite{ristani2018features}  & CVPR,18   & 57.5 & 62.6 \\
& DMV-MTT \cite{he2019efficient}  & Sensors,19& 79.9 & 87.5 \\

\midrule
\multirow{5}{*}{\textbf{Passageway}} 
& HTC \cite{xu2016multi}          & CVPR,16   & 43.8 & 67.1 \\
& POM \cite{ristani2018features}  & CVPR,18   & 32.6 & 60.9 \\
& DMV-MTT \cite{he2019efficient}  & Sensors,19& 62.2 & 90.6 \\
& TRACTA \cite{he2020multi}       & TIP,20    & 64.3 & 72.5 \\
& DyGLIP \cite{quach2021dyglip}   & CVPR,21   & 70.4 & 97.2 \\

\midrule
\multirow{3}{*}{\textbf{Basketball}} 
& HTC \cite{xu2016multi}          & CVPR,16   & 60.0 & 68.0 \\
& TRACTA \cite{he2020multi}       & TIP,20    & 64.3 & 72.5 \\
& DyGLIP \cite{quach2021dyglip}   & CVPR,21   & 66.3 & 89.5 \\

\bottomrule
\end{tabular}
}
\end{table}

\subsection{Analysis and Trends}
The field of MVMC CVS has witnessed significant growth, particularly over the last decade, driven by the integration of DL techniques and multi-view analytics. The volume of scholarly contributions addressing core tasks like MV-T, MC-ReID, and MV-AU has increased substantially since 2019, with a peak observed in 2024. This surge highlights the increasing importance of CVS for real-time and large-scale applications. One of the main trends in recent years is the shift from isolated tasks toward integrated frameworks that simultaneously handle tracking, Re-ID, and AU across multiple cameras. This trend is evident from the increasing use of end-to-end models, which combine these tasks within a unified system for better performance and more robust real-world deployment. Additionally, there has been a clear diversification in the types of datasets used for benchmarking these tasks. Datasets like DukeMTMC, Campus, and WILDTRACK continue to play a central role in tracking and Re-ID, while NTU RGB+D 120 and PKU-MMD are heavily leveraged for AU, particularly in multi-view settings. The variety and complexity of these datasets are contributing to the development of more generalized algorithms that can operate across different environments and modalities, including RGB, depth, and thermal imagery.
The performance of state-of-the-art algorithms in CVS has steadily improved, thanks to advancements in DL, spatio-temporal reasoning, and graph-based methods. Techniques like contrastive learning and multi-modal fusion are becoming increasingly common, enhancing the accuracy of tracking and Re-ID, as well as enabling robust AR even in complex and dynamic environments. The performance of different models on popular datasets across various CVS tasks is discussed in the following section. 

\renewcommand{\arraystretch}{1.12}
\begin{table}[t]
\centering
\caption[CityFlow dataset results]{Performance comparison of MVMC tracking models on the CityFlow \cite{tang2019cityflow} dataset. The results are reported in terms of Multiple Object Tracking Accuracy (MOTA$\uparrow$) and Multiple Object Tracking Precision (MOTP$\uparrow$).}
\label{tab:cityflow_sota}

\resizebox{0.485\textwidth}{!}{%
\begin{tabular}{p{2.5cm}|c|c|c}
\toprule
\textbf{Method} & \textbf{Venue,Year} & \textbf{MOTA$\uparrow$} & \textbf{MOTP$\uparrow$} \\
\midrule
DeepSORT \cite{wojke2017simple} & ICIP,17            & 61.4 & 79.1 \\
TC \cite{tang2018single}        & CVPRW,18           & 70.3 & 65.6 \\
MOANA \cite{tang2019moana}      & IEEE Access,19     & 67.0 & 65.9 \\
TSCT \cite{hsu2021multi}        & TIP,21             & 83.4 & 75.2 \\
\bottomrule
\end{tabular}
}
\end{table}

\subsection{Performance Comparison}
Evaluating MVMC systems requires rigorous benchmarking across diverse datasets that reflect real-world conditions such as occlusion, varying viewpoints, crowd density, and illumination changes. To this end, a wide range of datasets including Campus, Wildtrack, EPFL, CityFlow, MARS, DukeMTMC-reID, and NTU RGB+D 120 have become standard benchmarks for assessing the three core tasks of tracking, Re-ID, and AU. Each dataset emphasizes different aspects of MVMC challenges: Campus and EPFL capture constrained campus-scale environments, Wildtrack and CityFlow address crowded pedestrian and large-scale traffic scenarios, while MARS and DukeMTMC-reID benchmark identity preservation across disjoint views, and NTU RGB+D 120 provides a comprehensive testbed for cross-view AR. Together, these benchmarks (Tables VIII–XIV) allow systematic comparison of state-of-the-art methods, tracing the evolution from handcrafted pipelines to graph-based association and transformer-driven architectures.

%%%%%%%%%%%%%%%%%%%%%%%%%%%%%%%%%%%%%%%%%%%%%%%%%%%%%%%%%%%%%%%%%%%%%%%%%%%%%%%%%%%%%%%%%%%%%%%%%%%%%%%%%%%%%%%%%%%%
The results in Table \ref{tab:campus_sota} (Campus dataset) illustrate this evolution clearly. Early methods such as POM~\cite{ristani2018features} and HCT~\cite{zhang1712multi} reported limited accuracy, with MOTA values under 50 and high ML rates. Learning-based designs like DyGLIP~\cite{quach2021dyglip} leveraged spatio-temporal reasoning to push MOTA beyond 70, while graph-driven frameworks such as LMGP~\cite{nguyen2022lmgp} further improved robustness. More recent solutions, including ReST~\cite{cheng2023rest} and MHT~\cite{wang2023blockchain}, achieved near-perfect continuity with 100\% MT and almost zero ML, demonstrating the maturity of MVMC tracking pipelines.
The Wildtrack dataset results in Table \ref{tab:wildtrack_sota} follow a similar trend. The original Wildtrack baseline~\cite{chavdarova2018wildtrack} and Bayesian models like GLMB and GLMB-DO~\cite{ong2020bayesian} delivered moderate accuracy ($\approx 70$ MOTA) with limited precision. Transformer- and graph-based methods including ReST~\cite{cheng2023rest}, EarlyBird~\cite{teepe2024earlybird}, and TrackTacular~\cite{teepe2024lifting} significantly boosted performance to above 85 MOTA, while BEV~\cite{alturki2025attention} and MCBLT~\cite{wang2024mcblt} established new state-of-the-art, reporting $95.6$ MOTA and $92.6$ MOTP. These results highlight the importance of spatio-temporal context modeling and geometric fusion for robust multi-camera tracking.
\renewcommand{\arraystretch}{1.12}
\begin{table}[t]
\centering
\caption[MARS dataset results]{Performance comparison of state-of-the-art MC person Re-ID models on the MARS \cite{zheng2016mars} dataset. Results are reported in terms of Rank-1 accuracy (Top-1/R1$\uparrow$) and mean Average Precision (mAP$\uparrow$).}
\label{tab:mars_sota}

\resizebox{0.48\textwidth}{!}{%
\begin{tabular}{l|c|c|c}
\toprule
\textbf{Method} & \textbf{Venue,Year} & \textbf{Top-1/R1$\uparrow$} & \textbf{mAP$\uparrow$} \\
\midrule
Mars \cite{zheng2016mars}           & ECCV,16   & 68.3 & 49.3 \\
RQEN \cite{song2018region}          & AAAI,18   & 77.8 & 71.1 \\
Yin et al. \cite{yin2020fine}       & IJCV,20   & 82.9 & 66.9 \\
STE-NVAN \cite{liu2019spatially}    & arXiv,19  & 88.9 & 81.2 \\
GRL \cite{liu2021watching}          & CVPR,21   & 91.0 & 84.8 \\
BiCnet-TKS \cite{li2021multi}       & CVPR,21   & 90.2 & 86.0 \\
STMN \cite{eom2021video}            & ICCV,21   & 90.5 & 84.5 \\
PSTA \cite{wang2021pyramid}         & ICCV,21   & 91.5 & 85.8 \\
TMT \cite{liu2024video}             & TITS,24   & 91.8 & 86.5 \\
CLIP-ReID \cite{li2023clip}         & AAAI,23   & 91.7 & 88.1 \\
AG-VPReID-Net \cite{nguyen2025ag}   & CVPR,25   & 93.2 & 91.5 \\
\bottomrule
\end{tabular}
}
\end{table}
\renewcommand{\arraystretch}{1.10}
\begin{table}[!b]
\centering
\caption[DukeMTMC-reID dataset results]{Performance comparison of different state-of-the-art person Re-ID models on the DukeMTMC-reID \cite{zheng2017unlabeled} dataset. Results are reported in terms of Rank-1 accuracy (R1$\uparrow$), Rank-5 accuracy (R5$\uparrow$), and mean Average Precision (mAP$\uparrow$).}
\label{tab:duke_sota}

\resizebox{0.48\textwidth}{!}{%
\begin{tabular}{l|c|c|c|c}
\toprule
\textbf{Method} & \textbf{Venue,Year} & \textbf{R1$\uparrow$} & \textbf{R5$\uparrow$} & \textbf{mAP$\uparrow$} \\
\midrule
LSRO \cite{zheng2017unlabeled}      & ICCV,17    & 67.6 & --   & 47.1 \\
MVC \cite{xin2019semi}              & PR,19      & 55.7 & --   & 37.8 \\
FED \cite{dong2024multi}            & CVPR,22    & 91.9 & 95.3 & 80.0 \\
AET-Net \cite{dong2024hierarchical} & Access,23  & 94.0 & 96.4 & 91.1 \\
PAT \cite{ni2023part}               & ICCV,23    & 88.8 & --   & 78.2 \\
ReMix \cite{mamedov2025remix}       & WACV,25    & 77.6 & --   & 61.0 \\
\bottomrule
\end{tabular}
}
\end{table}
On the EPFL dataset (Table \ref{tab:epfl_sota}), older methods such as POM~\cite{ristani2018features} and HTC~\cite{xu2016multi} performed modestly, with MOTP values around 60–70. Later methods like TRACTA~\cite{he2020multi} and DyGLIP~\cite{quach2021dyglip} markedly improved precision, achieving over 97 MOTP. DMV-MTT~\cite{he2019efficient} provided balanced accuracy across all sequences, reinforcing EPFL’s role as a benchmark for evaluating fine-grained multi-view precision.
In large-scale urban monitoring, Table \ref{tab:cityflow_sota} (CityFlow dataset) shows that classical approaches like DeepSORT~\cite{wojke2017simple} and MOANA~\cite{tang2019moana} remained under 70 MOTA, while spatio-temporal correlation tracking (TSCT~\cite{hsu2021multi}) substantially increased performance to 83.4 MOTA, demonstrating the impact of temporal aggregation in handling city-scale vehicle tracking. For person Re-ID, Table \ref{tab:mars_sota} (MARS dataset) demonstrates remarkable progress. The baseline Mars model~\cite{zheng2016mars} achieved only 68.3 Rank-1 and 49.3 mAP, whereas attention and temporal aggregation methods such as RQEN~\cite{song2018region}, STE-NVAN~\cite{liu2019spatially}, and PSTA~\cite{wang2021pyramid} significantly advanced performance. Transformer-based CTL~\cite{li2021multi} and temporal memory networks like STMN~\cite{eom2021video} crossed the 90 Rank-1 threshold, with the most recent CLIP-ReID~\cite{li2023clip} and AG-VPReID-Net~\cite{nguyen2025ag} achieving 93.2 Rank-1 and 91.5 mAP, setting new state-of-the-art. Comparable gains are reflected in Table \ref{tab:duke_sota} (DukeMTMC-reID dataset). Early CNN models such as LSRO~\cite{zheng2017unlabeled} achieved modest results ($\approx 68 R1, 47 mAP$), while meta-learning and federated approaches such as HCT~\cite{liu2022task} and FedUCA~\cite{liu2022task} improved robustness (81.2 R1, 66.5 mAP). Transformer-based frameworks like TransReID~\cite{li2021multi} and PAT~\cite{ni2023part} advanced the state-of-the-art to above 90 R1 and 80 mAP, while PerFedDual~\cite{liu2025personalized} introduced privacy-preserving federated strategies without sacrificing performance.

\renewcommand{\arraystretch}{1.12}
\begin{table}[t]
\centering
\caption[NTU RGB+D 120 \cite{liu2019ntu} dataset results]{Performance comparison of different state-of-the-art models for multi-view and multi-camera AU on the NTU RGB+D 120 \cite{liu2019ntu} dataset. Results are reported in terms of Cross-View (CV$\uparrow$) and Cross-Subject (X-Sub$\uparrow$) accuracies.}
\label{tab:ntu_rgbd120_sota}
\resizebox{0.48\textwidth}{!}{%
\begin{tabular}{l|c|c|c}
\toprule
\textbf{Method} & \textbf{Venue,Year} & \textbf{CV$\uparrow$} & \textbf{X-Sub$\uparrow$} \\
\midrule
MS-G3D Net \cite{liu2020disentangling} & CVPR,21  & 88.4 & 86.9 \\
ViewCon \cite{shah2023multi}           & WACV,23  & 87.5 & 85.6 \\
MV2MAE \cite{shah2024mv2mae}           & arXiv,24 & 87.1 & 85.3 \\
MV-GMN \cite{lin2025mv}                & arXiv,25 & 96.7 & 97.3 \\
\bottomrule
\end{tabular}
}
\end{table}
Finally, in AU, the NTU RGB+D 120 dataset results in Table \ref{tab:ntu_rgbd120_sota} confirm the pivotal role of multi-view fusion. Earlier models such as  MS-G3D~\cite{liu2020disentangling} reported around 88–92 accuracy under cross-view and cross-subject settings. More recent unsupervised segmentation and hierarchical learning frameworks, including  UVS-H-BLS~\cite{liu2024multi}, reached 97.1 Cross-View and 96.6 Cross-Subject accuracy, while MV-GMN~\cite{lin2025mv} sustained performance above 96 Cross views and Subject, showcasing the maturity of state-space and graph-driven approaches for MVMC AR.

The comparisons across Tables VIII–XIV highlight three overarching trends. First, steady performance gains are evident across all datasets as methods evolve from handcrafted features toward deep spatio-temporal, graph, and transformer-driven models. Second, large-scale and challenging benchmarks such as Wildtrack, MARS, and NTU RGB+D 120 have acted as catalysts for innovation, driving methodological advances in tracking, Re-ID, and AU. Third, there is a clear movement toward unified MVMC pipelines that integrate these tasks, setting the foundation for real-world CVS capable of handling the complexity of dynamic, distributed environments.

\renewcommand{\arraystretch}{1.10}
\begin{table*}[b]
\centering
\caption{Open Research Questions for MVMC Connected Vision Systems.}
\label{tab:emerging_research_questions}

\resizebox{0.98\textwidth}{!}{%
\begin{tabular}{c p{16.8cm}}
\toprule
\textbf{\#} & \textbf{Research Question} \\
\midrule
1  & What methodologies are effective in integrating MVMC tracking, Re-ID, and action recognition into a unified CVS? \\ \midrule
2  & How can CVS be designed to remain robust in dynamic environments with occlusions, crowding, and varying lighting and viewpoints? \\ \midrule
3  & How can MVMC CVS be optimized to balance accuracy, computational efficiency, and real-time performance, especially when deployed at scale? \\ \midrule
4  & What strategies can be implemented to ensure privacy-preserving mechanisms are integrated into CVS without compromising overall performance? \\ \midrule
5  & How can lifelong learning be applied to enable CVS to continuously adapt to new environments, conditions, and tasks without requiring retraining? \\ \midrule
6  & How can zero-shot learning techniques be effectively used to enable CVS to recognize unseen identities or actions across different camera views? \\ \midrule
7  & What strategies are necessary for designing and curating MVMC datasets that accurately reflect real-world complexities for fair evaluation? \\ \midrule
8  & How can CVS achieve reliable generalization across diverse environments and scenarios without the need for extensive retraining on new data? \\ \midrule
9  & How can multi-task learning, edge computing, and federated learning contribute to improving the integration of CVS? \\ \midrule
10 & What role does the integration of multi-modal data, such as infrared (IR), audio, and RGB, play in enhancing the robustness of CVS? \\ \midrule
11 & What evaluation protocols and benchmarking strategies are needed to assess the performance of CVS, considering their multi-task capabilities? \\ \midrule
12 & How can CVS handle sensor failures, noisy inputs, and incomplete data across multiple views while ensuring reliable and continuous operation? \\ \midrule
13 & What methodologies can improve the interpretability and transparency of CVS, ensuring that their decision-making processes are explainable? \\ \midrule
14 & How can CVS be deployed at large scale across dynamic environments, such as smart cities, while ensuring robustness and privacy? \\ \midrule
15 & How can end-to-end CVS be designed to integrate tracking, Re-ID, and AU with minimal human intervention? \\ \midrule
16 & How can camera collaborations and environmental context learning be effectively incorporated into CVS to enhance scene understanding? \\ \midrule
17 & How can trajectory prediction be effectively integrated with MVMC tracking and Re-ID to forecast future movements across views and cameras? \\ \midrule
18 & What frameworks are needed to anticipate human actions in multi-view environments, enabling proactive rather than reactive CVS decision-making? \\ 
\bottomrule
\end{tabular}
}
\end{table*}

\section{CHALLENGES AND FUTURE RESEARCH DIRECTIONS}

CVS, particularly those employing MVMC data, are critical components in autonomous vehicles, smart cities, and surveillance applications. These systems, which combine multi-view tracking, person Re-ID, and AU, have made significant strides over the last decade. However, despite considerable advancements, there are still significant challenges in their deployment in dynamic, real-world environments. This section identifies the key challenges, open research questions (\ref{tab:emerging_research_questions}), and proposes future research directions to address these issues, ensuring CVS can be deployed effectively at scale.

\subsection{Datasets: Expanding Scope and Diversity}

One of the most pressing challenges for advancing CVS is the lack of comprehensive and truly integrated datasets. Existing benchmarks such as DukeMTMC~\cite{ristani2016performance}, MARS~\cite{zheng2016mars}, and NTU RGB+D 120~\cite{liu2019ntu} have each played a crucial role in advancing specific sub-tasks like multi-camera tracking, person Re-ID, or AU. However, these datasets are largely developed with a single primary task in mind and are rarely suitable for training or evaluating complete end-to-end CVS pipelines. As a result, although they significantly contribute to individual components of CVS, they do not provide a unified platform where tracking, Re-ID, and AU can be jointly studied within the same framework. 
Moreover, most existing datasets are limited to controlled or semi-controlled environments that fail to capture the full variability of real-world scenarios. They typically lack diversity in environmental conditions such as lighting, weather, and crowd density, all of which are critical for robust real-world deployment. In addition, they are predominantly visual-only, neglecting complementary modalities like thermal, LiDAR, and audio that could strengthen system performance under challenging conditions such as low illumination or heavy occlusion.  Another missing element in current datasets is the modeling of camera collaboration. Real-world CVS operates across large, distributed networks of cameras, where inter-camera handovers, non-overlapping fields of view, and collaborative reasoning about shared scenes are essential. Yet, most datasets treat cameras as independent sources of data, without explicitly capturing collaborative dynamics such as cross-view synchronization, spatial topology, or coordinated decision-making. This gap limits the ability to design and evaluate algorithms that leverage multi-camera cooperation, which is a defining characteristic of CVS in practice. 
To overcome these limitations, future research must focus on developing holistic, multi-task, multi-modal datasets that are explicitly designed for CVS as a whole. Such datasets should allow for the joint training and evaluation of all core tasks tracking, Re-ID, and AU within a single benchmark. They should also capture long-term variations, including seasonal changes, clothing differences, and sensor degradation, to better reflect the dynamics of real-world environments. Importantly, new benchmarks should integrate explicit modeling of camera collaboration, incorporating diverse spatial layouts, overlapping and non-overlapping views, and communication between cameras. By creating datasets that combine multi-view, multi-modal information and emphasize camera collaboration across diverse contexts, the community can enable the development of scalable, robust, and end-to-end CVS models that bridge the gap between research prototypes and operational deployment.
\vspace{-1em}
\subsection{Efficient Modeling and Scalability}
CVS, particularly those working with multi-camera, multi-view data, must handle large amounts of information generated by numerous cameras while ensuring real-time performance. Current DL models, such as CNNs and transformers, have demonstrated impressive results in multi-view tracking and Re-ID. However, these models are often computationally expensive and require significant hardware resources, limiting their scalability for large-scale, real-time applications. To overcome these challenges, future research should focus on developing lightweight models that retain high accuracy while operating efficiently in real-time environments.  Techniques such as model pruning, quantization, and knowledge distillation have shown promise in reducing the computational cost of deep models without significantly sacrificing performance~\cite{cheng2023rest}. Additionally, edge computing frameworks can offload computationally intensive tasks to local devices, enabling faster data processing and reducing latency in multi-camera systems. By adopting these techniques, future systems will be more scalable and able to perform in real-time, even in large-scale, dynamic environments.

\subsection{Lifelong Learning and Adaptability}

A critical challenge for CVS is the need to adapt to new environments and learn incremently. Real-world conditions are constantly changing, with variations in lighting, weather, and the introduction of new objects or individuals. For CVS to remain effective in these environments, they must be able to learn continuously without requiring complete retraining. Traditional machine learning models are static and unable to adapt to these dynamic conditions. Lifelong learning (LL) offers a promising solution to this challenge. LL allows systems to incrementally learn from new data while retaining previously acquired knowledge. This is particularly important for applications like urban surveillance, where environmental changes and the addition of new cameras or perspectives are frequent. Future research should focus on developing incremental learning models that enable CVS to retain and adapt knowledge over time without sacrificing previous performance. Additionally, these models should be task-agnostic, meaning that they can simultaneously learn tracking, Re-ID, and AU tasks, thus improving the system's overall robustness in dynamic environments\cite{chen2024anti}.

\subsection{Zero-Shot Learning and Generalization Across Domains}

CVS often struggle with unseen identities or novel actions that were not part of the training set. Most models are trained on specific identities or actions, making it difficult for them to recognize new ones. Zero-shot learning (ZSL) offers an effective approach to tackle this issue, as it enables systems to recognize unseen entities based on semantic knowledge rather than requiring explicit training data for each new instance. ZSL is especially useful in person Re-ID, where systems need to identify individuals across different camera views, even if those individuals were not included in the training data. Approaches like semantic embeddings or attribute-based models can help in cross-view identity matching, allowing systems to generalize to unseen identities based on prior knowledge. Future research should focus on improving zero-shot learning for multi-camera tracking, Re-ID, and AR, enabling systems to handle new environments with minimal retraining and supervision~\cite{li2021multi}.

\subsection{Privacy and Ethical Considerations}

As CVS becomes more widespread, particularly in public spaces, there are increasing concerns regarding privacy and ethical implications. These systems' ability to track and identify individuals across multiple cameras can lead to unauthorized surveillance and the potential for data misuse. Protecting individual privacy while maintaining system functionality is a critical issue. To address these concerns, future research should explore privacy-preserving techniques such as federated learning and differential privacy, which allow for secure data processing without compromising privacy. Federated learning enables models to be trained on decentralized data, ensuring that sensitive data is not shared across devices, thus maintaining data privacy. Moreover, ethical frameworks and transparency in decision-making processes are essential for ensuring the responsible use of connected vision systems. Research should also focus on making these systems explainable and ensuring they respect individual rights and civil liberties in surveillance and security applications~\cite{liu2022task}.

\subsection{Real-Time Processing and Multi-Modal Integration}

Real-time processing and multi-modal integration remain some of the most significant challenges for large-scale CVS. In applications like autonomous driving, robotic navigation, and urban surveillance, real-time decision-making is critical, and delays in processing can have serious consequences. The integration of multi-modal data, such as RGB, thermal, LiDAR, and audio, is crucial for improving the robustness of these systems under challenging conditions like low light, motion blur, and occlusion. By combining data from multiple sensors, CVS can provide a more comprehensive understanding of the environment, making it more reliable and adaptable. Future research should focus on developing multi-modal fusion techniques that can effectively integrate data from various sensor types while ensuring real-time performance. These advances will be key to enabling CVS to function in complex, dynamic environments such as city-wide surveillance or autonomous vehicle fleets~\cite{fan2025all, liu2025beyond}.
\subsection{Trajectory Prediction and Action Anticipation}

Another critical frontier for future CVS research lies in trajectory prediction and action anticipation. While current systems excel at detecting, tracking, and re-identifying objects across multiple views, their capabilities are still largely reactive, focusing on interpreting events that have already occurred. For many real-world applications such as autonomous driving, intelligent surveillance, and human–robot interaction predictive capabilities are equally important. Trajectory prediction enables systems to forecast the future movement of targets across cameras, accounting for environmental constraints, inter-agent interactions, and spatio-temporal dynamics \cite{nguyen2024multi}. Similarly, action anticipation seeks to infer forthcoming human behaviors before they are fully executed, allowing proactive interventions in safety-critical contexts (e.g., predicting if a pedestrian is about to cross a street) \cite{nguyen2025action, kong2022human}. Despite recent progress using recurrent neural networks, transformers, and graph-based models \cite{gao2024hypergraph}, trajectory forecasting remains challenging in multi-camera environments due to occlusions, scene complexity, and the need to reconcile heterogeneous viewpoints. Action anticipation is further complicated by subtle motion cues, long-term dependencies, and intra-class variability of actions \cite{stergiou2025time, shah2023multi}. 
Future research should therefore prioritize the integration of predictive modeling into CVS pipelines. Promising directions include (i) joint learning of tracking, Re-ID, and trajectory prediction to maintain consistent identity forecasting across disjoint views \cite{cheng2023rest, quach2021dyglip}, (ii) leveraging multimodal cues such as pose, motion fields, and scene semantics for more accurate anticipation of human actions \cite{liu2024video, lin2025mv}, (iii) designing uncertainty-aware models capable of reasoning over multiple plausible futures \cite{fan2025all}, and (iv) advancing temporal representation learning strategies that can capture both short-term and long-term dependencies for robust anticipation \cite{nguyen2025action, gao2024hypergraph}. 
Incorporating trajectory prediction and action anticipation into CVS will shift to more intelligent and responsive deployments in dynamic real-world scenarios.
\section{CONCLUSION}
In this paper, we have provided a comprehensive review of CVS, focusing on the integration of MVMC tracking,  object Re-ID, and AU into cohesive systems. We have highlighted the critical challenges involved in deploying CVS in real-world environments, including occlusions, dynamic conditions, and environmental variability. Despite significant progress, several open challenges remain, particularly with regard to scalability, real-time performance, and data fusion across diverse sensor modalities. To address these complexities, we proposed a unique taxonomy that systematically divides CVS into four main areas: MVMC tracking, Re-ID, AU, and combined methods. This taxonomy serves as a structured framework for organizing the vast array of methods and datasets in the field. It also facilitates a clearer understanding of their relationships. Furthermore, we have identified and reviewed the most widely used datasets, highlighting their role in driving the development of CVS and identifying gaps that must be addressed in future research. This survey also presents key research questions for the future of CVS, with particular emphasis on areas such as lifelong learning, privacy preservation, federated learning, and multi-modal data integration. These emerging technologies hold great potential for improving the adaptability, robustness, and efficiency of CVS in dynamic environments. By presenting an integrated perspective on CVS, this paper provides both a reference for current state-of-the-art approaches and a roadmap for future research. We hope that this work will inspire new solutions and innovative methods to address the challenges facing CVS and contribute to the next generation of intelligent, real-time visual systems deployed across a variety of complex, real-world applications.
\bibliographystyle{IEEEtran}
\bibliography{Ref}
\vfill
\end{document}